\documentclass{article} 
\usepackage{iclr2027_conference,times}

\usepackage{amsmath,amsfonts,bm}

\def\eqref#1{equation~\ref{#1}}

\def\1{\bm{1}}

\def\rr{{\textnormal{r}}}

\def\rw{{\textnormal{w}}}

\DeclareMathAlphabet{\mathsfit}{\encodingdefault}{\sfdefault}{m}{sl}
\SetMathAlphabet{\mathsfit}{bold}{\encodingdefault}{\sfdefault}{bx}{n}

\usepackage{hyperref}
\usepackage{url}
\usepackage{amsmath}    
\makeatletter
\renewcommand{\eqref}[1]{\textup{Eq.~\tagform@{\ref{#1}}}}
\makeatother
\usepackage{amssymb} 
\usepackage{multirow}  
\usepackage[table]{xcolor} 
\usepackage{booktabs}   
\usepackage{graphicx}   
\usepackage{titlesec}
\usepackage{graphicx,tikz,ifthen}
\usepackage{array}
\usepackage{arydshln}
\usepackage{xcolor}
\usepackage{wrapfig}
\usepackage{subcaption}
\titlespacing*{\section}    {0pt}{5pt plus 1pt minus 4pt}{1pt}
\titlespacing*{\subsection} {0pt}{5pt plus 1pt minus 4pt}{1pt}

\definecolor{capgreen}{RGB}{60,140,80}
\definecolor{capblue}{RGB}{45,95,175}

\usepackage{tikz}
\usetikzlibrary{tikzmark}
\newcommand{\dbx}[2]{%
  \begin{tikzpicture}[remember picture, overlay]
    \draw[densely dashed, line width=0.6pt, gray!70, rounded corners=1.5pt]
      ([shift={(-3pt,9pt)}]pic cs:#1) rectangle ([shift={(3pt,-3.5pt)}]pic cs:#2);
  \end{tikzpicture}}

\usepackage{enumitem}
\setlist[itemize]{topsep=2pt, itemsep=1pt, parsep=0pt, partopsep=0pt}

\title{Light Field Primitive for Novel View Synthesis}

\author{Liang Chen$^1$~ 
Jiahui Ning~ 
Xun Jiang$^1$~ 
Xing Xu$^{1}$~ 
Jimmy Ren$^{2}$~ 
Fenglei Fan$^{3}$~ 
Heng Tao Shen$^{1}$\\
 {$^1$~Tongji University}\quad  {$^2$~Hong Kong Metropolitan University}\quad
 {$^3$~City University of Hong Kong}
}

\newlength{\panelw}
\newlength{\panelh}

\renewcommand{\paragraph}[1]{\par\smallskip\noindent\textbf{#1}\ \ignorespaces}

\iclrfinalcopy 
\begin{document}

\definecolor{cbest}{rgb}{1.0,0.55,0.55}
\definecolor{csnd}{rgb}{1.0,0.78,0.52}
\definecolor{ctrd}{rgb}{1.0,0.95,0.62}
\newcommand{\best}[1]{\cellcolor{cbest}#1}
\newcommand{\snd}[1]{\cellcolor{csnd}#1}
\newcommand{\trd}[1]{\cellcolor{ctrd}#1}

\maketitle

\begin{abstract}
We present \textbf{L}ight \textbf{F}ield \textbf{P}rimitives (\textbf{LFP}), a formulation for novel view synthesis that replaces the dense ray database \citep{levoy1996light} with a compact set of differentiable primitives in the classical two-plane parameterization. Each primitive condenses a group of rays into one learned record, and its response to a query is governed by how closely that query belongs to the group. Rendering a camera ray then reduces to compositing all responses it elicits, and a scene can be optimized directly from posed images and rendered in real time with rays. Beyond its competitive performance on standard benchmarks, the main advantage of LFP is structural: its primitives reside directly in the 4D ray space, so optical and appearance effects that are already operations on the light field become behaviors of a single shared renderer. With minimal changes to that renderer, LFP supports multi-scale anti-aliasing, defocus deblurring with refocusing, rendering for fisheye cameras, and even transparent object reconstruction with ray refraction, matching specialized frameworks that devote substantial machinery to these effects.
\end{abstract}

\section{Introduction}
In recent years, novel view synthesis (NVS) has been reshaped by neural
radiance fields and point-based rendering. NeRF~\citep{mildenhall2021nerf}
represents a scene as a continuous radiance field in 3D space and queries it
along camera rays, and 3DGS~\citep{kerbl20233d} represents it as a set of
optimizable 3D Gaussians and rasterizes them for real-time rendering.
Despite their different designs, both paradigms anchor
the scene in 3D space. A camera, however, does not measure a quantity at a
point in space; it measures the radiance arriving along each of its rays.
Their renderers must therefore mediate between the two, restating one side in terms the other can evaluate: NeRF turns a ray into point samples, the form a field takes values on; 3DGS restates each Gaussian as a screen-space
footprint that a pixel can evaluate, and later variants restate the ray in each Gaussian's own frame, solving for where it crosses
one~\citep{huang20242d,moenneloccoz20243d}. Mechanisms differ, but each connects a ray measurement to a representation anchored elsewhere.

Under the ideal model where a pixel is represented by one ray, their optimizations can learn scene parameters that read out correctly through the mediations. But every departure from the ideal model is a statement about rays: a coarser resolution widens the rays a pixel gathers into aliasing; a finite aperture spreads a ray bundle into defocus. Neither statement can be translated directly into space-anchored scene variables in NeRF nor 3DGS, and each must instead be reintroduced by rebuilding the renderer with extra complexity: dedicated anti-aliasing filters for scaling \citep{yu2024mip, barron2021mip}, explicit defocus models for aperture \citep{wang2025dof,lee2024sharp}.

These complexities can be dissolved naturally once the scene is indexed by rays. As promised by classical light field rendering (LFR)~\citep{levoy1996light}, which parameterizes radiance by the intersections of rays with two parallel planes, anti-aliasing and defocus deblurring reduce to their clean optical definitions: prefiltering the ray database naturally produces antialiased renderings~\citep{levoy1996light}; averaging rays admitted by a synthetic aperture gives post-hoc refocusing~\citep{isaksen2000dynamically}, and omitting that averaging automatically yields deblurring. Both departures can be settled in the coordinates the representation already carries.
However, classical light fields have to be captured on a dense grid of rays, incurring prohibitive storage overhead and restricting novel views to a fixed viewing volume. Lacking an underlying continuous representation, they could only interpolate between recorded rays rather than being optimized against a set of posed images, remaining static lookup tables.

We bridge this historical divide with LFP, a formulation that brings the differentiable point-based optimization to the 4D ray space. LFP represents a scene as a compact set of primitives defined in the classical two-plane parameterization. Unlike classical light fields, which store radiance independently for each ray, a primitive is a learnable record shared by a group of rays: a 4D coordinate anchors one ray in the group, a
disparity~\citep{chai2000plenoptic} supplies a coefficient, and together they define the affine coupling between the two plane coordinates of every ray in that group, the epipolar geometry~\citep{bolles1987epipolar} written in these coordinates; an anisotropic covariance grades the response to a query (e.g., a camera ray) by how far it falls outside the ray group; and the record holds an opacity, a base radiance, and a directional term expressed with anisotropic spherical Gaussians (ASG)~\citep{xu2013anisotropic}, so the radiance can differ from ray to ray within the group.

\newcommand{\tpanel}[2]{%
  \begin{tikzpicture}[inner sep=0pt,outer sep=0pt]
    \node (img) {\includegraphics[width=\panelw,height=\panelh]{#1}};
    \ifx\relax#2\relax\else
      \node[anchor=south east,inner xsep=2pt,inner ysep=1.5pt,align=right,
            font=\fontsize{5}{5.5}\selectfont,text=white,
            fill=black,fill opacity=0.45,text opacity=1]
        at (img.south east) {#2};
    \fi
  \end{tikzpicture}%
}
\newcommand{\tnone}[1]{%
  \begin{tikzpicture}[inner sep=0pt,outer sep=0pt]
    \node[minimum width=\panelw,minimum height=\panelh,fill=black!5] (b) {};
    \node[font=\fontsize{5}{5.5}\selectfont,text=black!45,align=center] at (b.center) {#1};
  \end{tikzpicture}%
}

\newcommand{\rowlabel}[1]{\raisebox{25pt}{\rotatebox[origin=c]{90}{\scriptsize #1}}}
\newcommand{\rowlabelb}[1]{\raisebox{25pt}{\rotatebox[origin=c]{90}{\scriptsize\bfseries #1}}}

\begin{figure}[t]
\centering
\setlength{\tabcolsep}{0pt}
\renewcommand{\arraystretch}{0}
\begin{tabular}{@{}c@{\hspace{2pt}}c@{\hspace{1.5pt}}c@{\hspace{1.5pt}}c@{\hspace{1.5pt}}c@{\hspace{1.5pt}}c@{}}
& \footnotesize Novel views & \footnotesize Anti-aliasing & \footnotesize Deblurring
& \footnotesize Fisheye NVS & \footnotesize Refractive NVS \\[2pt]
\rowlabel{GT} &
\tpanel{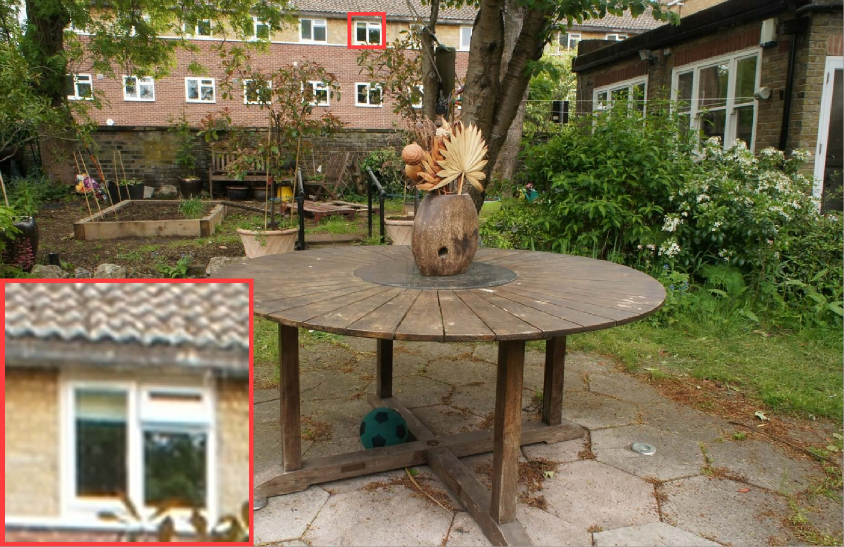}{} &
\tpanel{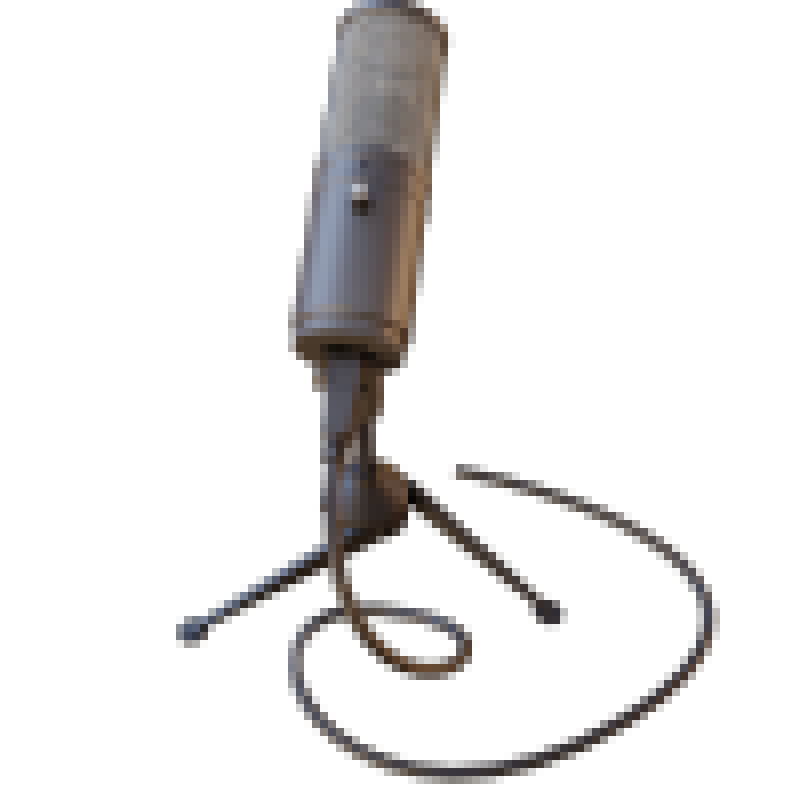}{} &
\tpanel{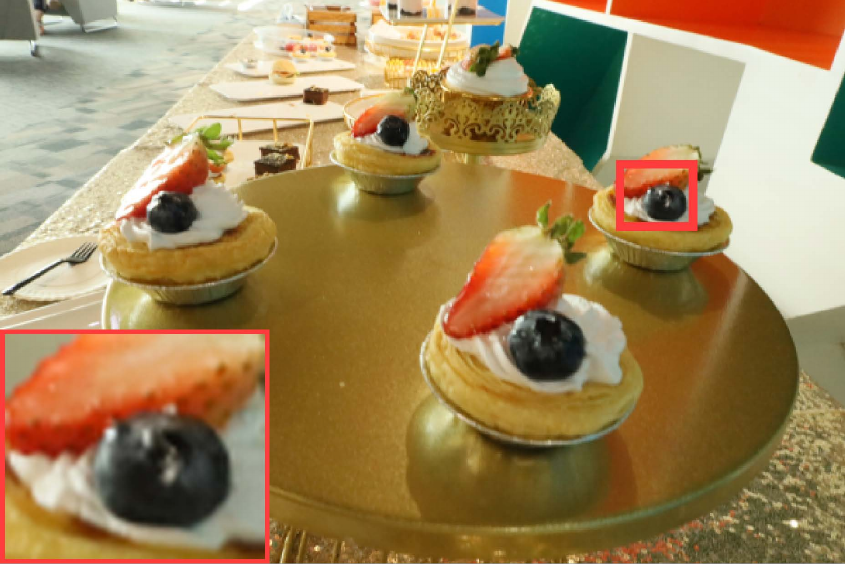}{} &
\tpanel{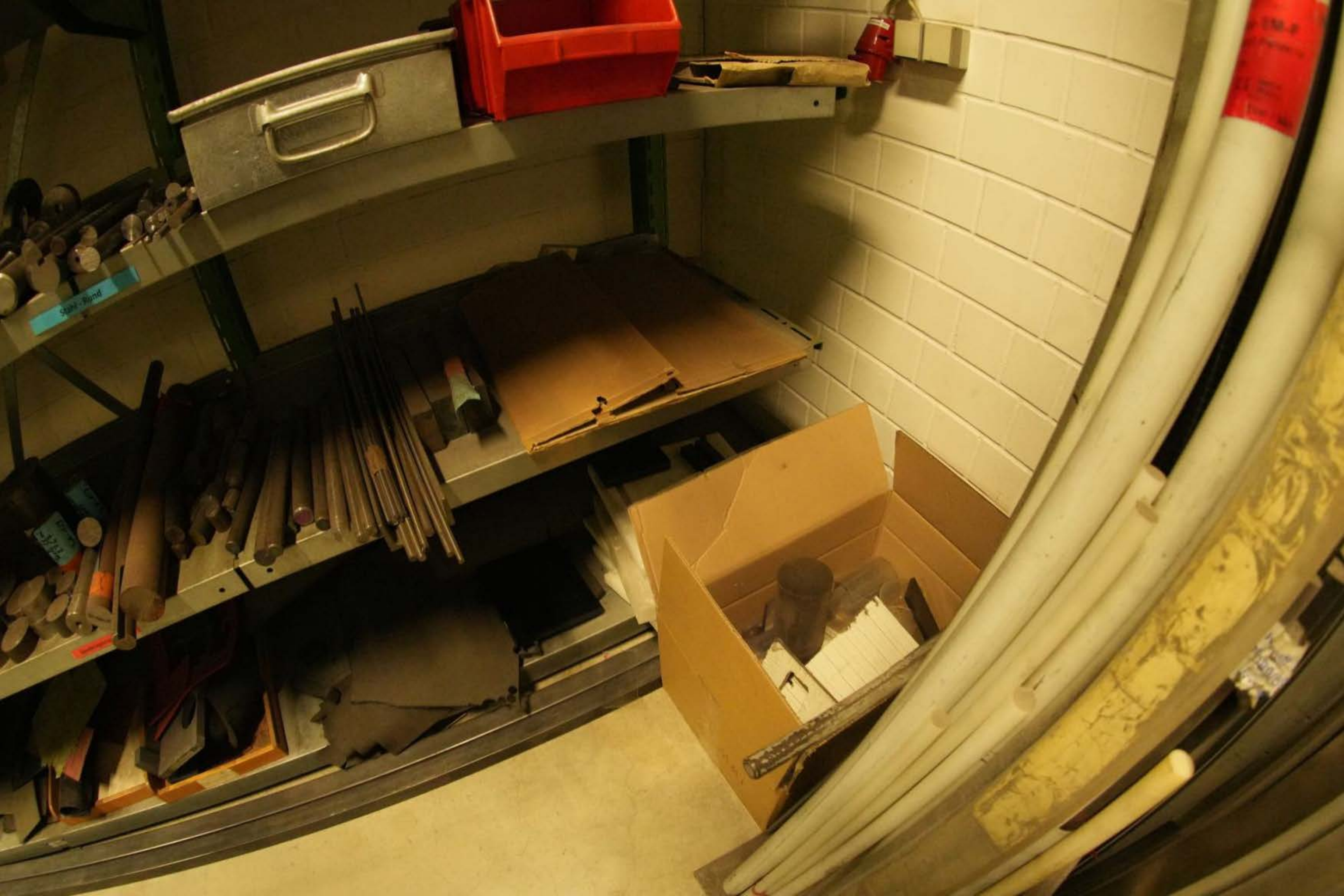}{} &
\tpanel{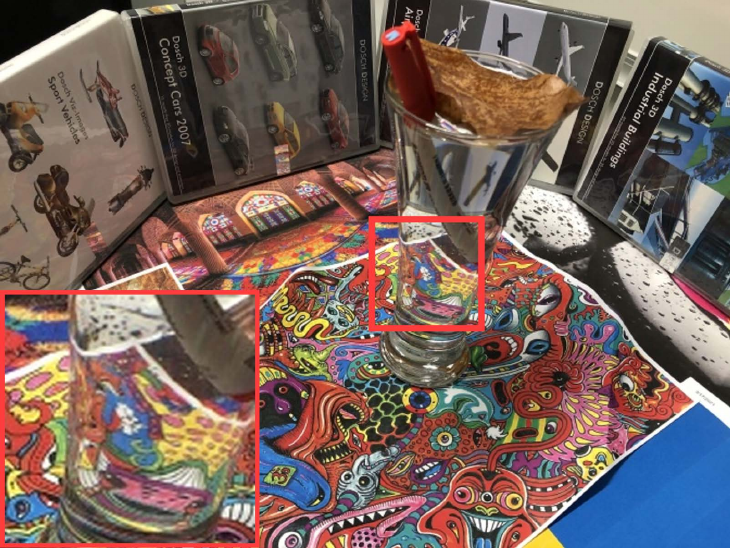}{} \\[1.5pt]  
\rowlabel{3D-anchored} &
\tpanel{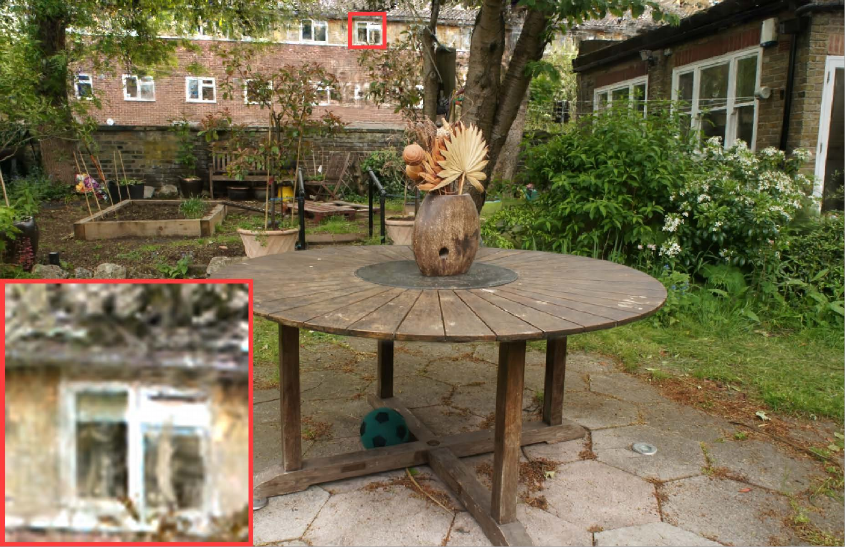}{3DGS} &
\tpanel{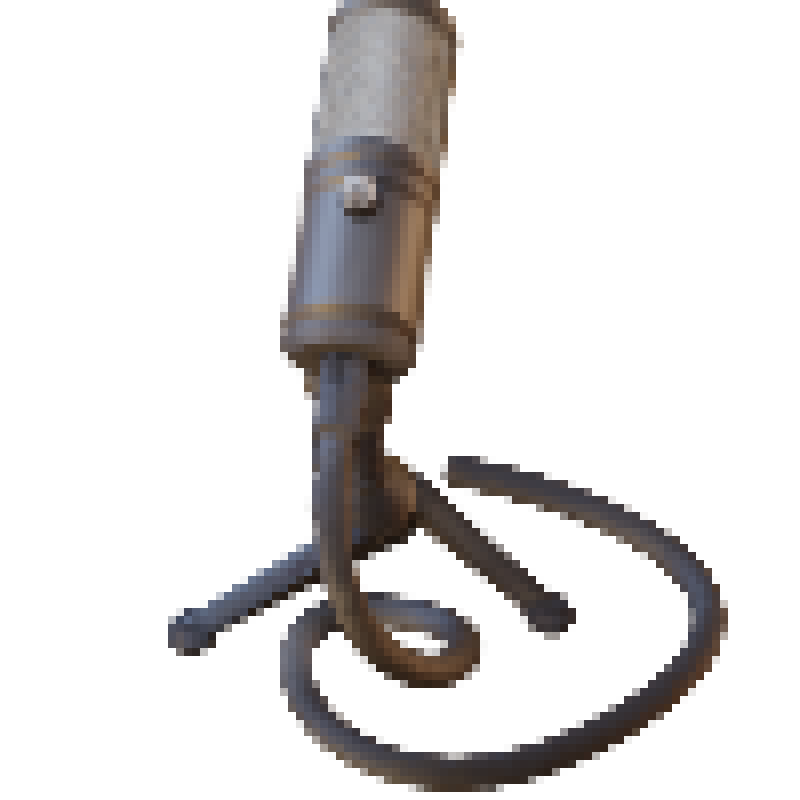}{3DGS} &
\tpanel{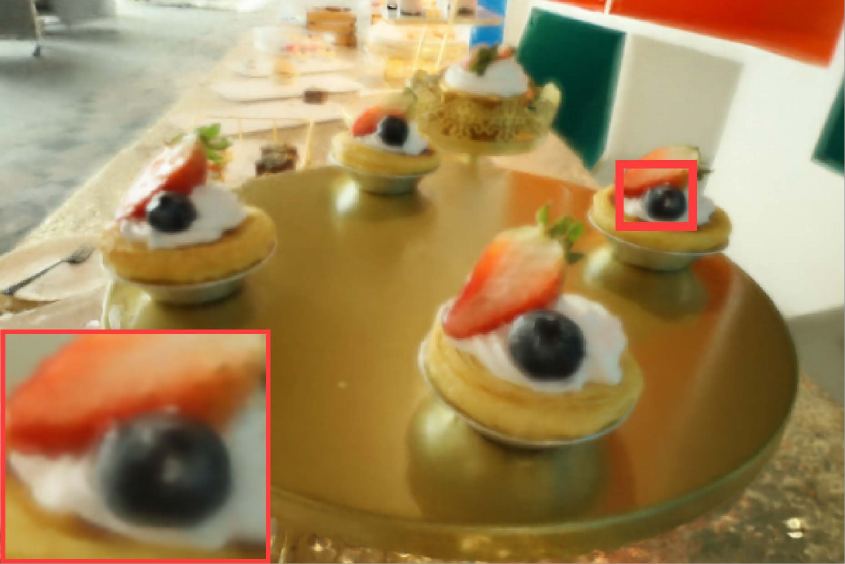}{NeRF} &
\tpanel{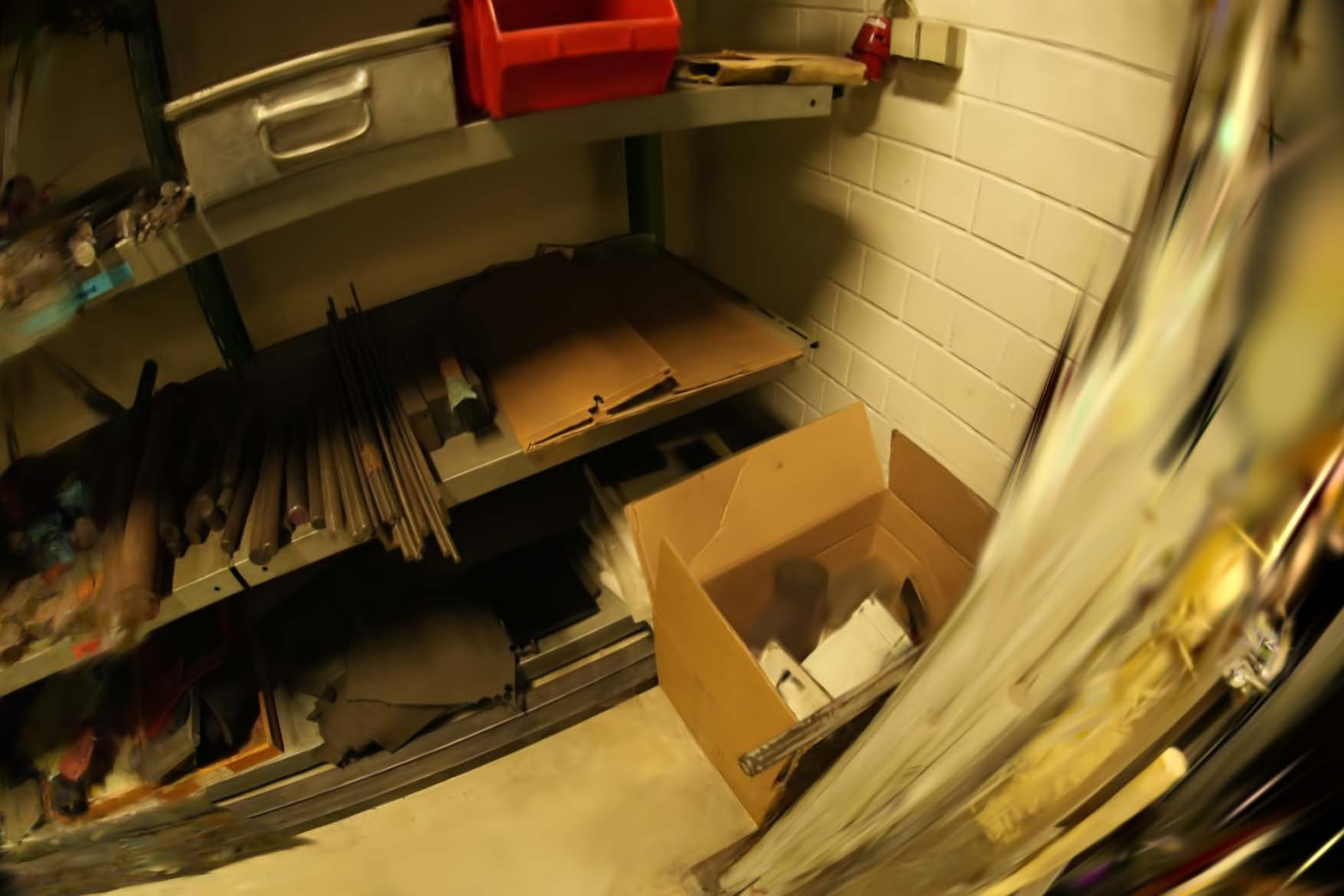}{3DGS} &
\tpanel{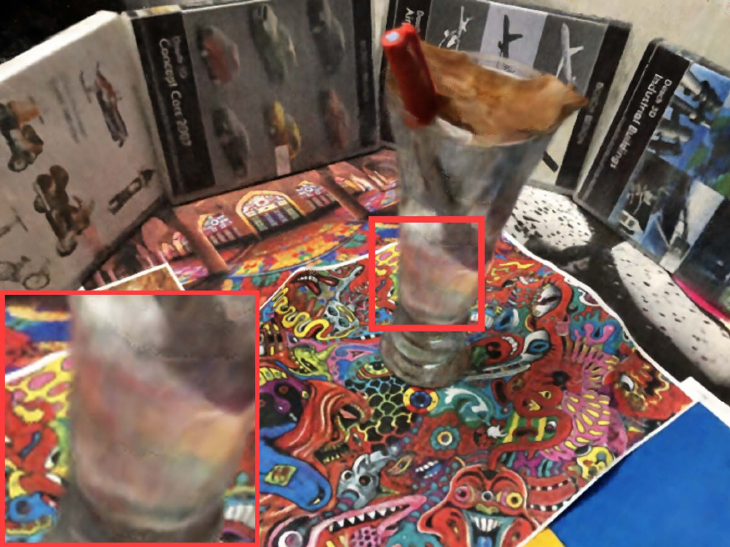}{NeRF}\\[1.5pt]  
\rowlabel{Specialized} &
\tnone{no method needed} &
\tpanel{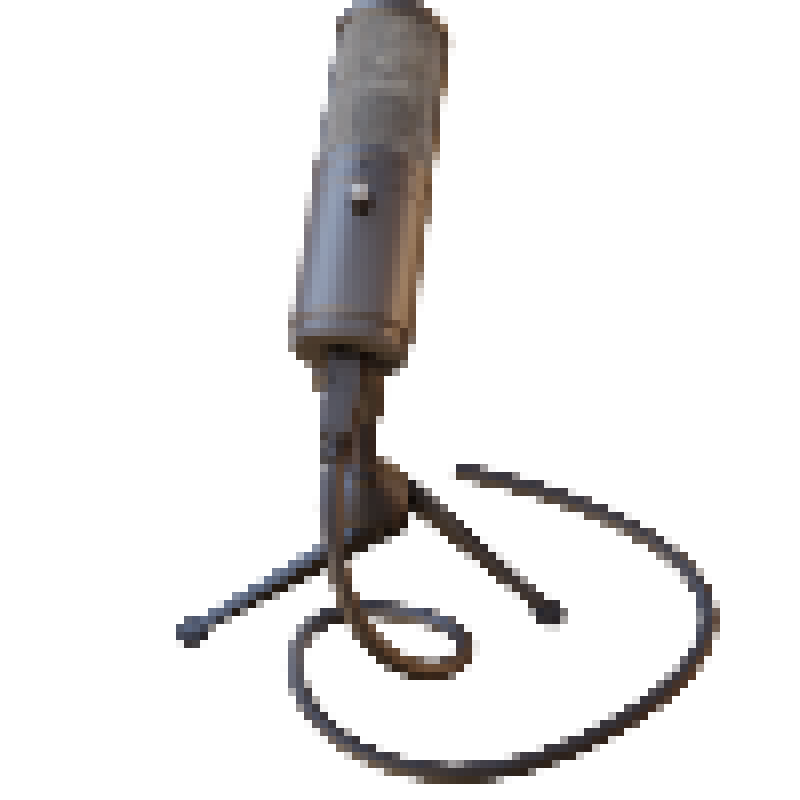}{Mip-Splatting\\filters + sample rate} &
\tpanel{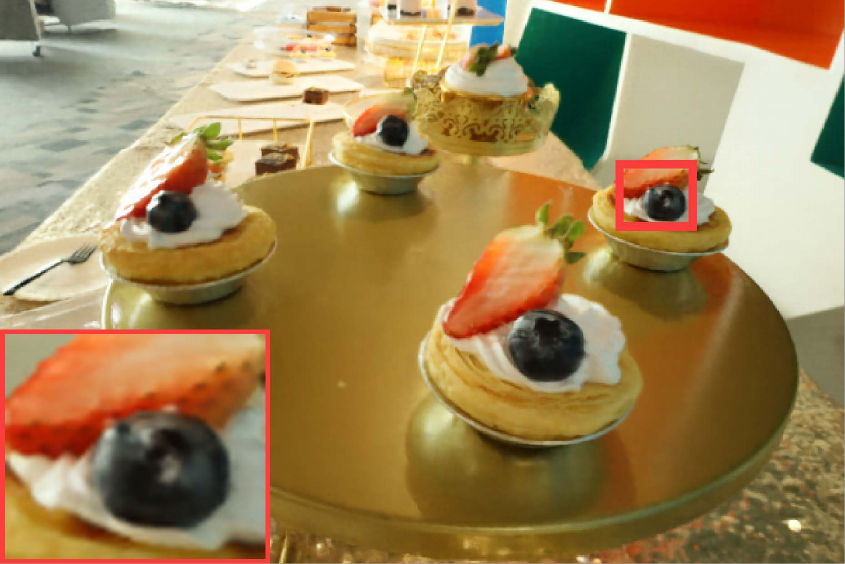}{Deblur-NeRF\\network + resample} &
\tpanel{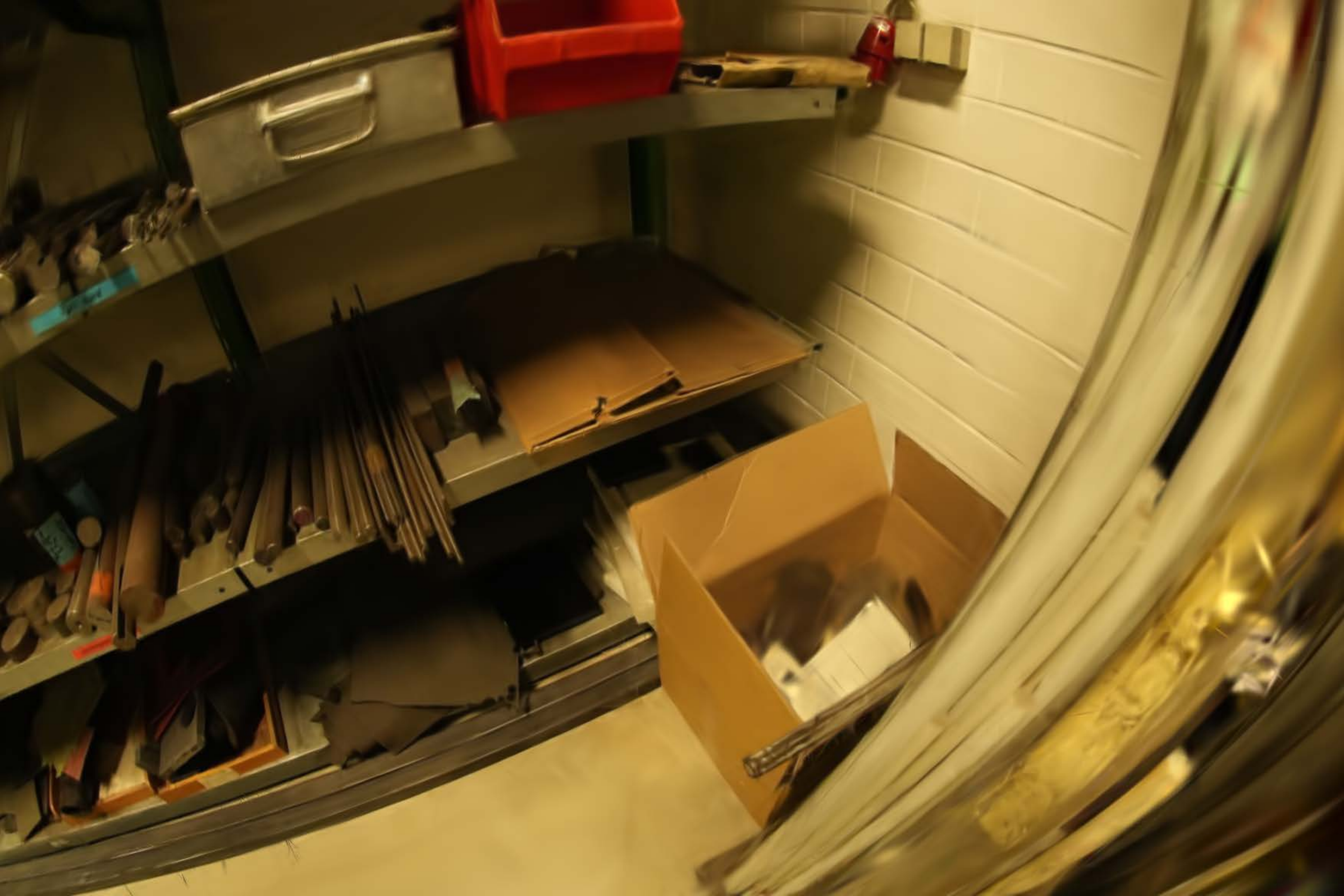}{3DGUT\\unscented rasterizer} &
\tpanel{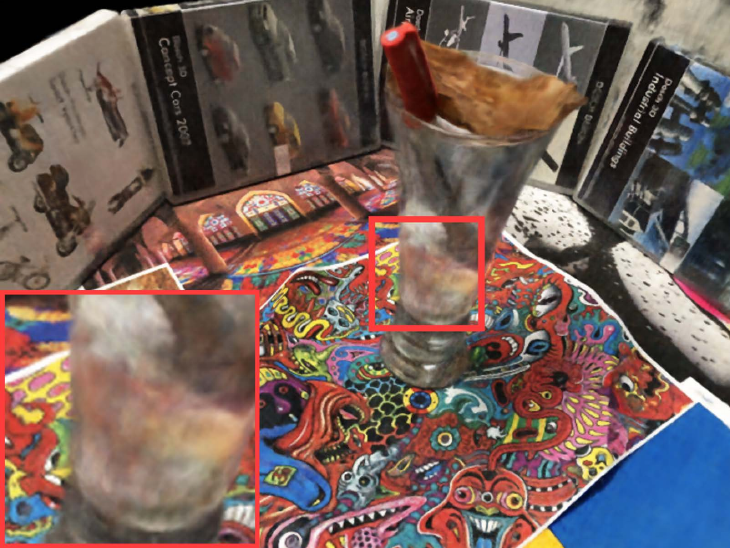}{Eikonal Fields\\refraction model} \\[1.5pt] 
\rowlabelb{LFP} &
\tpanel{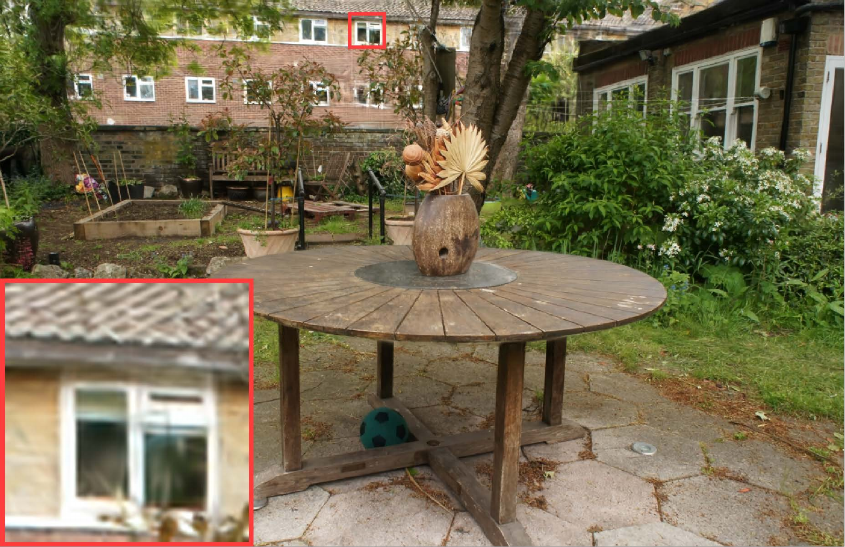}{\itshape nothing added} &
\tpanel{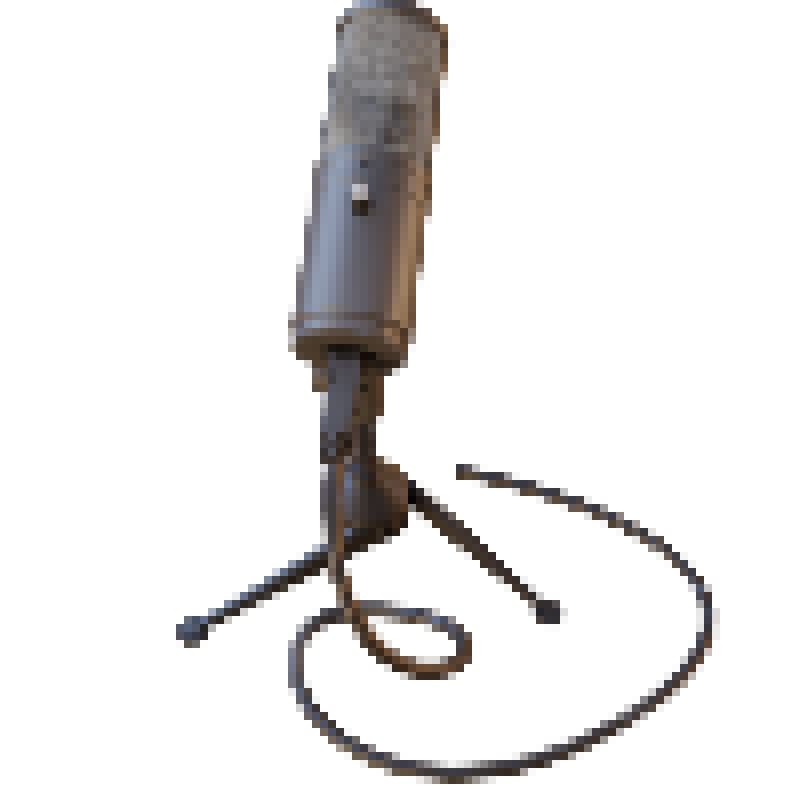}{\itshape one value changed} &
\tpanel{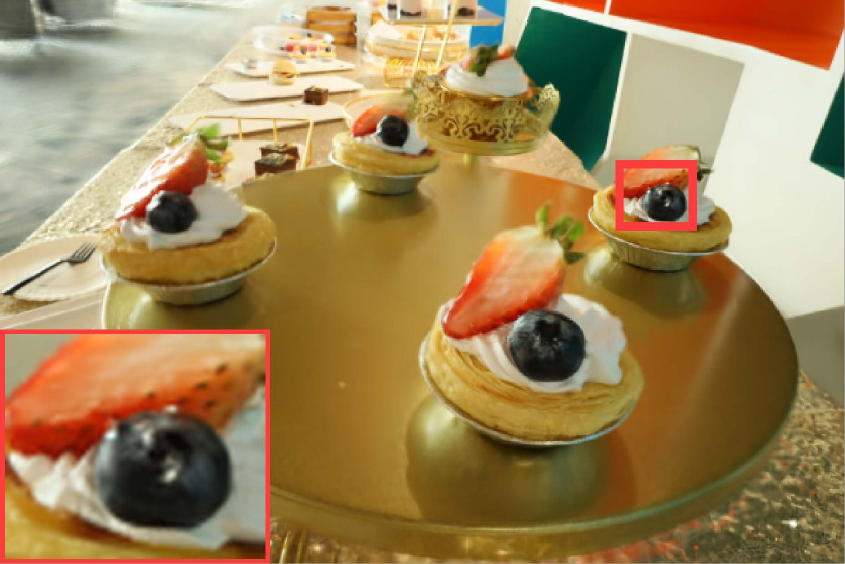}{\itshape two scalars per image} &
\tpanel{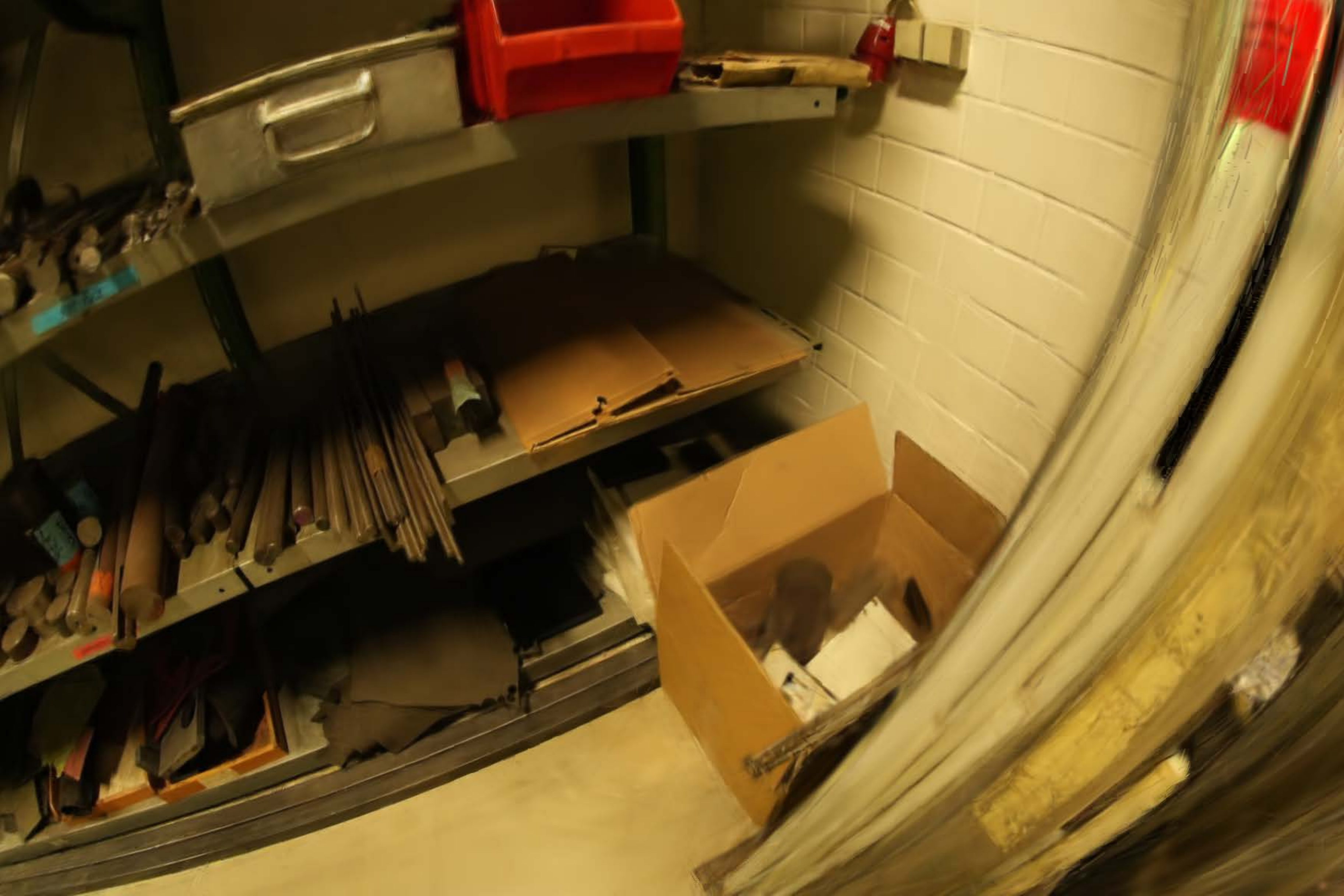}{\itshape fisheye pixel-to-ray  mapping} & 
\tpanel{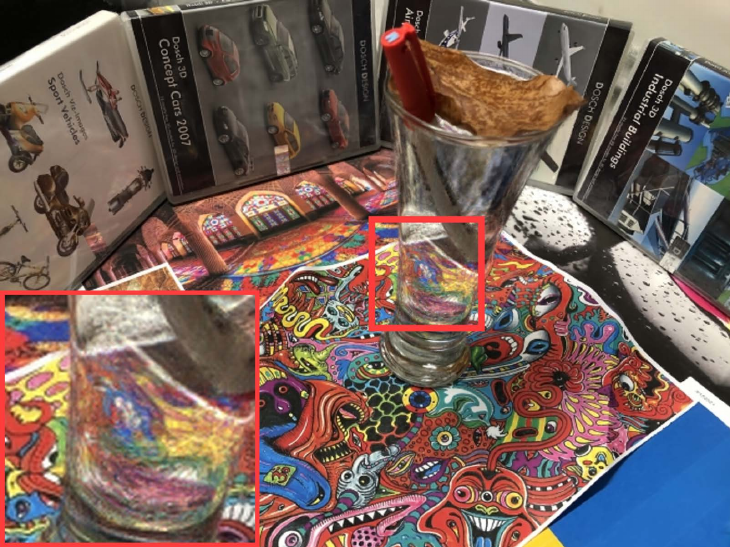}{\itshape nothing added}\\  
\end{tabular}
\vspace{-0.2cm}
\caption{\textbf{One renderer, five capabilities.} Panels in the ``Specialized" row are annotated with what each method adds over its base in the row above, and in ``LFP" with what LFP adds over its base.}
\vspace{-0.6cm}
\label{fig:tease}
\end{figure}

Meanwhile, since primitives are not parameterized by explicit 3D geometry, the projective splatting of 3DGS does not apply, requiring the interaction between a primitive and a query be established directly within the ray space. The interaction resides in a primitive's own parameters: its response to a query falls off as the query departs from the coupling that defines its ray group, measured under the covariance. Rendering then reduces to compositing all responses a query elicits, and it can be optimized from posed images and rendered in real time, retaining the optimizable nature of modern point-based methods while fully operating within the native coordinate system of light field.

As seen in Fig.~\ref{fig:tease}, beyond performing competitively on standard NVS benchmarks~\citep{barron2022mip,knapitsch2017tanks}, the advantage of LFP is structural: statements about rays are settled by the ray representation
itself without rebuilding the renderer.
How many rays a pixel gathers is one such statement, and both departures,
i.e., aliasing and defocus blur, ask only that a primitive be read over a
wider tolerance. A primitive already carries that tolerance, so multi-scale
anti-aliasing and defocus deblurring with post-hoc refocusing can be eased by simply
supplying different values (see Sec.~\ref{sec:aa_and_deblur}), at a competitive quality~\citep{yu2024mip,ma2022deblur}.
Which ray a pixel stands for is another, costing even less: a fisheye
capture changes how that ray is computed, but once computed it carries the
same coordinates as any other, so the renderer handles it exactly as in the ideal model, while the point-based family must rebuild its projection for such a len~\citep{wu20253dgut}.
A third statement is what a ray carries. The
grouping asks only that the radiance of a set of rays cohere, not what
produced it, so a ray bent by glass is recorded like any other, and
transparent objects can therefore be reconstructed without modeling the
refraction explicitly~\citep{bemana2022eikonal}.

Main contributions of this work are threefold:
\vspace{-0.2cm}
\begin{itemize}
\item We introduce LFP, an NVS method that brings the differentiable point-based optimization to the 4D ray space by modeling a scene as a set of primitives, each defined in the classical two-plane parameterization and holding one learned record for a group of rays.
\item We exploit the affine coupling that defines a ray group both to gather the rays sharing a primitive's record and to grade a query against them, giving a response function that can render in real time within the 4D ray space.
\item We demonstrate that LFP is effective on standard NVS benchmarks and, with minimal modifications to its renderer, can be extended to multi-scale anti-aliasing, defocus deblurring with post-hoc refocusing, fisheye rendering, and transparent-object reconstruction, at a quality comparable to frameworks specialized for each effect.
\end{itemize}

\section{Light field representation}

\label{sec:m-prim}

\begin{figure}[t]
  \centering
  \setlength{\tabcolsep}{4pt}
  \setlength{\dashlinedash}{5pt}
  \setlength{\dashlinegap}{2pt}
  \begin{tabular}{@{}c:c:c@{}}
    \includegraphics[width=0.29\textwidth]{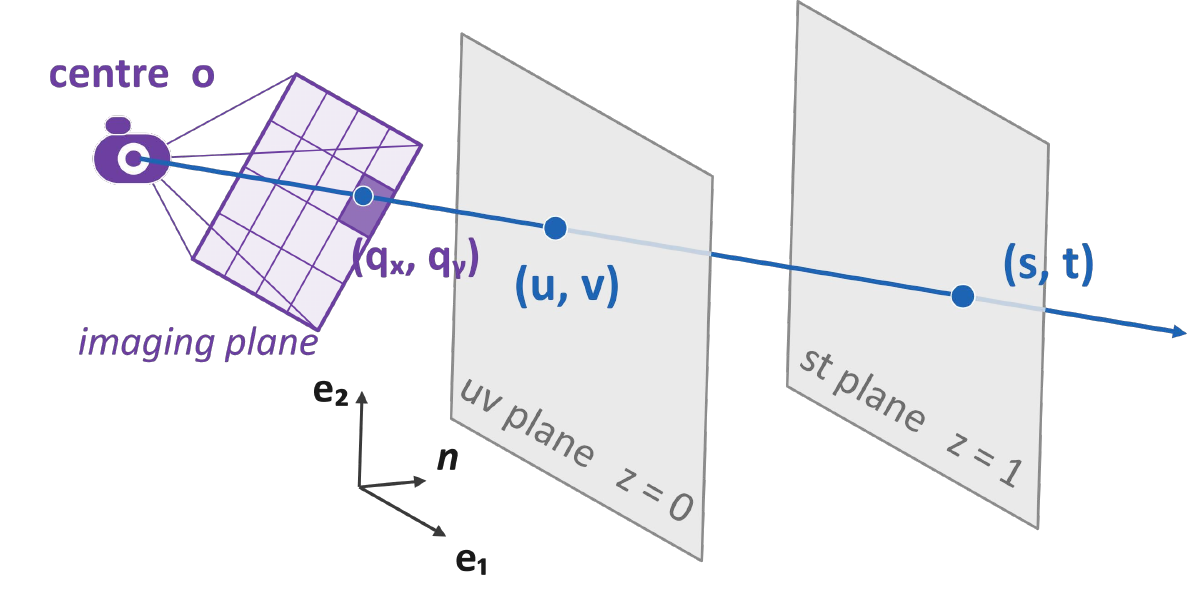} &
    \includegraphics[width=0.25\textwidth]{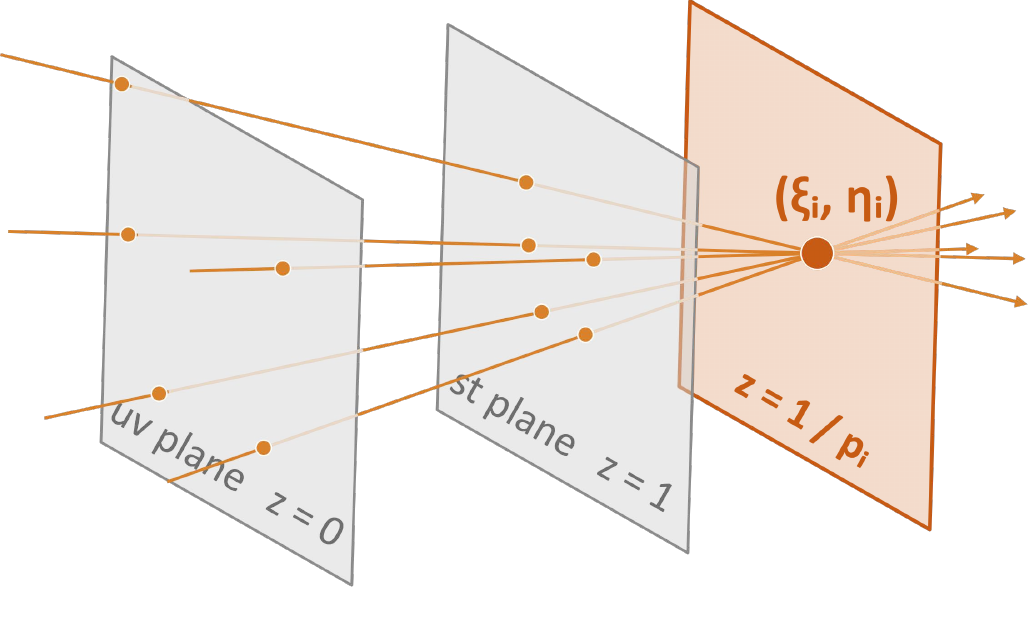} &
    \includegraphics[width=0.33\textwidth]{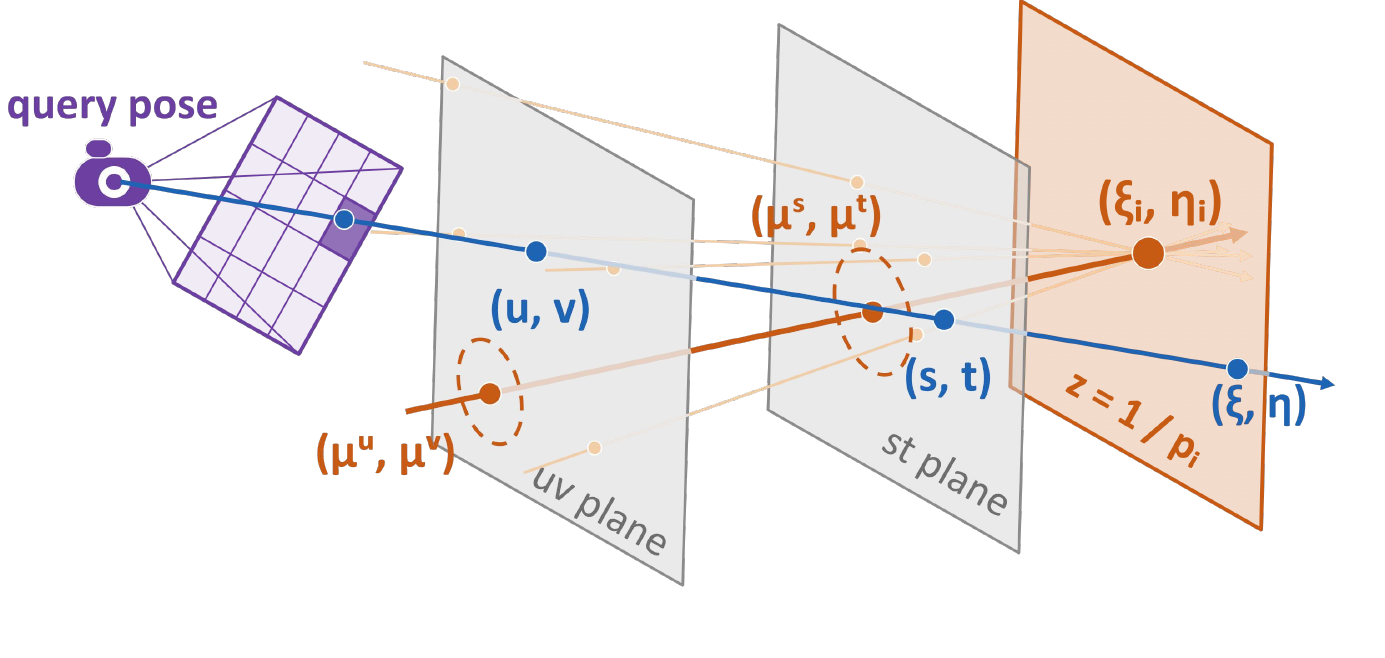} \\[-4pt]
    \multicolumn{1}{c}{(a)} & \multicolumn{1}{c}{(b)} & \multicolumn{1}{c}{(c)} \\
  \end{tabular}
  \vspace{-0.2cm}
  \caption{\textbf{Light field primitives for NVS.}
\textbf{(a)} Two parallel planes index a ray by its crossings at them.
\textbf{(b)} A ray group collects the rays that cross a third parallel  plane at one point.
\textbf{(c)} A primitive responds to a query (blue) by how far the query departs from its ray group, measured against the anchor ray it stores (thick orange), using their crossings at the initial two planes.}
  \label{fig:pipeline}
\end{figure}

\paragraph{Rays as light field coordinates.} 
As shown in Fig.~\ref{fig:pipeline} (a), we adopt the two-plane parameterization of classical light fields
\citep{levoy1996light}, where a unit normal $\mathbf{n}\in\mathbb{R}^3$ fixes a pair of parallel planes one unit apart: the $uv$ plane at $\mathbf{n}^\top\mathbf{y}=0$ and the $st$ plane at $\mathbf{n}^\top\mathbf{y}=1$. The planes share an orthonormal tangent basis $\mathbf e_1,\mathbf e_2\in\mathbb R^3$, with
$(\mathbf e_1,\mathbf e_2,\mathbf n)$ defining the coordinate system. A ray can then be indexed by the tangential coordinates of its two intersections with the two planes.
Concretely, for a camera with intrinsics $K\in\mathbb{R}^{3\times3}$ and rotation $R\in\mathbb{R}^{3\times3}$, the pixel at
image coordinates $(q_x,q_y)$ defines a ray that leaves the camera center $\mathbf{o}$ along the ray direction $\mathbf{d}=R K^{-1}(q_x,q_y,1)^{\top}$ and is mapped to
\begin{equation}
\label{eq:ray-coord}
\mathbf{x}=(u,v,s,t)=\bigl(\Pi(\mathbf{X}(\lambda_0)),\,\Pi(\mathbf{X}(\lambda_1))\bigr),
\qquad
\lambda_0=-\frac{\mathbf{n}^\top\mathbf{o}}{\mathbf{n}^\top\mathbf{d}},
\qquad
\lambda_1=\lambda_0+\frac{1}{\mathbf{n}^\top\mathbf{d}},
\end{equation}
with $\mathbf{x}\in\mathbb{R}^4$ the light field coordinate indexing the ray; $\mathbf{X}(\lambda)=\mathbf{o}+\lambda\mathbf{d}$ tracing the ray as
$\lambda$ advances from the camera, forward for $\lambda>0$ and backward for
$\lambda<0$; $\lambda_0$ and $\lambda_1$ the ray parameters at which it meets the
$uv$ and $st$ planes; and
$\Pi(\mathbf{y})=(\mathbf{e}_1^\top\mathbf{y},\,\mathbf{e}_2^\top\mathbf{y})$
extracting the tangential coordinates there. Every pixel can thus be mapped to a point in this ray space. Coordinates in that space are not all unrelated, and the structure that groups them, derived next, is what our primitives are defined on.

\paragraph{Epipolar geometry in ray space.}
Which rays should be represented by a shared record? They should be those
whose radiance is coherent enough to be captured jointly by a single
representation. Classical light field analysis~\citep{chai2000plenoptic}
shows that such coherence is organized by epipolar
geometry~\citep{bolles1987epipolar}: rays that see the same content are
related across views by a constraint set by the disparity of that content.
In two-plane coordinates this constraint takes a closed form. Consider a ray
with coordinates $(u,v,s,t)$ and a third parallel plane $\mathbf{n}^{\top}\mathbf{y}=z$, they intersect at
\begin{equation}
\label{eq:incidence}
(\xi,\eta)^\top=(1-z)(u,v)^\top+z(s,t)^\top,
\end{equation}
which reduces to $(u,v)$ at $z=0$ and to $(s,t)$ at $z=1$. \eqref{eq:incidence} holds for any $z$.
Now assuming specific content at disparity $p_i$, it sits on the plane with $z=1/p_i$, and every ray that sees it crosses that plane at the same point, whose tangential coordinate we write $(\xi_i,\eta_i)$, as depicted in Fig.~\ref{fig:pipeline}~(b).
Rearranging \eqref{eq:incidence} at that plane, they all satisfy
$(s,t)^\top=(1-p_i)(u,v)^\top+p_i(\xi_{i},\eta_{i})^\top$,
which is an affine coupling between the two plane coordinates, determined by
the disparity and the shared crossing through its coefficient $1-p_i$ and
offset $p_i(\xi_{i},\eta_{i})^\top$. This is the
epipolar constraint made explicit: with $v$ and $t$ held fixed it is a line
in the $(u,s)$ plane, the line an epipolar-plane
image~\citep{bolles1987epipolar} shows for content at that disparity, with
slope $1-p_i$. We call the rays satisfying one such coupling a ray group.
Given the disparity, any one of its rays recovers the offset through
\eqref{eq:incidence}, so a group is fully specified by its disparity and any
one of its rays.
Note that which disparity a group should take is not arbitrary: a light field
is coherent only along particular disparity, and grouping rays at any other
value mixes rays of unrelated radiance~\citep{wanner2012globally}.

\paragraph{Primitives.} 
LFP models a scene as a compact set of primitives, each attached to one ray
group defined above and holding the record that group shares. Primitive $i$ is specified by
\begin{equation}
\label{eq:primitive}
\mathcal{P}_i=\bigl\{\;\boldsymbol{\mu}_i,\; p_i,\; \Sigma_i,\; o_i,\; \mathbf{c}_i \;\bigr\},
\end{equation}
where the first two entries identify the group, the third shapes its response to a query, and the last two form the shared record. Here
$\boldsymbol{\mu}_i=(\mu^u_i,\mu^v_i,\mu^s_i,\mu^t_i)\in\mathbb{R}^4$ is the
coordinate of one ray of the group, which we used as an anchor for the ray group, with $(\mu^u_i,\mu^v_i)$ and
$(\mu^s_i,\mu^t_i)$ its intersections on the $uv$ and $st$ planes;
$p_i>0$ is the disparity of the group, setting the coefficient $1-p_i$ of
the affine coupling that holds among its rays,  tying their
$st$ coordinates to their $uv$ coordinates;
$\Sigma_i\in\mathbb{R}^{2\times2}$ is an anisotropic covariance, setting how
sharply the response falls off for queries outside the ray group;
$o_i\in(0,1)$ is the opacity of the record, modulating its peak contribution
during compositing;
$\mathbf{c}_i\in\mathbb{R}^3$ is its base radiance.
Sec.~\ref{sec:m-impl} details a directional term that absorbs the residual
variation within a group, while $\mathbf{c}_i$ carries the coherent part.
%

\newcommand{\vsep}{\raisebox{-2.4cm}[0pt][0pt]{\rule{0.4pt}{2.6cm}}}

\section{Method}
\subsection{Rendering}
\label{sec:m-render}

\paragraph{Response.}
Given a query $\mathbf{x}=(u,v,s,t)$, e.g., a camera ray traced from a pixel, the response it elicits from primitive $i$ is evaluated directly in ray space. Since the primitive and the query share the same light field coordinate system, no projection or screen-space linearization is required. Instead, we measure how far the query deviates from the ray group represented by the primitive.

At a fixed disparity $p$, the tangential coordinate at which a ray crosses the
plane $z=1/p$ partitions the ray space into distinct ray groups, and every
ray belongs to one such group. The deviation of $\mathbf{x}$ from a ray
group can thus be read off by comparing the differences between two groups, the one $\mathbf{x}$ falls in and the one the primitive represents, and we measure it under the partition at primitive $i$'s own disparity $p_i$,
through the residual
\begin{equation}
\label{eq:residual}
\mathbf{r}_i(\mathbf{x})=p_i\bigl((\xi,\eta)^\top-(\xi_i,\eta_i)^\top\bigr)
=(s-\mu^s_i, t-\mu^t_i)^\top 
-(1-p_i) (u-\mu^u_i, v-\mu^v_i)^\top,
\end{equation}
with $(\xi,\eta)$ and $(\xi_i,\eta_i)$ the tangential coordinates at which
$\mathbf{x}$ and $\boldsymbol{\mu}_i$ meet the plane $z=1/p_i$. The left form says what the residual is, which is the displacement between the two groups on that plane, scaled by $p_i$ so that it measures deviation in angular rather than in world units; the right form says how it is evaluated, from the coordinates of the query and the parameters the primitive stores, with no actual intersection computed, as displaced in Fig.~\ref{fig:pipeline} (c).
We convert the two-dimensional displacement into a scalar response using a Gaussian kernel:
\begin{equation}
\small
\label{eq:response}
G_i(\mathbf{x})=
\exp\!\Bigl(
-\tfrac12
\mathbf r_i(\mathbf{x})^\top
\Sigma_i^{-1}
\mathbf r_i(\mathbf{x})
\Bigr).
\end{equation}
This formulation assigns the maximum response to rays belonging to the
primitive's ray group and smoothly decreases with their deviation from the
encoded coupling, where $\Sigma_i$ controls the extent of the
response. 
Note that nothing constrains $\Sigma_i$ from below. If it is narrow along
some direction, an appreciable response is confined to queries very close
to the ray group in that direction, and once this range falls below the
pixel spacing of a view, the primitive can be missed by that view and
captured by another, which appears as flicker under a novel view. 
We therefore render with $\Sigma_i+\rho_i^2\mathbf{I}$,
where $\rho_i$ is the width of half a pixel, measured as a displacement of
the residual rather than of the image, so that
no response is read more narrowly than a view can resolve. The same guard appears in EWA splatting, where the reconstruction kernel is
convolved with a low-pass filter before it is
sampled~\citep{zwicker2002ewa}, and in 3DGS as a fixed dilation of the
projected covariance~\citep{kerbl20233d}. The conversion between the two widths is given in Sec.~\ref{sec:aa_and_deblur}.
The response from each primitive is then accumulated through compositing.

\paragraph{Compositing.}
The opacity $o_i$ caps primitive $i$'s contribution. A query receives a
discounted share,
\begin{equation}
\label{eq:alpha}
\alpha_i(\mathbf{x})=o_i\,G_i(\mathbf{x})\le o_i ,
\end{equation}
with equality when the query lies in the primitive's ray group. Many
primitives may respond to the same query, and their ordering affects the
result, as in standard front-to-back alpha
compositing~\citep{max1995optical}. Since the interaction with primitive $i$
takes place on the plane $z=1/p_i$ (\eqref{eq:residual} is a displacement on
that very plane), the query meets primitive $i$ at
$\lambda_i=\lambda_0+(\lambda_1-\lambda_0)/p_i$ by \eqref{eq:ray-coord}, in
which only $p_i$ varies between primitives, so the disparities alone fix the
sequence for each query, and a single sort by $p_i$ serves the entire view.
However, when a scene requires multiple light-field coordinate systems
(Sec.~\ref{sec:m-impl}), disparities along different normals are no longer
comparable, and the ordering must be resolved along a common axis. The camera
axis is the natural choice, but a plane $z=1/p_i$ does not have a single
depth along it; a specific site on the plane must be chosen. We take the crossing point where the response peaks, i.e., $(\xi_i,\eta_i)$ where the ray group is defined, which sits at $\xi_i\mathbf{e}_1+\eta_i\mathbf{e}_2+\mathbf{n}/p_i$ in the coordinate system
$(\mathbf{e}_1,\mathbf{e}_2,\mathbf{n})$, and read its depth in the camera
coordinate system. This depth is the sorting key, applied once per view, with
interactions behind the camera discarded. The query then accumulates
\begin{equation}
\label{eq:composite}
C=\sum\nolimits_{i=1}^{M} T_i\,\alpha_i\,\mathbf{c}_i
+T_{M+1}\,\mathbf{c}_{\mathrm{bg}},
\qquad
T_i=\prod\nolimits_{j<i}(1-\alpha_j),
\end{equation}
where $T_i$ is the fraction the first $i{-}1$ primitives leave to the rest,
$M$ the number of ordered primitives, and $\mathbf{c}_{\mathrm{bg}}$ the
background radiance receiving the remaining transmittance.

Note every step from \eqref{eq:residual} to \eqref{eq:composite} takes place in the
ray space where the query already lives, so LFP needs no mediation: neither
the ray marching of NeRF nor the projection of 3DGS. What a camera does to its rays, such as widening the
bundle a pixel gathers, spreading it across an aperture, or bending the map
from pixels to directions, therefore lands directly on quantities the renderer
already reads, which is what Sec.~\ref{sec:exp} relies on for anti-aliasing,
defocus deblurring, and fisheye NVS.

\subsection{Model Details and Optimization}
\label{sec:m-impl}

\paragraph{View-dependent appearance.} The base radiance $\mathbf{c}_i$ is constant across a ray group. To capture the variation, we use 3 ASG lobes~\citep{xu2013anisotropic} for each group, which is natural for modeling view-dependent appearance~\citep{yang2024spec}. Each lobe responds to a small set of directions, so a narrow highlight costs a single lobe, whereas the spherical harmonics of 3DGS spread over all directions and need many high-order terms for the same effect. The two bandwidths of a lobe are independent, letting it stretch along one tangent direction and stay narrow along the other.
The lobes are evaluated in a reflected direction, which requires an axis to
reflect about. We use the anchor $\boldsymbol{\mu}_i$ as that direction,
denoting it $\mathbf{m}_i$. Note nothing ties $\mathbf{m}_i$ to a surface, and the photometric loss is free to place it wherever it best explains how radiance varies across the group, so it acts as a shading normal rather than a geometric one. A unit query direction $\hat{\mathbf{d}}$ is mirrored about
$\mathbf{m}_i$ into
$\mathbf{h}_i=2(\mathbf{m}_i\!\cdot\!\hat{\mathbf{d}})\mathbf{m}_i
-\hat{\mathbf{d}}$. Following the standard ASG formulation, the actual radiance term becomes
\begin{equation}
\small
\label{eq:appearance}
\tilde{\mathbf{c}}_i(\hat{\mathbf{d}})=\sigma\Bigl(\mathbf{c}_i
+\sum\nolimits_{k}\mathbf{g}^k_i\,
\max\bigl(\mathbf{h}_i^\top\boldsymbol{\omega}^k_i,\,0\bigr)\,
\exp\bigl(-\lVert(\mathbf{B}^k_i)^\top\mathbf{h}_i\rVert^{2}\bigr)\Bigr),
\end{equation}
where lobe $k$ carries a unit axis $\boldsymbol{\omega}^k_i$, an RGB
amplitude $\mathbf{g}^k_i$, and a matrix $\mathbf{B}^k_i$ whose two columns
are its tangent axes scaled by the square roots of the bandwidths. All
of $\boldsymbol{\omega}^k_i$, $\mathbf{g}^k_i$, $\mathbf{B}^k_i$, and
$\mathbf{c}_i$ are optimized per primitive. Each lobe brightens
$\mathbf{c}_i$ over a narrow band of reflected directions.

\paragraph{Multi-atlas parameterization.}
A single plane pair cannot index rays that run nearly parallel to its
planes, as their light field coordinates diverge. We therefore fix a set of
atlases before optimization (16 atlas in practice), each with its own normal, tangent basis, and plane pair, with normals at the cluster centers of the training camera directions. A primitive keeps the atlas it is assigned, so the camera terms of \eqref{eq:ray-coord} are computed once per camera and atlas rather than once
per primitive.

\paragraph{Initialization.}
A ray group is specified by a plane parallel to the two parameterization
planes and a tangential coordinate on that plane (Sec.~\ref{sec:m-prim}). A
point in space naturally provides such a pair: it determines the plane
$\mathbf{n}^{\top}\mathbf{y}=z$ through its position along $\mathbf{n}$ and
the tangential coordinate $\Pi(\mathbf{y})$ on that plane. A point therefore
provides an initial identifier for a ray group, allowing a point cloud to
initialize a set of primitives. We use the sparse output of structure
from motion~\citep{snavely2006photo} for our intialization. Note the initial points are not part of the representation and serve only as starting values, different from that in 3DGS. 
Each point is assigned to an atlas according to the
alignment between a normal estimated from its neighbours and the atlas
normal. In that atlas, the point gives $1/p_i=\mathbf{n}^{\top}\mathbf{y}$
and $(\mu^u_i,\mu^v_i)=(\mu^s_i,\mu^t_i)=\Pi(\mathbf{y})$. The resulting
anchor ray runs along the atlas axis and therefore has identical tangential
coordinates on the two planes. This anchor direction is shared by all
primitives within the atlas and is not a per-point surface normal. The covariance is initialized isotropic, at a scale set from the local point
spacing, so that a primitive's response falls off no faster than the gap to
its neighbours. Since the group offset can be adjusted by
optimizing $(\mu^s_i,\mu^t_i)$, we fix $(\mu^u_i,\mu^v_i)$ after initialization
and optimize only $(\mu^s_i,\mu^t_i)$, $p_i$, $\Sigma_i$, and the record.

\paragraph{Optimization.}
Rendering follows the tile-based design in~\citep{kerbl20233d}: each primitive is assigned to the tiles it can affect, and each query composites the responses in its tile through \eqref{eq:composite}. While 3DGS reads this assignment off
the projected extent of a Gaussian, our primitives carry no explicit
geometry to project, so we obtain it from the response itself: evaluating
\eqref{eq:ray-coord} at neighbouring pixels gives the rate at which the
residual grows per pixel step, which converts $\Sigma_i$ into a screen
radius (see Sec.~\ref{sec:a-raster} for details). Every primitive is optimized against
\begin{equation}
\label{eq:loss}
\mathcal{L}=(1-\lambda_{\mathrm{d}})\,\mathcal{L}_{1}
+\lambda_{\mathrm{d}}\,\mathcal{L}_{\text{D-SSIM}}
+\lambda_{\mathrm{g}}\lVert\mathbf{g}\rVert^{2},
\end{equation}
where $\lambda_d$ and $\lambda_g$ are the weights, and the last term budgets the highlight amplitudes $\mathbf{g}$ of the
appearance model so that the directional lobes do not take over color that
belongs to the base.
We also use densification during our optimization: cloning, splitting, and pruning for primitives. The criterion for selecting which primitives to densify cannot follow 3DGS, which uses the gradient w.r.t a projected 2D mean, since no such quantity enters our response. We accumulate instead the gradients from the parameters that define a ray group, i.e., $\boldsymbol{\mu}_i$ and $p_i$, as the densification signal.

\begin{table}[t]
\centering
\small
\caption{Average NVS results. Numbers are cited from~\citep{kulhanek2026nerfbaselines} except those boxed with dash lines, which we ran ourselves. FPS is measured on our device for every method.}
\vspace{-0.3cm}
\label{tab:nvs}
\setlength{\tabcolsep}{2pt}
\scalebox{0.85}{
\begin{tabular}{llcc*{10}{c}}
\toprule
& & & & \multicolumn{5}{c}{Mip-NeRF 360 (9 scenes)} & \multicolumn{5}{c}{Tanks and Temples (21 scenes)} \\
\cmidrule(lr){5-9}\cmidrule(lr){10-14}
& Method & Space & Mediation
& PSNR $\uparrow$ & SSIM $\uparrow$ & LPIPS $\downarrow$ & FPS & Mem
& PSNR $\uparrow$ & SSIM $\uparrow$ & LPIPS $\downarrow$ & FPS & Mem \\
\midrule
\multirow{3}{*}{\rotatebox[origin=c]{90}{NeRF}}
& Instant NGP & \multirow{3}{*}{3D field} & \multirow{3}{*}{ray-marching}
& 25.51 & 0.684 & 0.398 & 1.33 & 38.1MB & 21.62 & 0.712 & 0.340 & 1.62 & 39.3MB \\
& NeRFStudio & & &
26.39 & 0.731 & 0.343 & 1.72 & 68.4MB & 22.04 & 0.743 & 0.270 & 3.88 & 68.4MB \\
& M-NeRF 360 & & &
\snd{27.68} & 0.792 & 0.272 & 0.12 & 34.4MB
& \tikzmark{m1}23.43 & 0.784 & 0.215\tikzmark{m2} & 0.27 & 34.4MB \\
\midrule
\multirow{3}{*}{\rotatebox[origin=c]{90}{GS}}
& M-Splatting & \multirow{3}{*}{3D point} & \multirow{3}{*}{projection}
& \trd{27.49} & \snd{0.815} & 0.258 & 149 & 782MB
& \trd{23.93} & \snd{0.833} & 0.166 & 203 & 452MB \\
& 3DGS & & &
\tikzmark{a1}27.45 & \trd{0.814} & \trd{0.257} & 166 & 789MB
& \tikzmark{b1}23.84 & \trd{0.831} & \trd{0.165} & 229 & 418MB \\
& 3DCS & & &
27.22 & 0.800 & \best{0.244} & 39.1 & 670MB
& \snd{24.26} & 0.828 & \best{0.137} & 36.6 & 515MB \\
\midrule
& LFP & 4D ray & unnecessary
& \best{28.02} & \best{0.820} & \snd{0.247}\tikzmark{a2} & 101 & 313MB
& \best{24.38} & \best{0.835} & \snd{0.151}\tikzmark{b2} & 173 & 150MB \\
\bottomrule
\end{tabular}
\dbx{m1}{m2}\dbx{a1}{a2}\dbx{b1}{b2}}
\vspace{-0.2cm}
\end{table}

\begin{figure}[t]
  \centering
  \setlength{\tabcolsep}{1pt}
  \begin{tabular}{@{}c@{\hspace{1pt}}c@{\hspace{1pt}}c@{\hspace{1pt}}c@{\hspace{1pt}}c@{\hspace{1pt}}c@{}}
    \includegraphics[width=0.16\textwidth]{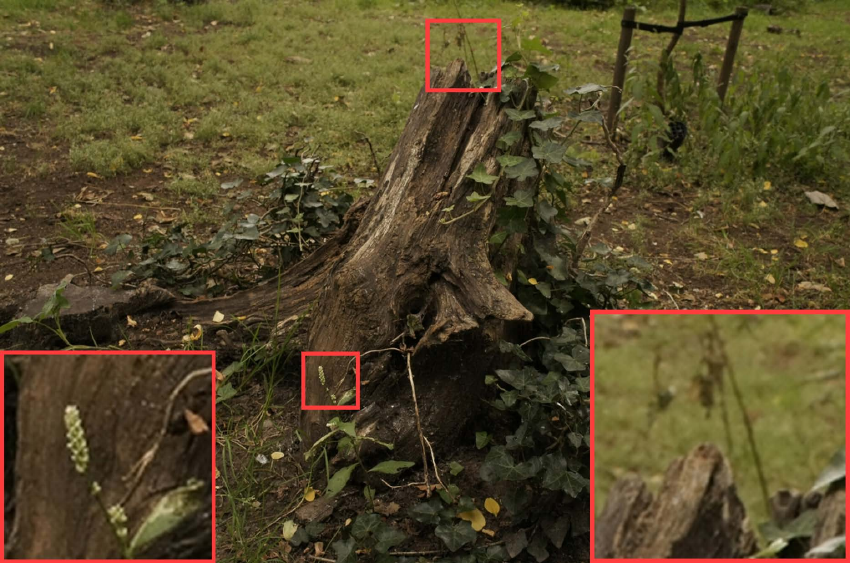} &
    \includegraphics[width=0.16\textwidth]{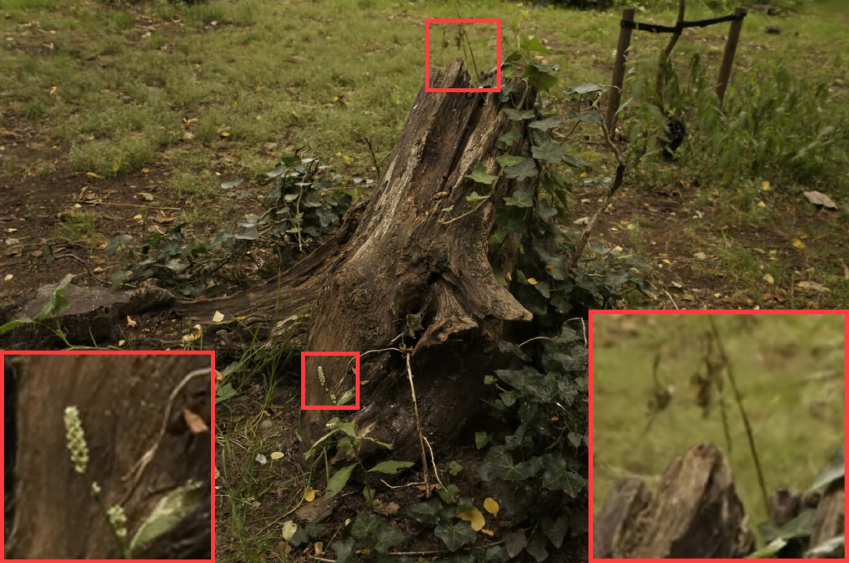} &
    \includegraphics[width=0.16\textwidth]{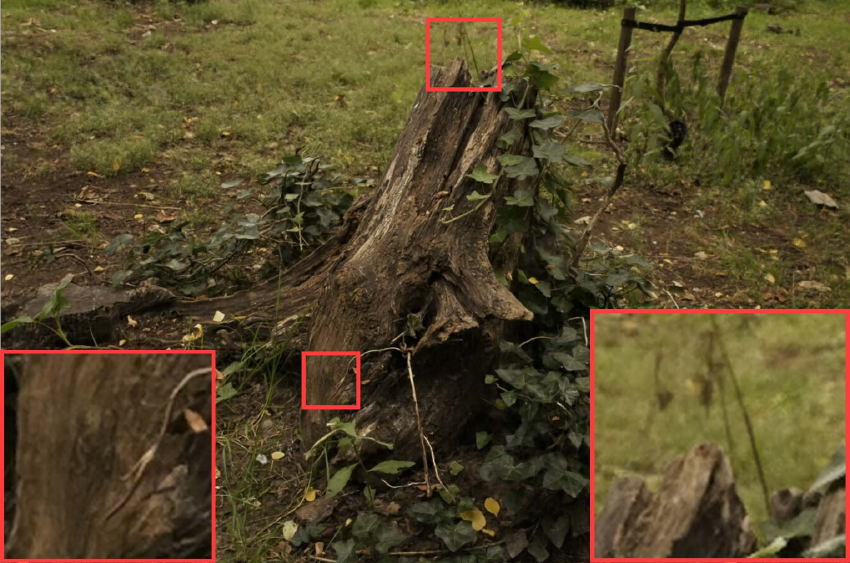} &
    \includegraphics[width=0.16\textwidth]{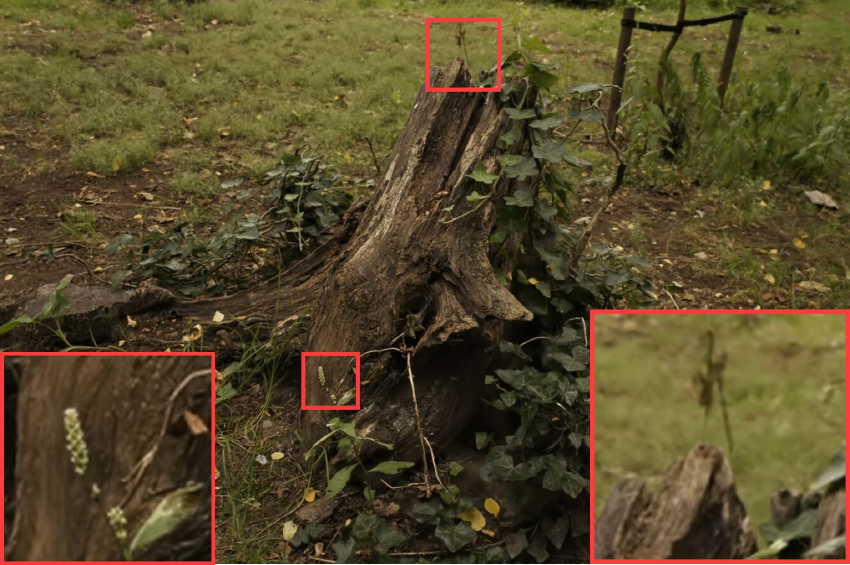} &
    \includegraphics[width=0.16\textwidth]{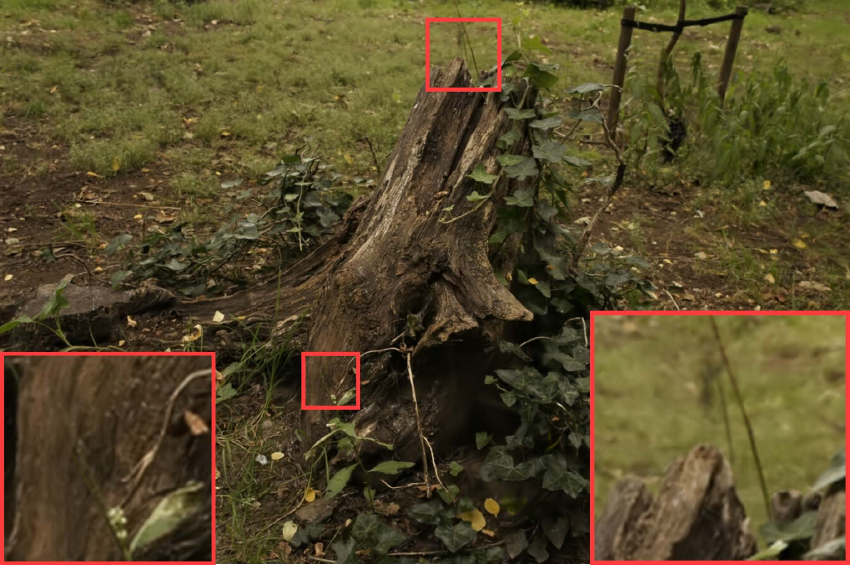} &
    \includegraphics[width=0.16\textwidth]{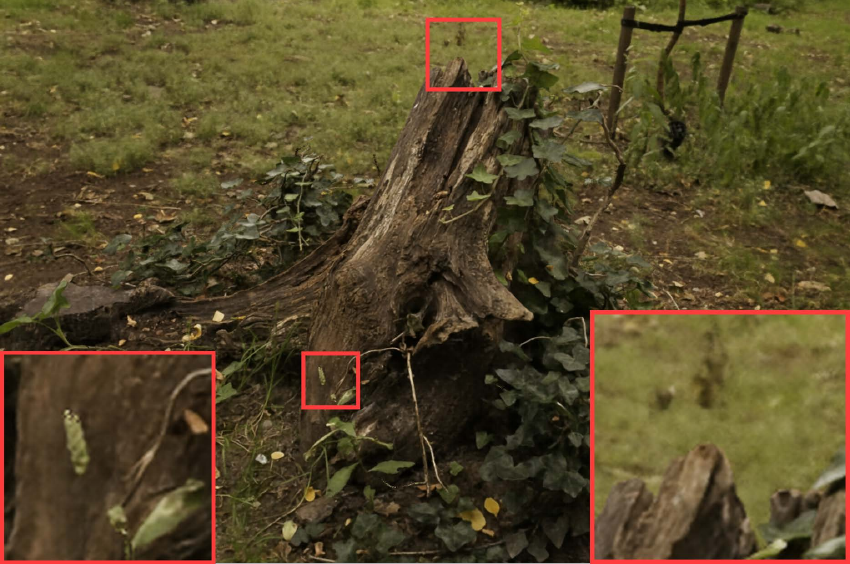} \\
    \includegraphics[width=0.16\textwidth]{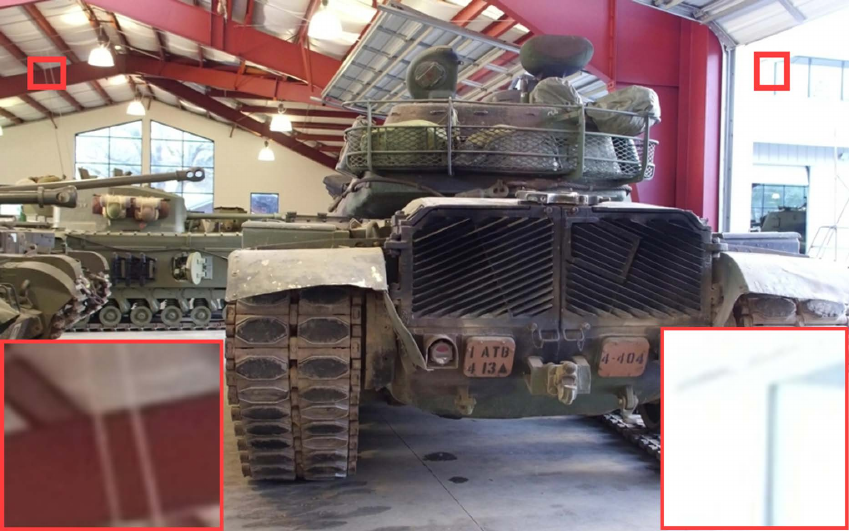} &
    \includegraphics[width=0.16\textwidth]{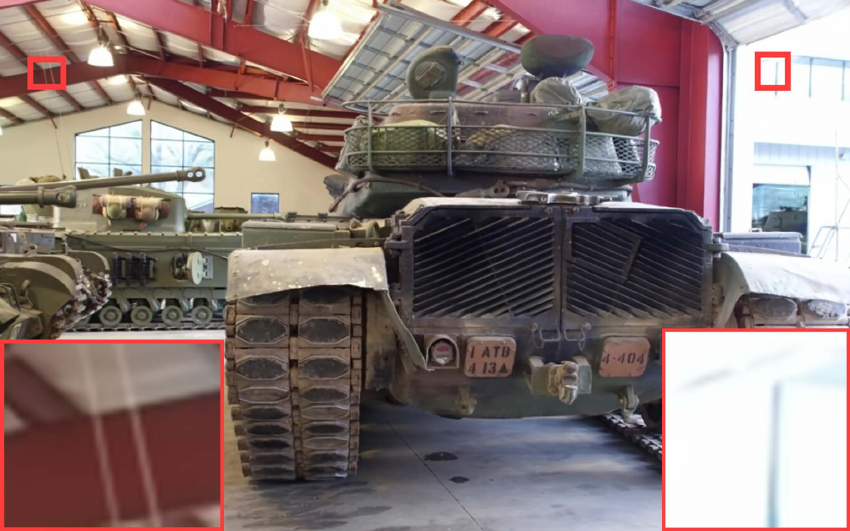} &
    \includegraphics[width=0.16\textwidth]{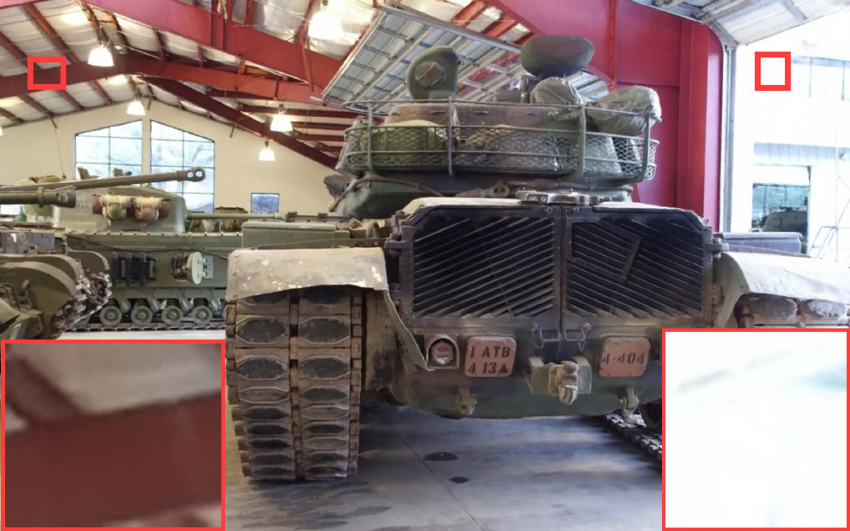} &
    \includegraphics[width=0.16\textwidth]{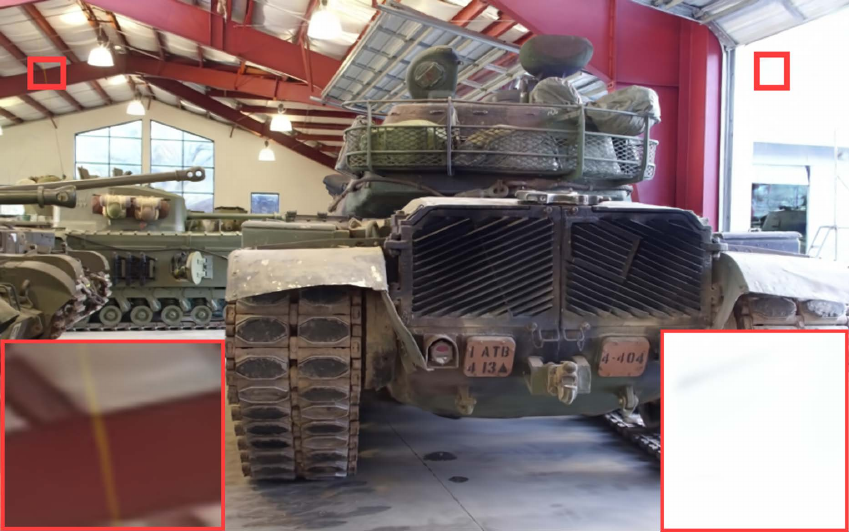} &
    \includegraphics[width=0.16\textwidth]{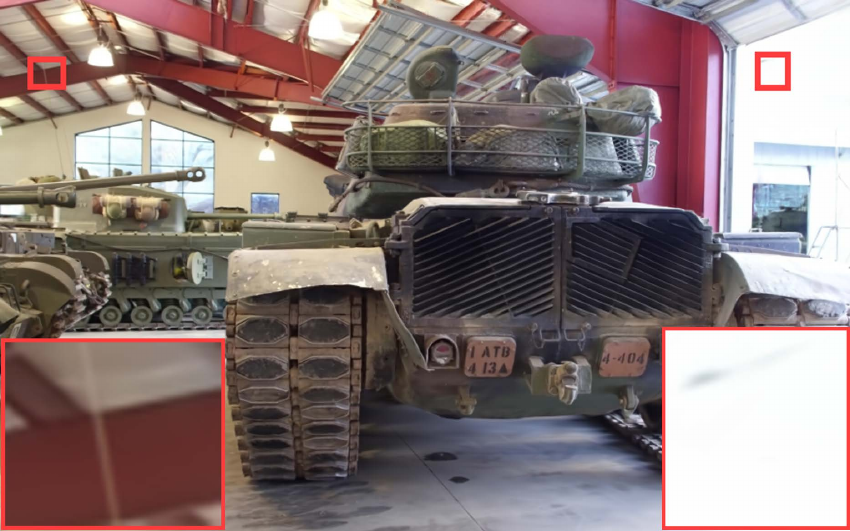} &
    \includegraphics[width=0.16\textwidth]{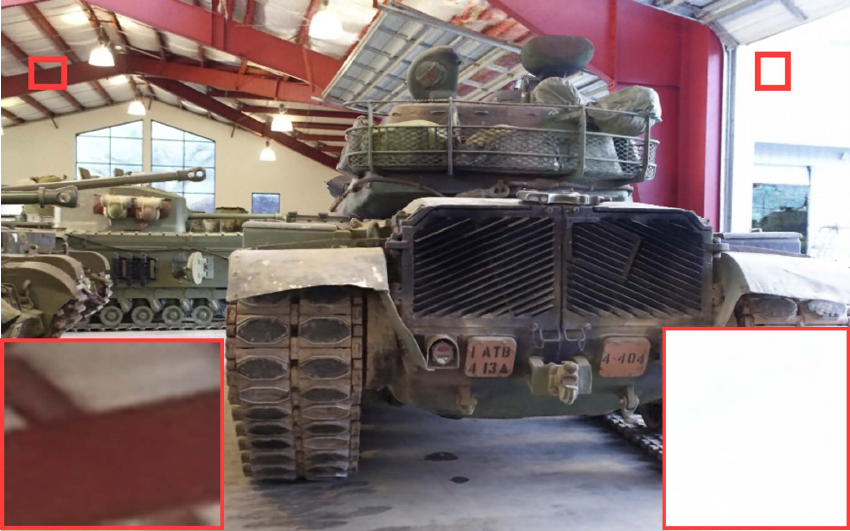} \\
    \scriptsize{(a) GT} & \scriptsize{(b) LFP} & \scriptsize{(c) 3DCS} & \scriptsize{(d) 3DGS} & \scriptsize{(e) Mip-Splatting} & \scriptsize{(f) Mip-NeRF 360}
  \end{tabular}
  \vspace{-0.2cm}
  \caption{NVS examples on \citet{barron2022mip} (\emph{top}) and \citet{knapitsch2017tanks} (\emph{bottom}).}
  \label{fig:nvs_sample}
  \vspace{-0.5cm}
\end{figure}

\section{Experiments}
\label{sec:exp}

\paragraph{Datasets and Implementation.} We evaluate LFP on 5 distinct setups: standard NVS on the 9 Mip-NeRF 360 scenes~\citep{barron2022mip} and 21 Tanks and Temples scenes~\citep{knapitsch2017tanks}, following protocol in~\citet{kulhanek2026nerfbaselines}; multi-scale anti-aliasing under the single-scale training and multi-scale testing protocol of~\citet{yu2024mip}, with zoom-in on the 9 Mip-NeRF 360 scenes and zoom-out on the 8 synthetic scenes of Blender~\citep{mildenhall2021nerf}; defocus deblurring on the 10 real defocus scenes of Deblur-NeRF~\citep{ma2022deblur}; fisheye NVS on the selected 6 scenes from ScaneNet++~\citep{yeshwanth2023scannet++} following \citet{liao2024fisheye}, and transparent-object reconstruction on the 4 glass and liquid scenes from~\citep{bemana2022eikonal}. 

LFP is built in PyTorch with the renderer as a single CUDA kernel. The same kernel serves all setups, including the two covariance edits in anti-aliasing and deblur that are applied in the Python layer before the kernel is called, so anti-aliasing and defocus reuse the unchanged base renderer (see Sec.~\ref{sec:aa_and_deblur} for details). The loss weights are the same in every experiment, $\lambda_{\mathrm{d}}=0.05$ and $\lambda_{\mathrm{g}}=0.005$, and each scene trains for $30$k iterations, evaluated at the last one, except the defocus setup, which trains for $20$k. Within a setup no setting is tuned per scene, and every setup is initialized from COLMAP points unless otherwise mentioned. All experiments run on a single H100 GPU. 

\paragraph{Standard NVS results.}
Tab.~\ref{tab:nvs} shows results against representatives of both
families~\citep{mueller2022instant,barron2022mip,tancik2023nerfstudio,yu2024mip,kerbl20233d,held20253d}. Figures are taken from~\citet{kulhanek2026nerfbaselines}, which evaluates every method under one protocol, except for the boxed rows, which we ran ourselves: the two point-based baselines, and Mip-NeRF~360 on Tanks and Temples, which the benchmark does not report. Our reevaluated 3DGS run lands within 0.02\,dB of the cited one, indicating the two protocols agree.
LFP is first on PSNR and SSIM on both datasets and second on LPIPS, performing competitively against many state-of-the-art methods. 
Quality aside, the two families differ in what they cost to render. The NeRF
family stays below two frames per second, so the comparison to make is with
the point-based methods, which rasterize as we do and differ in the space
their primitives are defined in. Against 3DGS, LFP renders at about two thirds
of the frame rate, still in real time, and holds under half the memory, and
its variants make the same trade.
Fig.~\ref{fig:nvs_sample} shows one scene from each dataset, where LFP
preserves fine structures and image details while remaining visually
comparable to current baselines.

\emph{What the ray-space parameterization buys is not in these numbers.} Both
NeRF and 3DGS families reach a primitive through a step the renderer must take before anything can be compared, marching the ray through the field or projecting each primitive to the image plane, and LFP needs neither: a query and a primitive are already written in the same coordinates, and their interaction is a subtraction. The capabilities this admits are the subject of the remaining experiments.

\paragraph{Multi-scale anti-aliasing.} In this setting, a
scene is trained at one resolution and rendered at others. Two directions are tested separately. A scene is trained once and rendered at $n\in\{1,2,4,8\}$ times the training resolution for zoom-in, and at $1/n$ of it for zoom-out. In both directions the same trained scene serves every output resolution, and the covariance and opacity edits in Sec.~\ref{sec:supp-aa} are applied at render time with no gradient step.
Results are in Tab.~\ref{tab:aa}. At the training scale all methods are
close, and the differences appear once the rendering resolution departs from
it. 3DGS degrades sharply in both directions, since its screen-space dilation is tied to neither the sampling rate nor the primitive. The EWA filter recovers that loss but oversmooths, and Mip-NeRF stays stable across scales at a lower level throughout. LFP holds up in both directions and is close to the best Mip-Splatting.

\emph{However, what it takes to achieve the performance on either side is not
comparable.} LFP changes only the value of a term it already carries and
rescales the opacity once, with no parameter added and nothing recomputed in
training. Mip-Splatting needs two filters, one in world space and one in
screen space, together with a per-primitive sampling rate recovered from all
training cameras and maintained as primitives move and split, because a 3D
Gaussian carries the same covariance whichever camera looks at it. The
half-pixel width $\rho_i$ of Sec.~\ref{sec:m-render} already depends on the
group's disparity and the camera being rendered, so each view supplies its own
and neither filter is needed. Without either, LFP is competitive with
Mip-Splatting across both directions, ahead on PSNR at the scales furthest
from training and second on SSIM and LPIPS.
In all, without any of these complexities, LFP is still competitive with
Mip-Splatting, showing the potential of this new pipeline.

\begin{table}[t]
\centering
\small
\caption{Multi-scale rendering. A scene is trained once, at $n{=}1$, and rendered at four scales: zoom-in renders at $n\times$ the training resolution on Mip-NeRF 360, zoom-out at $1/n\times$ on Blender. All numbers are cited directly from~\citet{yu2024mip}, whose protocol we follow for LFP.}
\vspace{-0.3cm}
\label{tab:aa}
\setlength{\tabcolsep}{2.5pt}
\scalebox{0.88}{
\begin{tabular}{ll|*{5}{c}|*{5}{c}|*{5}{c}}
\toprule
& & \multicolumn{5}{c|}{PSNR $\uparrow$} & \multicolumn{5}{c|}{SSIM $\uparrow$} & \multicolumn{5}{c}{LPIPS $\downarrow$} \\
& Method
& $n{=}1$ & $n{=}2$ & $n{=}4$ & $n{=}8$ & Avg.
& $n{=}1$ & $n{=}2$ & $n{=}4$ & $n{=}8$ & Avg.
& $n{=}1$ & $n{=}2$ & $n{=}4$ & $n{=}8$ & Avg. \\
\midrule
\multirow{5}{*}{\rotatebox[origin=c]{90}{Zoom-in}}
& Mip-NeRF      & 29.26 & 25.18 & 24.16 & 24.10 & 25.67 & 0.860 & 0.727 & 0.670 & 0.706 & 0.741 & 0.122 & 0.260 & 0.370 & 0.428 & 0.295 \\
& 3DGS          & 29.19 & 23.50 & 20.71 & 19.59 & 23.25 & \snd{0.880} & 0.740 & 0.619 & 0.619 & 0.715 & \snd{0.107} & 0.243 & 0.394 & 0.476 & 0.305 \\
& 3DGS + EWA    & 29.30 & 25.90 & 23.70 & 22.81 & 25.43 & \snd{0.880} & 0.775 & 0.667 & 0.643 & 0.741 & 0.114 & 0.236 & 0.369 & 0.449 & 0.292 \\
& Mip-Splatting & \snd{29.39} & \snd{27.39} & \best{26.47} & \best{26.22} & \snd{27.37} & \best{0.884} & \best{0.808} & \best{0.754} & \best{0.765} & \best{0.803} & 0.108 & \best{0.205} & \best{0.305} & \best{0.392} & \best{0.252} \\
\cmidrule(lr){2-17}
& LFP           & \best{29.75} & \best{27.44} & \best{26.48} & \snd{26.19} & \best{27.47} & 0.879 & \snd{0.794} & \snd{0.737} & \snd{0.753} & \snd{0.791} & \best{0.100} & \snd{0.213} & \snd{0.321} & \snd{0.409} & \snd{0.261} \\
\midrule
\multirow{5}{*}{\rotatebox[origin=c]{90}{Zoom-out}}
& Mip-NeRF      & 33.08 & 33.31 & 30.91 & 27.97 & 31.31 & 0.961 & 0.970 & 0.969 & 0.961 & 0.965 & 0.045 & 0.031 & 0.036 & 0.052 & 0.041 \\
& 3DGS          & 33.33 & 26.95 & 21.38 & 17.69 & 24.84 & \best{0.969} & 0.949 & 0.875 & 0.766 & 0.890 & \best{0.030} & 0.032 & 0.066 & 0.121 & 0.063 \\
& 3DGS + EWA    & \best{33.51} & 31.66 & 27.82 & 24.63 & 29.40 & \best{0.969} & 0.971 & 0.959 & 0.940 & 0.960 & \snd{0.031} & 0.024 & 0.033 & 0.047 & 0.034 \\
& Mip-Splatting & \snd{33.36} & \best{34.00} & \snd{31.85} & \snd{28.67} & \best{31.97} & \best{0.969} & \best{0.977} & \best{0.978} & \best{0.973} & \best{0.974} & \snd{0.031} & \best{0.019} & \best{0.019} & \best{0.026} & \best{0.024} \\
\cmidrule(lr){2-17}
& LFP           & 32.76 & \snd{33.55} & \best{32.07} & \best{29.29} & \snd{31.92} & \snd{0.966} & \snd{0.974} & \snd{0.976} & \snd{0.972} & \snd{0.972} & 0.035 & \snd{0.022} & \snd{0.023} & \snd{0.030} & \snd{0.028} \\
\bottomrule
\end{tabular}}
\vspace{-0.5cm}
\end{table}

\begin{figure*}[t]
\centering
%
%
\def\FIGSCALE{0.98}              
\def\VLABFONT{\scriptsize\sffamily\bfseries}   
\pgfmathsetmacro{\LABW}{0.022}   
\pgfmathsetmacro{\GAP} {0.0035}  
\pgfmathsetmacro{\VGAP}{0.006}   
\pgfmathsetmacro{\BGAP}{0.028}   
\pgfmathsetmacro{\APW} {0.150}   
\pgfmathsetmacro{\FDH} {0.0155}  
%
\pgfmathsetmacro{\ARdeblur}{1.500}
\pgfmathsetmacro{\ARaif}   {1.500}   
\pgfmathsetmacro{\ARgrid}  {1.608}   
\pgfmathsetmacro{\ARfish}  {1.500}
\pgfmathsetmacro{\ARtran}  {1.333}
%
\pgfmathsetmacro{\WSIX}{(1-\LABW-5*\GAP)/6}          
\pgfmathsetmacro{\HDEB}{\WSIX/\ARdeblur}
\pgfmathsetmacro{\HFSH}{\WSIX/\ARfish}
\pgfmathsetmacro{\HTRN}{\WSIX/\ARtran}
\pgfmathsetmacro{\HGRD}{\HDEB-\FDH}                  
\pgfmathsetmacro{\WGRD}{\HGRD*\ARgrid}
\pgfmathsetmacro{\XGRD}{\LABW+\WSIX+2.5*\GAP}        
\pgfmathsetmacro{\GGAP}{(1-\APW-\GAP-\XGRD-4*\WGRD)/3}
%
\pgfmathsetmacro{\HROWB}{\HDEB}
\pgfmathsetmacro{\WSIXP}{\WSIX*\FIGSCALE}
\pgfmathsetmacro{\WGRDP}{\WGRD*\FIGSCALE}
\begin{tikzpicture}[x={\FIGSCALE\textwidth},y={\FIGSCALE\textwidth},
    img/.style   ={anchor=north west,inner sep=0pt,outer sep=0pt},
    mlab/.style  ={anchor=north east,inner sep=1.1pt,outer sep=0pt,
                   fill=white,font=\tiny\sffamily,text=black},
    vlab/.style  ={rotate=90,anchor=center,font=\VLABFONT},
    flab/.style  ={anchor=north,font=\scriptsize\sffamily},
    sep/.style   ={line width=1.1pt,dash pattern=on 5pt off 3.5pt,black!55}]

\pgfmathsetmacro{\ya}{0}
\foreach \f/\l [count=\i from 0] in
    {gt/GT, lfp/Ours, dofgs/DoF-GS, deblurnerf/Deblur-NeRF, nerf/NeRF, 3dgs/3DGS}{
  \pgfmathsetmacro{\xx}{\LABW+\i*(\WSIX+\GAP)}
  \node[img] (D\i) at (\xx,\ya) {\includegraphics[width=\WSIXP\textwidth]{cherrypick_deblur_\f}};
  \node[mlab] at (D\i.north east) {\l};
}
\node[vlab] at ({\LABW/2},{\ya-\HDEB/2}) {(a) Deblur};

\pgfmathsetmacro{\yb}{\ya-\HDEB-\VGAP}

\node[img] (F0) at (\LABW,\yb) {\includegraphics[width=\WSIXP\textwidth]{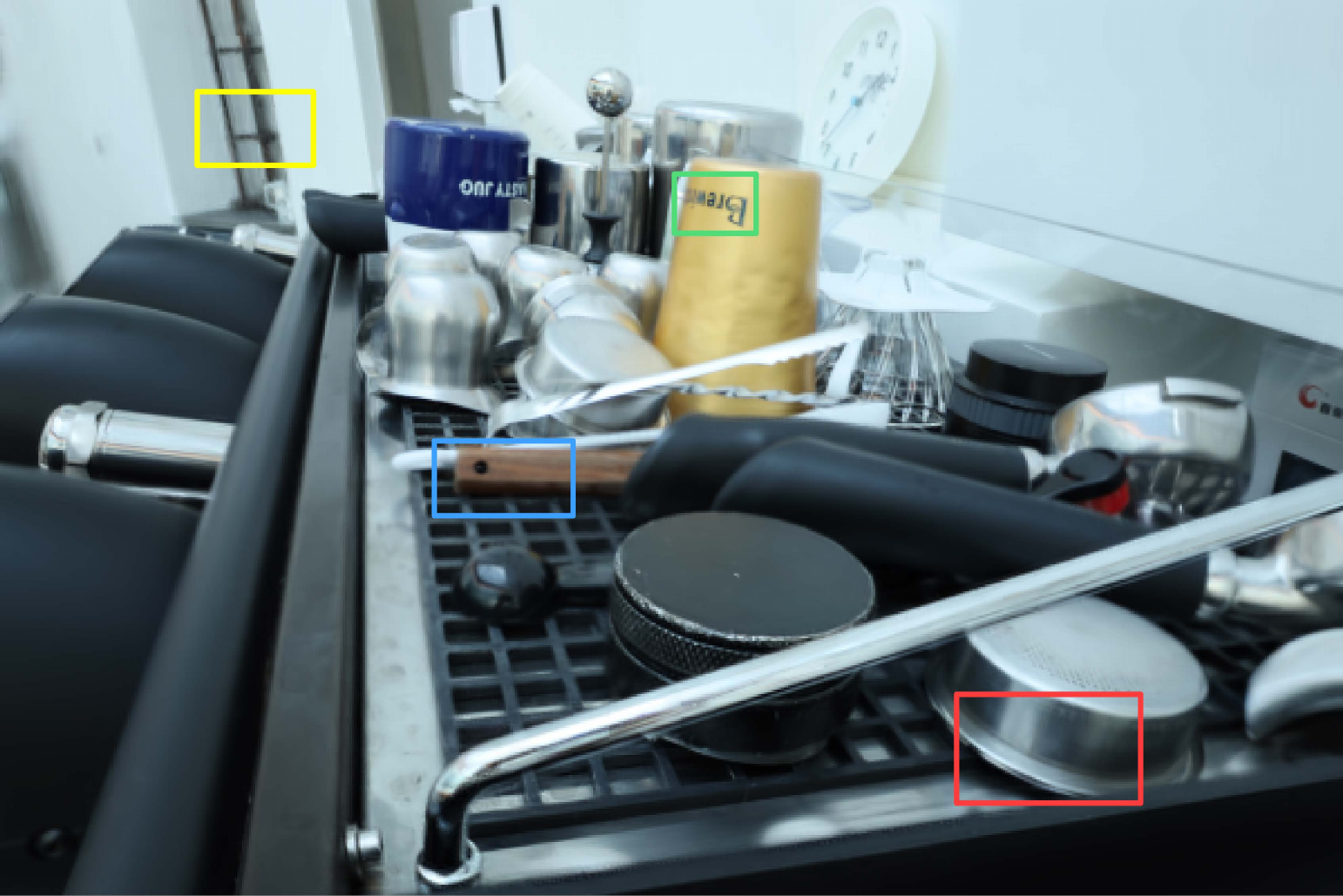}};
\node[mlab] at (F0.north east) {all-in-focus};

\foreach \k/\zf [count=\j from 0] in {1/23.5, 2/35.6, 3/42.9, 4/255.8}{
  \pgfmathsetmacro{\xx}{\XGRD+\j*(\WGRD+\GGAP)}
  \node[img] (G\j) at (\xx,\yb) {\includegraphics[width=\WGRDP\textwidth]{cherrypick_dof_focol\k}};
  \node[flab] at ({\xx+\WGRD/2},{\yb-\HGRD-0.0012}) {focal distance \zf};
  \ifnum\j>0
    \pgfmathsetmacro{\xs}{\xx-\GGAP/2}
    \draw[line width=.3pt,black!45,dash pattern=on 1.6pt off 1.4pt]
          (\xs,{\yb-0.001}) -- (\xs,{\yb-\HDEB});
  \fi
}
\node[vlab] at ({\LABW/2},{\yb-\HROWB/2}) {(b) Refocus};

\pgfmathsetmacro{\irisR}{0.0105}
\pgfmathsetmacro{\xarr}{1-\APW+0.007}                 
\pgfmathsetmacro{\xap} {\xarr+0.010+\irisR}           
\pgfmathsetmacro{\xnum}{\xap+\irisR+0.007}            
\foreach \a/\op [count=\r from 0] in {1/0.26, 1.5/0.46, 2.2/0.68, 3/0.90}{
  \pgfmathsetmacro{\yg}{\yb-(\r+0.5)*\HGRD/4}
  \begin{scope}[shift={(\xap,\yg)}]
    \pgfmathsetmacro{\rr}{\op*\irisR}
    \fill[black!14] (0,0) circle (\irisR);
    \fill[white]    (0,0) circle (\rr);
    \draw[line width=.4pt,black!60] (0,0) circle (\irisR);
    \foreach \b in {0,60,...,300}{\draw[line width=.32pt,black!60] (\b:\rr)--({\b+60}:\irisR);}
  \end{scope}
  \node[anchor=west,font=\small,inner sep=0pt] at (\xnum,\yg) {\a$\times$};
}
\pgfmathsetmacro{\atip}{\yb-\HGRD+0.003}
\draw[line width=.55pt,black!60] (\xarr,{\yb-0.004}) -- (\xarr,{\atip+0.005});
\fill[black!60] (\xarr,\atip) -- ++(-0.0035,0.0065) -- ++(0.007,0) -- cycle;
\node[flab,black!75] at ({(\xarr+\xnum+0.030)/2},{\yb-\HGRD-0.0012}) {~~~aperture};

\pgfmathsetmacro{\ysepa}{\yb-\HROWB-\BGAP/2}
\draw[sep] (0,\ysepa) -- (1,\ysepa);

\pgfmathsetmacro{\yc}{\ysepa-\BGAP/2}
\foreach \dir [count=\rr from 0] in {fisheye, fisheye2}{
  \pgfmathsetmacro{\yy}{\yc-\rr*(\HFSH+\VGAP)}
  \foreach \f/\l [count=\i from 0] in
      {gt/GT, lfp/Ours, fisheyegs/Fisheye-GS, 3dgut/3DGUT, mipnerf360/mip-NeRF 360, 3dgs/3DGS}{
    \pgfmathsetmacro{\xx}{\LABW+\i*(\WSIX+\GAP)}
    \node[img] (H\rr\i) at (\xx,\yy) {\includegraphics[width=\WSIXP\textwidth]{cherrypick_\dir_\f}};
    \node[mlab] at (H\rr\i.north east) {\l};
  }
}
\foreach \rw/\cl/\bxa/\bya/\bxb/\byb/\side in {%
    0/2/0.004/0.820/0.235/1.000/TR,
    0/3/0.004/0.820/0.235/1.000/TR,
    0/4/0.004/0.820/0.235/1.000/TR,
    0/5/0.004/0.820/0.235/1.000/TR,
    0/2/0.004/0.004/0.105/0.120/BR,
    1/2/0.004/0.828/0.120/1.000/TR}{
  \pgfmathsetmacro{\yy}{\yc-\rw*(\HFSH+\VGAP)}
  \pgfmathsetmacro{\xfg}{\LABW+\cl*(\WSIX+\GAP)}
  \pgfmathsetmacro{\gxa}{\xfg+\bxa*\WSIX}\pgfmathsetmacro{\gxb}{\xfg+\bxb*\WSIX}
  \pgfmathsetmacro{\gya}{\yy-\bya*\HFSH} \pgfmathsetmacro{\gyb}{\yy-\byb*\HFSH}
  \ifthenelse{\equal{\side}{TR}}
    {\draw[green!60!black,line width=1.7pt,dash pattern=on 3.4pt off 2.2pt]
           (\gxa,\gya) -- (\gxb,\gya) -- (\gxb,\gyb);}
    {\draw[green!60!black,line width=1.7pt,dash pattern=on 3.4pt off 2.2pt]
           (\gxa,\gyb) -- (\gxb,\gyb) -- (\gxb,\gya);}
}
\pgfmathsetmacro{\yfb}{\yc-2*\HFSH-\VGAP}
\node[vlab] at ({\LABW/2},{(\yc+\yfb)/2}) {(c) Fisheye};

\pgfmathsetmacro{\ysepb}{\yfb-\BGAP/2}
\draw[sep] (0,\ysepb) -- (1,\ysepb);

\pgfmathsetmacro{\yd}{\ysepb-\BGAP/2}
\foreach \dir [count=\rr from 0] in {transparent1, transparent2}{
  \pgfmathsetmacro{\yy}{\yd-\rr*(\HTRN+\VGAP)}
  \foreach \f/\l [count=\i from 0] in
      {gt/GT, lfp/Ours, rtsplatting/RT-Splatting, 3dgrt/3DGRT, eikon/Eikonal Fields, 3dgs/3DGS}{
    \pgfmathsetmacro{\xx}{\LABW+\i*(\WSIX+\GAP)}
    \node[img] (T\rr\i) at (\xx,\yy) {\includegraphics[width=\WSIXP\textwidth]{cherrypick_\dir_\f}};
    \node[mlab] at (T\rr\i.north east) {\l};
  }
}
\pgfmathsetmacro{\ytb}{\yd-2*\HTRN-\VGAP}
\node[vlab] at ({\LABW/2},{(\yd+\ytb)/2}) {(d) Transparent};

\end{tikzpicture}
\vspace{-0.2cm}
\caption{Qualitative comparisons across different settings.}
\label{fig:qualitative}
\vspace{-0.7cm}
\end{figure*}

\paragraph{Defocus deblurring and refocusing.}
In this setting, training images are captured with a wide aperture and the held-out views are sharp, so a method must explain blurred observations with a scene that renders sharp. We apply the aperture term in Sec.~\ref{sec:supp-dof} when rendering the training views and drop it at test time, adding two scalars per training image, aperture and focal distance, which are optimized alongside the scene from a single initialization shared by all scenes.
Results are in Tab.~\ref{tab:dof}. Methods that do not model the
aperture stay ineffective, neither 3DGS nor Mip-Splatting can ease the blur. Restoring their renderings afterwards with
Restormer~\citep{zamir2022restormer} recovers part of the gap but works on
images rather than on the representation. LFP is among methods built for this task, ranking third on PSNR and second on SSIM.
Fig.~\ref{fig:qualitative} (a) shows the deblurred results: LFP recovers sharp structures and fine image details from the blurred observations, while the general-purpose baselines retain noticeable blur and the specialized methods produce comparable sharpness.

\emph{What this takes on either side is again not comparable.} Those specific methods carry the aperture in the operator or attached to the primitives: the blur is predicted by an auxiliary network~\citep{ma2022deblur}, guided by a prior computed over the training images~\citep{lee2024sharp}, or carried per primitive through the projection with the losses that keep it consistent~\citep{wang2025dof}. Differently, in LFP, the whole model of the aperture is simply two scalars per training image, with no auxiliary network and no added loss. What changes between the two settings is only when that term is applied: it is present when rendering the blurred training views and dropped at test time.

\begin{table}[t]
\centering
\begin{minipage}[t]{0.28\textwidth}
\centering
\small
\caption{Defocus NVS results on~\citet{ma2022deblur}.}
\vspace{-0.3cm}
\label{tab:dof}
\setlength{\tabcolsep}{2pt}
\scalebox{0.78}{
\begin{tabular}{llcc}
\toprule
& Method & PSNR $\uparrow$ & SSIM $\uparrow$ \\
\midrule
\multirow{3}{*}{\rotatebox[origin=c]{90}{General}}
& 3DGS             & 20.00 & 0.593 \\
& 3DGS+deblur & 21.78 & 0.667 \\
& NeRF             & 22.40 & 0.667 \\
\midrule
\multirow{3}{*}{\rotatebox[origin=c]{90}{Specific}}
& DOF-GS      & \best{23.48} & \best{0.742} \\
& Deblur-NeRF & \snd{23.40} & \trd{0.716} \\
& Sharp-NeRF  & 22.82 & 0.694 \\
\midrule
& LFP         & \trd{23.33} & \snd{0.728} \\
\bottomrule
\end{tabular}}
\end{minipage}
\hfill
\begin{minipage}[t]{0.33\textwidth}
\centering
\small
\caption{Fisheye NVS results on~\citet{yeshwanth2023scannet++}.}
\vspace{-0.3cm}
\label{tab:fisheye}
\setlength{\tabcolsep}{2pt}
\scalebox{0.78}{
\begin{tabular}{llccc}
\toprule
& Method & PSNR $\uparrow$ & SSIM $\uparrow$ & LPIPS $\downarrow$ \\
\midrule
\multirow{3}{*}{\rotatebox[origin=c]{90}{NeRF}}
& Instant NGP &26.53 &0.847 &0.273\\
& Nerfstudio    & 25.14 & 0.848 &0.247 \\
& M-NeRF 360    &27.54 &0.878 &0.192    \\
\midrule
\multirow{3}{*}{\rotatebox[origin=c]{90}{GS}}
& 3DGS      & {23.70} & {0.805} &0.178 \\
& FisheyeGS & \trd{28.14} & \best{0.894} &\best{0.152} \\
& 3DGUT  & \snd{28.47} & \snd{0.892} &\trd{0.169} \\
\midrule
& LFP         & \best{28.52} & \trd{0.889} &\snd{0.167} \\
\bottomrule
\end{tabular}}
\end{minipage}
\hfill
\begin{minipage}[t]{0.33\textwidth}
\centering
\small
\caption{Transparent object NVS results on~\citet{bemana2022eikonal}.}
\vspace{-0.3cm}
\label{tab:transparent}
\setlength{\tabcolsep}{2pt}
\scalebox{0.78}{
\begin{tabular}{llccc}
\toprule
& Method
& PSNR $\uparrow$ & SSIM $\uparrow$ & LPIPS $\downarrow$ \\
\midrule
\multirow{3}{*}{\rotatebox[origin=c]{90}{General}}
& NeRF          & 22.05 & 0.689 & 0.285 \\ 
& Mip-Splatting & 22.93 & 0.819 & 0.097 \\ 
& 3DGS          & \trd{23.26} & \snd{0.830} & \snd{0.090} \\ 
\midrule
\multirow{3}{*}{\rotatebox[origin=c]{90}{Specific}}
& Eikonal Fields & 21.54 & 0.700 & 0.258 \\ 
& 3DGRT          & 22.83 & 0.812 & 0.107 \\ 
& RT-Splatting   & \snd{23.37} & \trd{0.827} & \trd{0.091} \\ 
\midrule
& LFP            & \best{23.66} & \best{0.831} & \best{0.084} \\ 
\bottomrule
\end{tabular}}
\end{minipage}
\vspace{-0.3cm}
\end{table}

%
%
%

Refocusing then costs nothing further. Where the blur is fitted per
primitive or per view, changing the focus after training means fitting it
again. DOF-GS also supports post-capture refocusing~\citep{wang2025dof}, but its aperture modeling remains part of the rendering machinery. In LFP the aperture never enters the representation: the two scalars belong to the camera, and rendering at other values is the same closed-form update evaluated differently (Sec.~\ref{sec:aa_and_deblur}).
Fig.~\ref{fig:qualitative} (b) shows a refocusing sweep at different focal distances. LFP moves the sharp region consistently with the selected focus, producing the expected transition from near to far objects while preserving the appearance of the scene.

\paragraph{Fisheye NVS.}
In this setting, the calibrated lens departs far from a linear projection. Nothing is added for the setting: the
pixel-to-ray step of \eqref{eq:ray-coord} follows the fisheye model instead
of a pinhole one, and every subsequent stage, the residual, the response,
and the compositing, is the one used everywhere else.
Results are in Tab.~\ref{tab:fisheye}. The two families separate here for a
reason: a radiance field is queried along whatever rays it is given, so the fisheye model enters only where the rays are generated and the NeRF family also handles the setting without modification~\citep{mueller2022instant,tancik2023nerfstudio,barron2022mip}. A projection cannot be given rays. 3DGS has to be run on undistorted images instead, which discards the periphery the lens was chosen for and resamples what remains, and it falls well behind the rest of its family~\citep{liao2024fisheye,wu20253dgut}; the methods built for this setting keep the raw capture,
either by replacing the projection with one the lens
defines~\citep{liao2024fisheye} or by propagating each primitive through the
distortion~\citep{wu20253dgut}. Both rebuild the step that reaches a
primitive, and both are specific to the lens model they were written for.
LFP leads on PSNR and follows closely on the other two, with no stage of the
renderer changed: a ray is computed differently and then indexed by the same
four coordinates as any other. In this respect LFP holds both sides of the
table at once, taking rays as they come, as a field does, and rasterizing
them in real time, as the point-based family does.
Fig.~\ref{fig:qualitative} (c) compares two fisheye scenes against the baselines. While the central regions highlighted in the red boxes are reconstructed comparably well, some baselines exhibit noticeable artifacts or loss of detail toward the image boundaries, particularly in the corner regions highlighted in green. In contrast, LFP maintains consistent reconstruction quality across both central and peripheral regions.

\paragraph{Transparent objects.}
Through a refractive interface, rays reaching the camera are bent before
measurement, so the observed radiance is no longer organized around the
physical location of the object. Nevertheless, the refracted rays still
exhibit coherent radiance across views. Sec.~\ref{sec:m-prim} defines
primitives by grouping rays with coherent radiance and makes no assumption
about the source of this coherence. Consequently, the ray groups formed in a
refractive region are identical in definition to those elsewhere. A
refracted ray is therefore represented in exactly the same way as any other
ray, and the method applies without modification.
Tab.~\ref{tab:transparent} reports results on scenes with transparent objects. Methods that model the refraction do not clearly gain
from it: the eikonal formulation traces rays through a recovered
index-of-refraction field~\citep{bemana2022eikonal} and stays below the
methods that ignore the bending, while 3DGRT~\citep{moenneloccoz20243d} and RT-Splatting~\citep{shi2026rt}, which follow bent rays against the
Gaussians, match 3DGS rather than exceed it.
LFP is ahead of these arts, though by margins that leave refraction far from solved. The point is
not the size of the lead but where it comes from: LFP adds nothing but still offers better results than those specific methods, suggesting that refraction may be less a
modeling problem than a question of what a representation is indexed by.
Fig.~\ref{fig:qualitative} (d) shows two scenes cropped to the refractive regions. Despite without explicit modeling, LFP preserves the contours and internal appearance of the glass and liquid objects, while the baselines show stronger artifacts and loss of detail in these regions.

\section{Discussion and Conclusion}
\paragraph{Limitations.}
Two limits follow from what the representation leaves out. LFP records rays
rather than matter, so nothing in it is a surface, and tasks that need an
explicit geometry are not served by the representation as it stands. The second is a debt of the parameterization. A single pair of planes cannot
index rays that run nearly parallel to it, so a scene is given several
atlases. Their normals are chosen from the training cameras before
optimization and never change, and a primitive keeps the atlas it was
assigned, so the coverage a scene has follows the capture rather than the
content. Primitives in different atlases also fall back on a depth read in
the camera frame for their ordering. A parameterization that indexes every
direction without partitioning them would remove the second; the first is a
consequence of recording a light field rather than a scene.

\paragraph{Conclusion.}
We present LFP, which represents a scene as a compact set of
differentiable primitives in the classical two-plane parameterization. Each
primitive holds one learned record for a group of rays and answers a query by
how far the query departs from the coupling that defines the group, without
projecting a primitive, intersecting one, or marching along a ray. A scene is
optimized from posed images and rendered in real time, as with
point-based methods. Departures from the ideal imaging model are settled in
the coordinates the primitives are written in: a wider tolerance gives
multi-scale anti-aliasing and defocus deblurring with post-hoc refocusing, a
fisheye capture changes only how a ray is computed, and refraction asks for
nothing, in each case matching methods built for that effect
alone. Taken together, these are not four methods but one representation, and
they suggest that ray space is a place to anchor a scene that the NeRF and
3DGS families have left open.

\newpage
\subsection*{AI use statement}

We used generative AI tools to assist with writing: drafting and revising
prose in the main paper and the supplementary material, improving clarity and
grammar, and formatting tables and figure captions in \LaTeX. Every claim,
number, equation, and citation in the resulting text was written or verified
by the authors against the implementation and the cited sources.
We used generative AI tools to assist with code, specifically, e.g. evaluation scripts. Such code was read
and tested by the authors.
We did not use generative AI tools for research ideation or experimental
design. Retrieval of related work was carried out by the authors; where a tool was used to surface candidate references, each was located and read in the original before being cited.

We take responsibility for the final content of this work, including text,
claims, and artifacts produced with the aid of generative AI.

\bibliography{iclr2027_conference}
\bibliographystyle{iclr2027_conference}
\appendix
\section{Related Work}

\noindent\textbf{Novel view synthesis (NVS).}
NeRF~\citep{mildenhall2021nerf} and 3DGS~\citep{kerbl20233d} families dominate current NVS task, and they operate on different sides of the 3D-to-2D mapping: the former by integrating a field along the ray, the latter by projecting 3D Gaussians onto the image plane.

Specifically, NeRF stores a scene as a continuous radiance field indexed by 3D position and view direction. A pixel is rendered by marching along its camera ray and integrating the field at the sample points. Later work keeps that operator and changes what sits at the samples. Hash grids, sparse voxels and tensor factorizations replace the coordinate network and cut training from days to minutes \citep{mueller2022instant,fridovichkeil2022plenoxels,chen2022tensorf}. A second line makes the field scale-aware, encoding the cone a pixel subtends instead of a point on the ray, and carries that encoding to unbounded scenes and to feature grids \citep{barron2021mip,barron2022mip,barron2023zip};
3DGS regards a scene as a set of anisotropic Gaussians with opacity and view-dependent color. Each Gaussian is projected onto the image plane, where an affine approximation of the projection gives its screen-space footprint, following~\citep{zwicker2001ewa,zwicker2002ewa}. Sorting and alpha-compositing the projected Gaussians inside a tile-based rasterizer gives real-time rendering, and adaptive density control grows and prunes the set during optimization. Later work revises the Gaussians or the stages around it, replacing the footprint with a ray-disk intersection \citep{huang20242d}, adding an opacity field for surfaces \citep{yu2024gaussian}, or decoding attributes from anchor points \citep{lu2024scaffold}.

These two families differ in their rendering mechanisms, yet both anchor the scene in 3D space and rely on mapping operators to bridge that space with the sensor. We introduce LFP as a third option. LFP inherits the sparse differentiable optimization and tile-based rasterization of point-based methods, but defines its primitives natively in 4D ray space. Rather than projecting 3D primitives onto the image plane or integrating a field along a ray, LFP couples each primitive to a camera ray through an epipolar response function evaluated directly in ray space.

\noindent\textbf{Light field rendering (LFR).}
LFR records radiance as a function of rays rather than matter that emits it \citep{levoy1996light,gortler1996lumigraph}. A ray is parameterized by its intersections with two parallel planes, and a novel view is rendered by tracing each camera ray through those planes, querying the corresponding 4D coordinate, and resampling the stored radiance from its discrete neighbors. Because the representation is an explicit table of rays, optical effects reduce to operations on that table: integrating the rays admitted by a synthetic aperture yields depth of field and post-capture refocusing \citep{isaksen2000dynamically,ng2005light}; prefiltering the table before resampling suppresses aliasing \citep{levoy1996light}.

The cost of the simplicity is density. Alias-free reconstruction requires a sampling rate fixed by the scene depth range \citep{chai2000plenoptic}, so the table is large and views are bounded by the captured planes. Unrecorded rays can only be interpolated from neighbors, with hand-designed weights for unstructured capture \citep{buehler2001unstructured}. The table is also static: it cannot be optimized against a sparse set of images.
Learned successors escape the density at the cost of the explicit table. Layered representations collapse the 4D ray record into a stack of fronto-parallel depth planes \citep{zhou2018stereo,mildenhall2019local,wizadwongsa2021nex}: the radiance of a ray is assembled by compositing the planes it pierces, and the continuous ray coordinate is discretized into plane indices and pixel grids. Neither an aperture integral nor a 4D prefilter can act on such a stack ray by ray, so both effects must be approximated with per-plane warps and blur kernels. Neural light fields compress the entire table into the weights of a network that maps ray coordinates to color \citep{sitzmann2021light,attal2022learning}; with no explicit ray set left to integrate or prefilter, the effects either emerge as regularities learned from the training data or must be simulated by many network queries per pixel. Generalizable renderers sample source views along epipolar lines and blend their features \citep{suhail2022light,wang2021ibrnet}, and feed-forward models predict primitives directly \citep{charatan2024pixelsplat}. In each case, the optical effects are no longer manipulations of an explicit ray structure, but behaviors recovered through network capacity or auxiliary modules.

LFP restores what these successors gave up. The dense table returns as a sparse set of differentiable primitives, each holding one learned record for a whole group of rays rather than one sample for one ray, so the light field stays explicit yet is no longer dense, and it can now be optimized against posed images. The effects therefore come back as manipulations of the representation: defocus is again an integration over rays, and anti-aliasing is again a prefiltering of them. Epipolar geometry changes role along the way, from a depth cue or a view-sampling rule to the relation that gathers the rays sharing a primitive's record. This keeps the optical advantages of the light field without its density burden, with no separate mechanism for any effect.

\begin{table}[t]
\centering
\small
\caption{Fields of a primitive $i$. Floats are per primitive, with $K{=}3$ lobes.}
\label{tab:params}
\setlength{\tabcolsep}{5pt}
\begin{tabular}{llcl}
\toprule
Field & Stored as & Floats & Role \\
\midrule
$\boldsymbol{\mu}_i$ & $(\mu^u_i,\mu^v_i,\mu^s_i,\mu^t_i)$ & 4 & anchor ray; $(\mu^u_i,\mu^v_i)$ frozen after init \\
$p_i$ & $\log p_i$ & 1 & disparity of the group; coefficient of the coupling \\
$\Sigma_i$ & $(r_i,a_i,\theta_i)$ & 3 & extent of the response, via \eqref{eq:sigma-decode} \\
$o_i$ & logit & 1 & peak contribution in compositing \\
$\mathbf{c}_i$ & logit RGB & 3 & base radiance of the record \\
$\mathbf{g}^k_i$ & signed RGB & $3K$ & amplitude of lobe $k$ \\
$\boldsymbol{\omega}^k_i$ & unit vector & $3K$ & axis of lobe $k$ \\
$\mathbf{B}^k_i$ & $(\theta^k_i,\ell^k_i,a^k_i)$ & $3K$ & tangent frame and the two bandwidths \\
\midrule
atlas index & integer & 1 & selects the plane pair; not optimized \\
\midrule
\multicolumn{2}{l}{Total, $K{=}3$} & 40 & \\
\bottomrule
\end{tabular}
\end{table}

\section{Model details}
\label{sec:a-model}

\paragraph{Fields of a primitive.}
Sec.~\ref{sec:m-prim} specifies a primitive by
$\{\boldsymbol{\mu}_i, p_i, \Sigma_i, o_i, \mathbf{c}_i\}$ and
Sec.~\ref{sec:m-impl} adds the directional lobes. Three of these are
constrained, and none is optimized in the form it is used. The disparity is
stored as $\tilde p_i$ and decoded by $p_i=\exp\tilde p_i$, which keeps it
positive without bracketing its range. The opacity is stored as a logit and
passed through a sigmoid, giving $o_i\in(0,1)$. The covariance is stored as
three scalars with disjoint geometric roles, a log scale $r_i$, a log aspect
ratio $a_i$, and an angle $\theta_i$, from which
\begin{equation}
\label{eq:sigma-decode}
\Sigma_i=\mathbf{R}(\theta_i)
\operatorname{diag}\bigl(e^{2(r_i+a_i/2)},\,e^{2(r_i-a_i/2)}\bigr)
\mathbf{R}(\theta_i)^{\top},
\qquad
\mathbf{R}(\theta)=\begin{pmatrix}\cos\theta & -\sin\theta\\
\sin\theta & \cos\theta\end{pmatrix},
\end{equation}
symmetric positive definite for any value the three take, so no projection or
clipping is needed during optimization. The two exponentials are the squared
half-axes of the response, and $\mathbf{R}(\theta_i)$ orients them in the
tangent plane. The renderer consumes the Cholesky factor of $\Sigma_i$, which
\eqref{eq:sigma-decode} produces in closed form; the factor is clamped from
below at $10^{-4}$ on its diagonal, the same bound the kernel enforces.

Each lobe of the appearance model is stored likewise. Its amplitude
$\mathbf{g}^k_i$ is a signed RGB triple and its axis
$\boldsymbol{\omega}^k_i$ a vector normalized before use. The matrix
$\mathbf{B}^k_i$ of \eqref{eq:appearance}, whose columns are the two tangent
axes scaled by the square roots of the bandwidths, is assembled from three
scalars: an angle $\theta^k_i$ that rotates the tangent frame about
$\boldsymbol{\omega}^k_i$, and two log bandwidths kept as a mean and a
ratio, $\lambda^k_x=\exp(\ell^k_i+a^k_i/2)$ and
$\lambda^k_y=\exp(\ell^k_i-a^k_i/2)$. Splitting scale from anisotropy in
this way, for both $\Sigma_i$ and the lobes, keeps the two directions of a
covariance from competing through a shared parameter.

A primitive also carries an atlas index, assigned at initialization and
fixed thereafter. It is not optimized and enters nothing but the choice of
which $(\mathbf{e}_1,\mathbf{e}_2,\mathbf{n})$ and which precomputed camera
terms of \eqref{eq:ray-coord} to read. Tab.~\ref{tab:params} lists every
field with its role and its cost.

\paragraph{Adaptive control of the primitive count.}
When primitives are initialized from a sparse structure from motion cloud we
follow the schedule of 3DGS~\citep{kerbl20233d}, cloning and splitting every
$100$ iterations between iteration $500$ and $10\,000$, and resetting
opacities to $0.06$ every $3000$ iterations between iteration $3000$ and
$10\,000$. The criterion is the one described in Sec.~\ref{sec:m-impl}:
gradients are accumulated on $\boldsymbol{\mu}_i$ and $p_i$, the parameters
that place a ray group, since no projected 2D mean enters our response, and a
primitive is selected when the running average exceeds $2\times10^{-5}$. A
selected primitive is cloned when its larger radius stays below $0.005$ of the
scene extent and split otherwise.

A clone keeps the shape, orientation and atlas of its parent and is displaced
inside the $st$ plane: each of the two coordinates receives an independent
Gaussian offset of standard deviation $\tfrac{1}{2}\max(r^x_i,r^y_i)$, half the
primitive's own larger half-axis, so the pair separates by roughly its own
width instead of starting as a duplicate. Parent and child each keep half the
parent's opacity, which leaves the tangential coverage unchanged, and the next
opacity reset removes whichever of the two the photometric loss does not
support. Because the child sits elsewhere in the scene, its base radiance is
not inherited but resampled: the child's centre is projected into the $20$ most
recent training views and read off the ground truth, falling back to the
parent's value where it is not visible. The directional lobes are inherited
unchanged.

A split replaces the parent by two children that inherit its record, its atlas
and its base radiance. It is area-conserving and de-elongating: the children
keep the shorter radius and halve the longer one, so an elongated primitive
loses one factor of two in aspect and takes $1/\sqrt{2}$ of the parent's scale,
while an already round one shrinks isotropically by that same factor.

Primitives are pruned when their opacity falls below $0.005$, and additionally
when their larger radius exceeds $0.1$ of the scene extent. 

\begin{figure}[t]
  \centering
  \setlength{\tabcolsep}{1pt}
  \begin{tabular}{@{}c@{\hspace{20pt}}c@{\hspace{20pt}}c@{\hspace{20pt}}c@{}}
    \includegraphics[width=0.28\textwidth]{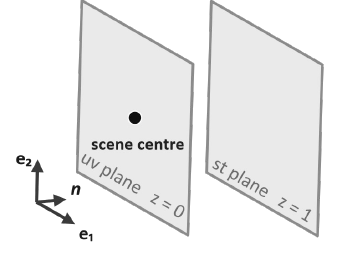} &
    \includegraphics[width=0.18\textwidth]{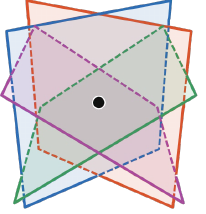} &
    \includegraphics[width=0.18\textwidth]{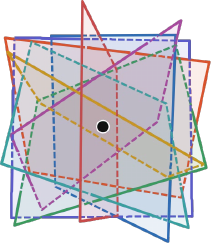} &
    \includegraphics[width=0.18\textwidth]{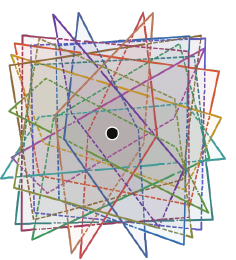} \\
    {(a) 1 atlas} & {(b) 4 atlas} & {(c) 8 atlas} & {(d) 16 atlas} 
  \end{tabular}
  \vspace{-0.2cm}
  \caption{\textbf{Atlas construction.} A single atlas, shown in (a), is the classical two-plane parameterization of LFR~\citep{levoy1996light}: a $uv$ plane through the scene centre and an $st$ plane one unit along its normal $\mathbf{n}$. Only the $uv$ planes are drawn in (b) - (d), each carrying its own $st$ plane one unit away. Every atlas passes through the scene centre, so they differ only in orientation. The orientations are spread evenly here for legibility, and they are the $K$-means centres of the training camera directions in implementation.}
  \label{fig:atlas}
  \vspace{-0.5cm}
\end{figure}

\paragraph{Atlas construction.}
Atlas normals are chosen before optimization from the forward directions of
the training cameras, and are fixed thereafter. The vertical axis is not
assumed: it is taken as the direction of least variance of that distribution,
so a scene is handled the same way whichever world axis happens to be up. The
normals are then the centres of a $k$-means clustering of the camera
directions, and we use $16$ clusters on every
benchmark. Fig.~\ref{fig:atlas} shows our atlas construction.

A point can belong to an atlas only when it lies in front of that atlas's $uv$
plane, as otherwise the disparity it would receive is not positive. Among the
atlases that remain, each point of the initialization cloud takes the one
whose normal best aligns with a surface normal estimated from its $10$ nearest
neighbours by local PCA, with alignment measured by the absolute dot product,
since a surface normal and its opposite describe the same plane. Once made,
the assignment is kept for the rest of training. Because the atlas count is small and each primitive's atlas is fixed, the
whole atlas table is built once per frame and held in shared memory for the
whole tile: the basis $(\mathbf{n},\mathbf{e}_1,\mathbf{e}_2)$ of each atlas,
which depends on neither the camera nor the query, together with the
projections $\mathbf{n}^\top\mathbf{o}$, $\mathbf{e}_1^\top\mathbf{o}$,
$\mathbf{e}_2^\top\mathbf{o}$ of the camera centre, which depend on the camera
alone. Note that if each primitive carried its own plane pair, the same quantities would be per-primitive, and neither the computation nor the memory traffic could be amortized over a tile.

\section{Tile assignment without projection}
\label{sec:a-raster}

\paragraph{What has to be replaced.}
A tile-based rasterizer needs, for each primitive, the set of pixels whose
queries can receive an appreciable response from it, so that the primitive
is listed only in the tiles it can affect. In 3DGS this set is read off the
projection: a 3D Gaussian is mapped to a 2D Gaussian on the image plane by
an affine approximation of the projection~\citep{zwicker2002ewa}, and the
screen extent follows from the projected covariance. LFP has no such object
to project. A primitive is a ray group and a record; its $\Sigma_i$ weighs
the residual of \eqref{eq:residual}, a displacement between two-plane
coordinates and not a screen quantity. What is available instead is the
response as a function of the pixel, and both quantities the rasterizer
needs are read from that function directly: its centre is the zero of the
residual, and its extent is the residual's slope.

\paragraph{The residual as a function of the pixel.}
A pixel $\mathbf{x}(x,y)$ gives a query by \eqref{eq:ray-coord}, so composing the two
makes the residual \eqref{eq:residual} a function of the pixel, which we write $\mathbf{r}_i(\mathbf{x})$. It vanishes exactly for the queries that belong to primitive $i$'s ray group, and a view contains one such query: the one passing through $(\xi_i,\eta_i)$ on the plane $z=1/p_i$, the tangential coordinate the group is defined by. We write $\mathbf{x}^{\mathrm{ref}}$ for its image coordinates and call it the reference position. Since $(\xi_i,\eta_i)$ is determined by $\boldsymbol{\mu}_i$ and $p_i$, that query is known, leaving the camera centre towards $\xi_i\mathbf{e}_1+\eta_i\mathbf{e}_2+\mathbf{n}/p_i$, so inverting the pixel-to-ray map of \eqref{eq:ray-coord} gives $\mathbf{x}^{\mathrm{ref}}$ directly, with no search over pixels, as written out below. It is where the response of primitive $i$ reaches one, its largest value.

Away from $\mathbf{x}^{\mathrm{ref}}$ the residual is not affine in $(x,y)$:
the ray parameters $\lambda_0,\lambda_1$ of \eqref{eq:ray-coord} are rational
in the direction, which is itself a projective function of the pixel, so the
exact residual over a neighbourhood is expensive to obtain. A bounding
rectangle does not need it, only how fast the response falls off around its
peak, so we expand about $\mathbf{x}^{\mathrm{ref}}$, where that peak sits,
and take
\begin{equation}
\label{eq:J-pix}
\mathbf{J}_i=\frac{\partial\mathbf{r}_i}{\partial(x,y)}\in\mathbb{R}^{2\times2},
\end{equation}
the rate at which the residual moves per pixel step, evaluated by finite
differences from \eqref{eq:ray-coord} at $\mathbf{x}^{\mathrm{ref}}$ and at
$\mathbf{x}^{\mathrm{ref}}$ displaced by one pixel along each axis. Those
coordinates need not be integers: \eqref{eq:ray-coord} is defined at any point
of the image plane, and a unit step is only the interval over which the
difference is taken. This costs three evaluations of a map that is already
needed for rendering, and avoids differentiating the projective composition.
Note that this linearization decides only which tiles a primitive is listed
in. Once a pixel is being shaded, its residual is computed from
\eqref{eq:ray-coord} and \eqref{eq:residual} directly, so a listed primitive
contributes exactly what \eqref{eq:response} says, and a slightly loose
rectangle costs only a few wasted evaluations.

\paragraph{From the residual to a screen extent.}
The rasterizer needs a range of pixels, while \eqref{eq:J-pix} gives the rate
at which the residual grows as the pixel moves away from
$\mathbf{x}^{\mathrm{ref}}$. What connects them is the point at which the
response becomes negligible: the kernel evaluates \eqref{eq:response} only
while the query stays within five standard deviations of the ray group and
discards it beyond, that is, while
\begin{equation}
\label{eq:cutoff}
\mathbf{r}_i^\top\bigl(\Sigma_i+\rho_i^2\mathbf{I}\bigr)^{-1}\mathbf{r}_i\le25,
\end{equation}
with $\Sigma_i+\rho_i^2\mathbf{I}$ the covariance the kernel reads
(Sec.~\ref{sec:m-render}). A rate and a cutoff together give a range.

Let $\Delta\mathbf{x}=\mathbf{x}-\mathbf{x}^{\mathrm{ref}}$ be the offset of a
pixel from the reference position of primitive $i$. Expanding
\eqref{eq:residual} about $\mathbf{x}^{\mathrm{ref}}$,
\begin{equation}
\label{eq:taylor}
\mathbf{r}_i(\mathbf{x})
=\underbrace{\mathbf{r}_i(\mathbf{x}^{\mathrm{ref}})}_{=\,\mathbf{0}}
+\;\mathbf{J}_i\,\Delta\mathbf{x}
+O(\lVert\Delta\mathbf{x}\rVert^{2}),
\end{equation}
where the leading term drops out because $\mathbf{x}^{\mathrm{ref}}$ is where
the residual vanishes. Substituting the first-order part into
\eqref{eq:cutoff}, the pixels that matter are those with
\begin{equation}
\label{eq:Jq}
\Delta\mathbf{x}^\top\mathbf{M}_i\,\Delta\mathbf{x}\le25,
\qquad
\mathbf{M}_i=\mathbf{J}_i^\top\bigl(\Sigma_i+\rho_i^2\mathbf{I}\bigr)^{-1}
\mathbf{J}_i .
\end{equation}
Read as a quadratic form, $\mathbf{M}_i$ takes an offset in pixels and returns
the squared Mahalanobis distance of the query at that pixel from the ray
group. It is therefore an inverse covariance in pixel coordinates: a primitive
whose residual moves quickly per pixel step crosses the cutoff within a few
pixels, while one whose residual barely moves covers many. The kernel already
holds the Cholesky factor $\mathbf{L}_i$ of $\Sigma_i+\rho_i^2\mathbf{I}$, so
$\mathbf{M}_i$ is formed as
$(\mathbf{L}_i^{-1}\mathbf{J}_i)^\top(\mathbf{L}_i^{-1}\mathbf{J}_i)$ and no
inverse is taken in the implementation.

\begin{figure}[t]
  \centering
  \setlength{\tabcolsep}{1pt}
  \begin{tabular}{@{}c@{\hspace{2pt}}c@{}}
    \includegraphics[width=0.52\textwidth]{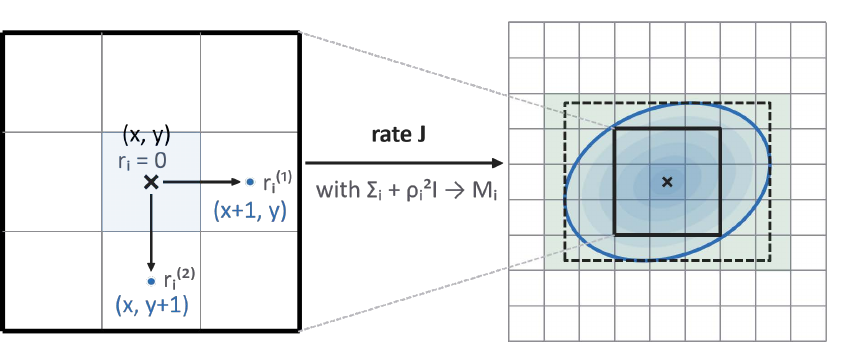} &
\includegraphics[width=0.47\textwidth]{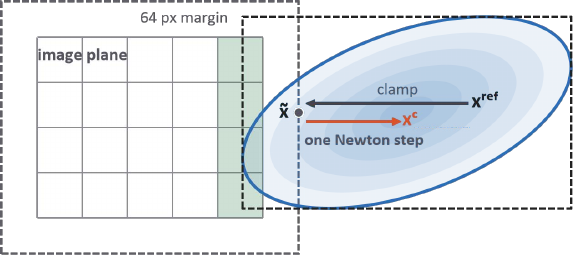}\\
    {\small{(a) From residual evaluations to the tiles}} & \small{{(b) Reference position falls off-image}} 
  \end{tabular}
  \vspace{-0.2cm}
  \caption{\textbf{Tile assignment without projection.} Each cell is a pixel.
\textcolor{capgreen}{Green marks the pixels a primitive can affect}.
\textbf{(a)} Three evaluations of \eqref{eq:ray-coord}, one pixel apart, give
the rate $\mathbf{J}_i$ by \eqref{eq:J-pix}, which together with $\Sigma_i+\rho_i^2\mathbf{I}$
forms the falloff $\mathbf{M}_i$ of \eqref{eq:Jq};
\textcolor{capblue}{the blue ellipse encloses the pixels satisfying
$\Delta\mathbf{x}^\top\mathbf{M}_i\Delta\mathbf{x}\le25$}, and the dashed
rectangle bounds it. \textbf{(b)} When $\mathbf{x}^{\mathrm{ref}}$ falls beyond
the image margin it is clamped to $\tilde{\mathbf{x}}$, and one Newton step
re-centres the rectangle at $\mathbf{x}^{\mathrm{c}}$, the zero of the
first-order model.The rectangle still reaches the image, so the primitive is kept and assigned to the tiles the rectangle covers.}
  \label{fig:tile}
  \vspace{-0.5cm}
\end{figure}

The pixels satisfying \eqref{eq:Jq} form an ellipse centred on
$\mathbf{x}^{\mathrm{ref}}$, since $\mathbf{M}_i$ is positive definite. Tiles
are axis aligned, so what the rasterizer needs from that ellipse is its
bounding rectangle, that is, the largest $\lvert\Delta x\rvert$ and
$\lvert\Delta y\rvert$ it contains. For an ellipse
$\Delta\mathbf{x}^\top\mathbf{M}\Delta\mathbf{x}\le c^{2}$, the extreme value
of a coordinate $\mathbf{a}^\top\Delta\mathbf{x}$ is
$c\sqrt{\mathbf{a}^\top\mathbf{M}^{-1}\mathbf{a}}$, so taking
$\mathbf{a}=(1,0)^\top$ and $(0,1)^\top$ gives the two half-widths, written
$\mathrm{ext}_x$ and $\mathrm{ext}_y$, as $c\sqrt{(\mathbf{M}^{-1})_{11}}$ and
$c\sqrt{(\mathbf{M}^{-1})_{22}}$. Writing
\begin{equation}
\small
\label{eq:extent}
\mathbf{M}_i=\begin{pmatrix} A & B\\ B & C \end{pmatrix},
\quad
\mathbf{M}_i^{-1}=\frac{1}{AC-B^{2}}\begin{pmatrix} C & -B\\ -B & A \end{pmatrix},
\quad
\mathrm{ext}_x=\sqrt{\frac{25\,C}{AC-B^{2}}},
\quad
\mathrm{ext}_y=\sqrt{\frac{25\,A}{AC-B^{2}}},
\end{equation}
with $c=5$ from \eqref{eq:cutoff}, so no eigendecomposition is needed. Both
are clamped to $[3,1024]$ pixels: the lower bound keeps a primitive listed in
at least the tile it peaks in, and the upper bound caps how many entries one
primitive can emit when $AC-B^{2}$ is small, the near-singular case treated
next. Every tile the rectangle overlaps receives one entry for $i$. We plot the tile assignment in Fig.~\ref{fig:tile} (a).

\paragraph{When the reference position falls far outside the image.}
Write $\mathbf{P}_i=\xi_i\mathbf{e}_1+\eta_i\mathbf{e}_2+\mathbf{n}/p_i$ for
the point on the plane $z=1/p_i$ the group is defined by. The query that
belongs to the group leaves the camera centre $\mathbf{o}_j$ towards
$\mathbf{P}_i$, so its direction is $\mathbf{d}=\mathbf{P}_i-\mathbf{o}_j$,
and writing $\mathbf{m}=KR^{-1}\mathbf{d}=(m_1,m_2,m_3)^{\top}$, which by
\eqref{eq:ray-coord} is proportional to $(q_x,q_y,1)^{\top}$,
\begin{equation}
\label{eq:refpix}
\mathbf{x}^{\mathrm{ref}}=\Bigl(\frac{m_1}{m_3},\,\frac{m_2}{m_3}\Bigr),
\end{equation}
with $m_3$ the depth of $\mathbf{P}_i$ in the camera frame, the same quantity used as the sorting key. Primitives with $m_3$ below $0.05$ are
culled before this point, but a small positive $m_3$ still sends
\eqref{eq:refpix} far outside the image, large enough to overflow the tile
arithmetic that follows. We therefore clamp $\mathbf{x}^{\mathrm{ref}}=(x,y)$ componentwise to a
rectangle $64$ pixels wider than the image on every side,
\begin{equation}
\label{eq:clamp}
\tilde{\mathbf{x}}=\bigl(\min(\max(x,-64),\,W+64),\;
\min(\max(y,-64),\,H+64)\bigr),
\end{equation}
for an image of width $W$ and height $H$. Each coordinate that already lies in
its range is left alone, and one that does not is replaced by the nearest
endpoint, so a value of $(300,50)$ on a $100\times100$ image becomes
$(164,50)$. This keeps the arithmetic finite, but it moves the centre off the
zero of the residual, where \eqref{eq:taylor} was expanded.

One step of Newton's method puts it back. At $\tilde{\mathbf{x}}$ the residual
$\mathbf{r}_i(\tilde{\mathbf{x}})$ is nonzero, and the first-order model
$\mathbf{r}_i(\mathbf{x})\approx\mathbf{r}_i(\tilde{\mathbf{x}})
+\mathbf{J}_i(\mathbf{x}-\tilde{\mathbf{x}})$ vanishes at
\begin{equation}
\label{eq:centre}
\mathbf{x}^{\mathrm{c}}
=\tilde{\mathbf{x}}-\mathbf{J}_i^{-1}\,
\mathbf{r}_i(\tilde{\mathbf{x}}).
\end{equation}
Both quantities on the right are already at hand: the three calls to
\eqref{eq:ray-coord} behind \eqref{eq:J-pix} are taken at $\tilde{\mathbf{x}}$
rather than at $\mathbf{x}^{\mathrm{ref}}$, which may be too large to
difference safely, so the same $\mathbf{J}_i$ serves the rectangle and this
step. When no clamping occurred the residual at $\tilde{\mathbf{x}}$ is zero
and $\mathbf{x}^{\mathrm{c}}$ returns $\mathbf{x}^{\mathrm{ref}}$ unchanged;
otherwise it recovers where the linearization places the zero. The rectangle is
centred at $\mathbf{x}^{\mathrm{c}}$, so a primitive whose reference position
lies outside the image is still listed in the tiles its falloff reaches.
Fig.~\ref{fig:tile} (b) plots this case.

Every primitive past the near cull follows these same three steps, whether or
not its reference position lies in the image: the three evaluations behind
$\mathbf{J}_i$, the solve of \eqref{eq:centre}, and the check of the rectangle
against the image. Nothing cheaper can be checked first, because whether a
primitive reaches the image depends on its extent, and the extent comes from
$\mathbf{J}_i$. The clamp of \eqref{eq:clamp} is two comparisons per
coordinate and leaves any primitive whose reference position is already within
the margin untouched.

The step is rejected in two cases. A step longer than $128$ pixels is outside
the range over which the linearization was intended to hold. And
$\mathbf{J}_i$ may be near singular, the grazing configuration: the view moves
the residual in essentially one direction, the response is nearly constant
along a line in the image, and no single zero exists to move towards. In both
cases $\tilde{\mathbf{x}}$ serves as the centre and both half-widths are
raised to at least $32$ pixels, so the primitive is listed over a band rather
than at a point.

\paragraph{Ordering and culling.}
The depth key is $m_3$, the depth of $\mathbf{P}_i$ in the camera frame, which
is the ordering rule of Sec.~\ref{sec:m-render}. Primitives with $m_3$ below $0.05$, or whose rectangle does not meet the image, are dropped. With bounds and keys in
hand the remainder is standard: each primitive emits one entry per tile its
rectangle overlaps, the entries are sorted by tile and then by depth, and each
tile composites its own list with \eqref{eq:composite}. The backward pass
replays that traversal in reverse. Only three steps differ from a point-based
rasterizer, the reference position, the rate $\mathbf{J}_i$, and the Newton
re-centring, and none of them requires a projected covariance.

\section{Anti-aliasing and defocus deblurring}
\label{sec:aa_and_deblur}

A pixel does not correspond to a single ray but a small bundle of them, and both effects
treated here stem from how that bundle spreads: aliasing appears when the pixel grid is too
coarse for its spatial spread, and defocus when a finite aperture spreads it over
directions. LFP accounts for the bundle through a term it already carries.
Sec.~\ref{sec:m-render} renders with $\Sigma_i+\rho_i^2\mathbf{I}$ so that no response is
read more narrowly than a view can resolve, and this same term admits a second reading. Let
the queries of a pixel's bundle differ from its central ray by $\boldsymbol{\delta}$, with
mean zero and covariance $P$ in residual space. Since the response \eqref{eq:response} is an
unnormalized Gaussian and Gaussians are closed under convolution, with covariances
adding~\citep{yu2024mip},
\begin{equation}
\label{eq:supp-conv}
\mathbb{E}_{\boldsymbol{\delta}}\bigl[G_i(\mathbf{r}_i-\boldsymbol{\delta})\bigr]
=\sqrt{\frac{\det\Sigma_i}{\det(\Sigma_i+P)}}\;
\exp\!\Bigl(-\tfrac12\mathbf{r}_i^{\top}(\Sigma_i+P)^{-1}\mathbf{r}_i\Bigr).
\end{equation}
Two things happen and no more: the covariance gains $P$, and the peak falls by the square
root of a determinant ratio. The first is why each effect below is a term added to
$\Sigma_i$, with the record, the response, and the compositing left untouched; the second is
a change in how much a primitive contributes, handled differently in the two settings for
the reason given in Sec.~\ref{sec:supp-peak}. Classical light field rendering obtained both
effects in just this way, by widening what a stored entry averages
over~\citep{levoy1996light,isaksen2000dynamically}.

The pixel's own bundle is the case $P=\rho_i^{2}\mathbf{I}$. The value of $\rho_i$ follows
from the units of \eqref{eq:residual}: a pixel of unit width subtends a half-angle $1/(2f)$,
which at a primitive of camera depth $Z^{\mathrm{cam}}_{ij}$ is a transverse displacement
$Z^{\mathrm{cam}}_{ij}/(2f)$, and \eqref{eq:residual} measures that displacement scaled by
$p_i$, so $\rho_i=Z^{\mathrm{cam}}_{ij}\,p_i/(2f)$, with $f=\max(f_x,f_y)$ when the two
differ. Taking the half-pixel as one standard deviation is a convention rather than a
consequence; what matters below is only the ratio $t_i/\rho_i^{2}$, which is the same under
any convention.

\subsection{Anti-aliasing}
\label{sec:supp-aa}

A coarser output enlarges the spatial spread of the bundle, so we render with
$\Sigma_i+t_i\mathbf{I}$. The peak factor of \eqref{eq:supp-conv} is a change in the total a
primitive contributes, since an unnormalized Gaussian integrates to $2\pi\sqrt{\det\Sigma}$
and preserving that total under a covariance change costs a factor
$\sqrt{\det\Sigma/\det\Sigma'}$ on the opacity~\citep{darmon2024robust}. During training it
is never applied explicitly: $o_i$ is free and is fitted with $\rho_i^{2}$ in place, so the
converged $o_i$ already contains it. Rendering at another scale replaces $\rho_i^{2}$ by
$t_i$, and what must be applied on top of the fitted opacity is the ratio of the two peak
factors, in which $\det\Sigma_i$ cancels,
\begin{equation}
\label{eq:aa}
\alpha_i\;\leftarrow\;\alpha_i
\sqrt{\det(\Sigma_i+\rho_i^{2}\mathbf{I})/\det(\Sigma_i+t_i\mathbf{I})},
\end{equation}
which is the correction \citet{yu2024mip} apply to their own low-pass filter for the same
reason. The cancellation makes it well defined without knowing the fitted $o_i$, and makes
it exactly one when $t_i=\rho_i^{2}$, so rendering at the training scale reproduces ordinary
rendering.

Calibrating $t_i$ takes no free parameter. At $1/s$ of the training resolution an output
pixel averages an $s\times s$ block of training pixels, the box filter of the imaging
process that \citet{yu2024mip} also model, whose variance is $(s^{2}-1)/12$ per axis, and
each of those pixels carries its own footprint of variance $1/4$ in the convention above.
The two are independent, so
\begin{equation}
\label{eq:supp-t}
t_i=\rho_i^{2}\Bigl(1+\frac{s^{2}-1}{3}\Bigr),
\end{equation}
which is one at $s=1$ and $22$ rather than $64$ at $s=8$, the difference being the part of
the widening the training footprint already supplied. Two details matter for reproduction:
both determinants are evaluated on the Cholesky factors the rasterizer draws with, which are
floored for numerical safety, since taking them from the unfloored parameters over-corrects
elongated primitives; and primitives culled by the near plane are given $\rho_i=0$ on both
sides.

At an output finer than the training views the same construction would give
$t_i<\rho_i^{2}$, narrowing every primitive below the width it was fitted at. Since no
observation supports a footprint narrower than the training one, we hold $t_i$ at
$\rho_i^{2}$ there, which leaves \eqref{eq:aa} at one. Either way a trained scene renders at
any resolution with no re-optimization.

Besides the quantitative results presented in Tab.~\ref{tab:aa}, we also present the qualitative results in Fig.~\ref{fig:aa-qual} to show the effectiveness of LFP for the anti-aliasing task. We notice that despite its simplicity, LFP performs favorably against Mip-splatting in both the zoom-in and zoom-out settings. In Fig.~\ref{fig:aa-qual} (a), where the models are trained at $1/8$ resolution on mip-NeRF~360 and rendered at $\times8$, the reference view marks each crop. 3DGS breaks into needle artifacts, 3DGS+EWA hallucinates scratch-like structures, and Mip-Splatting smears colour and washes out fine texture, while ours stays faithful to the ground truth. In Fig.~\ref{fig:aa-qual} (b), all models are trained at full resolution on Blender and rendered down to $1/8$; each image is split into horizontal bands rendered at the resolutions marked on the left ($1/8$ on top), low-resolution bands shown with their actual pixels. 3DGS brightens and dissolves thin structures at low resolutions, while ours stays close to the ground truth at every rate, similar to the specially designed Mip-Splatting.

\newlength{\AAz}\setlength{\AAz}{0.15\textwidth}     
\newlength{\AAr}\setlength{\AAr}{1.5\AAz}             
\newlength{\AAgap}\setlength{\AAgap}{0.0025\textwidth}
\newlength{\AAlbl}\setlength{\AAlbl}{0.015\textwidth} 
\newcommand{\AAQ}[1]{\includegraphics[width=\AAz]{cherrypick_aa_#1}}
\newcommand{\AAR}[1]{\includegraphics[width=\AAr]{cherrypick_aa_#1}}
\newcommand{\AAZIROW}[1]{%
  \AAR{#1_ref.pdf}\hspace*{\AAgap}%
  \AAQ{#1_gt.pdf}\hspace*{\AAgap}\AAQ{#1_lfp.pdf}\hspace*{\AAgap}%
  \AAQ{#1_mip.pdf}\hspace*{\AAgap}\AAQ{#1_ewa.pdf}\hspace*{\AAgap}%
  \AAQ{#1_3dgs.pdf}}
\newcommand{\AAZLBL}{%
  \makebox[\AAr]{\footnotesize\sffamily Reference}\hspace*{\AAgap}%
  \makebox[\AAz]{\footnotesize\sffamily GT}\hspace*{\AAgap}%
  \makebox[\AAz]{\footnotesize\sffamily Ours}\hspace*{\AAgap}%
  \makebox[\AAz]{\footnotesize\sffamily Mip-Splatting}\hspace*{\AAgap}%
  \makebox[\AAz]{\footnotesize\sffamily 3DGS\,+\,EWA}\hspace*{\AAgap}%
  \makebox[\AAz]{\footnotesize\sffamily 3DGS}}
\newlength{\AAc}\setlength{\AAc}{0.19\textwidth}     
\newcommand{\AAP}[1]{\includegraphics[width=\AAc]{cherrypick_aa_#1}}
\newlength{\AAch}\setlength{\AAch}{1.019\AAc}
\newcommand{\AABL}[1]{\rotatebox{90}{\makebox[0.25\AAch][c]{\scriptsize\sffamily #1}}}
\newcommand{\AABANDS}[4]{\makebox[\AAlbl][c]{%
  \vbox{\offinterlineskip
    \hbox{\AABL{#1}}\hbox{\AABL{#2}}\hbox{\AABL{#3}}\hbox{\AABL{#4}}}}}
\newcommand{\AAZOROW}[1]{%
  \AABANDS{$1/8$}{$1/4$}{$1/2$}{Full}\hspace*{2pt}%
  \AAP{#1_gt.pdf}\hspace*{\AAgap}\AAP{#1_lfp.pdf}\hspace*{\AAgap}%
  \AAP{#1_mip.pdf}\hspace*{\AAgap}\AAP{#1_ewa.pdf}\hspace*{\AAgap}%
  \AAP{#1_3dgs.pdf}}
\newcommand{\AAMLBL}[1]{\makebox[\AAc]{\footnotesize\sffamily #1}}
\newcommand{\AAMETHODS}{%
  \makebox[\AAlbl]{}\hspace*{2pt}%
  \AAMLBL{GT}\hspace*{\AAgap}\AAMLBL{Ours}\hspace*{\AAgap}%
  \AAMLBL{Mip-Splatting}\hspace*{\AAgap}\AAMLBL{3DGS\,+\,EWA}\hspace*{\AAgap}%
  \AAMLBL{3DGS}}
\begin{figure*}[t]
\centering
{\footnotesize\sffamily\bfseries (a) zoom-in: trained at $1/8$ resolution, rendered at
$\times8$}\\[2pt]
\AAZIROW{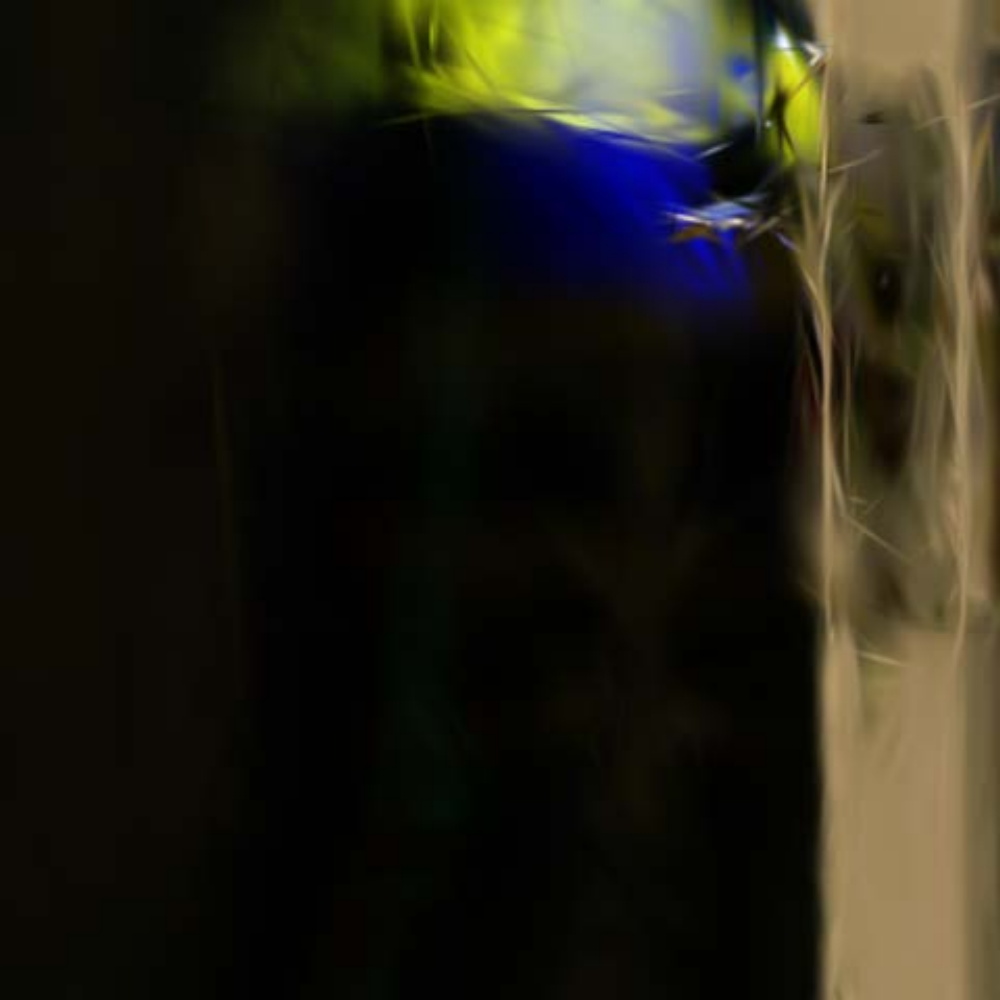}\\[2.5pt]
\AAZIROW{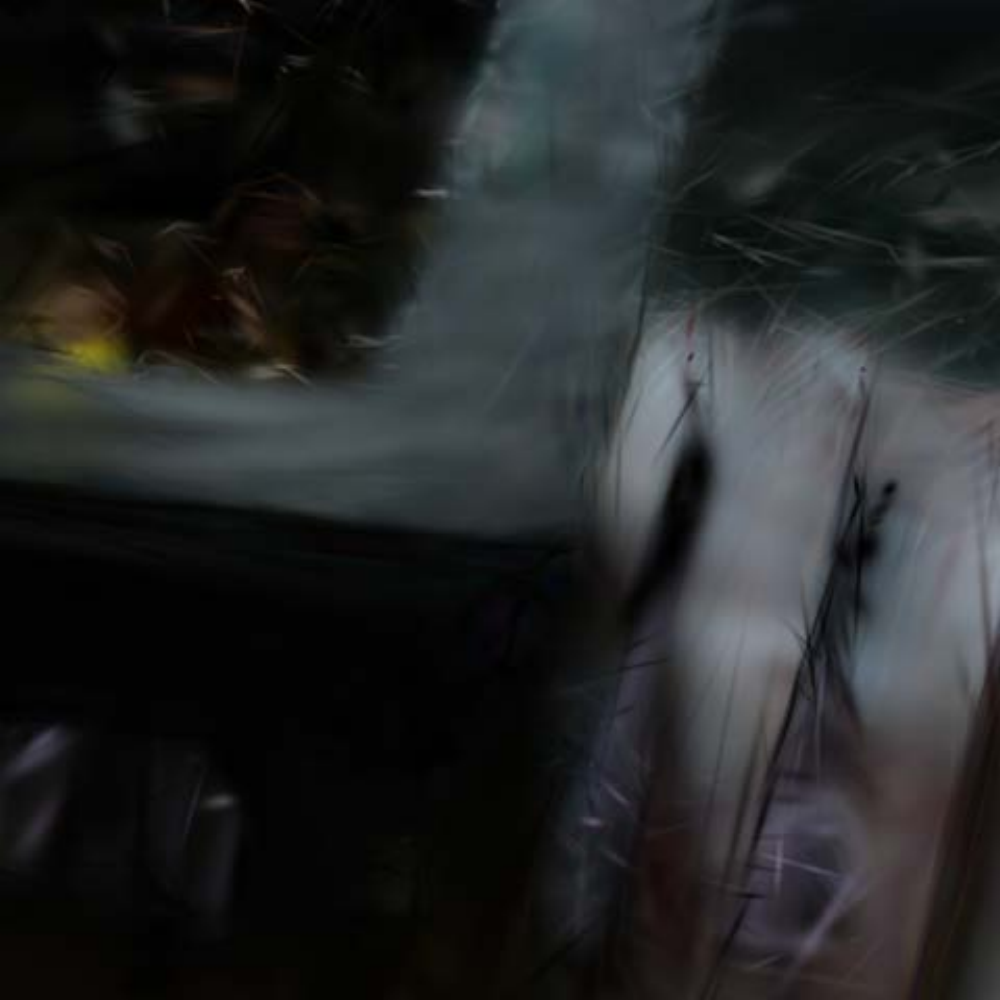}\\[2.5pt]
\AAZIROW{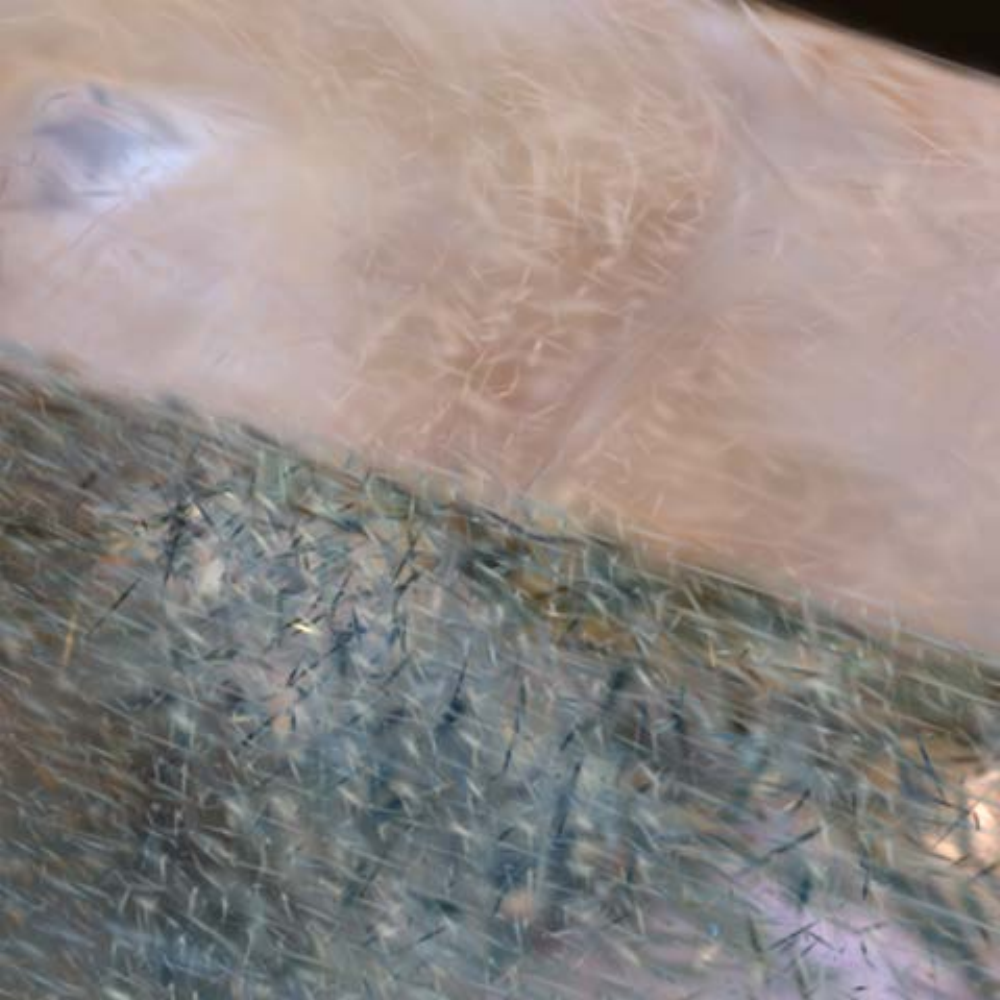}\\[2pt]
\AAZLBL\\[7pt]
{\footnotesize\sffamily\bfseries (b) zoom-out: trained at full resolution, rendered
down to $1/8$}\\[2pt]
\AAZOROW{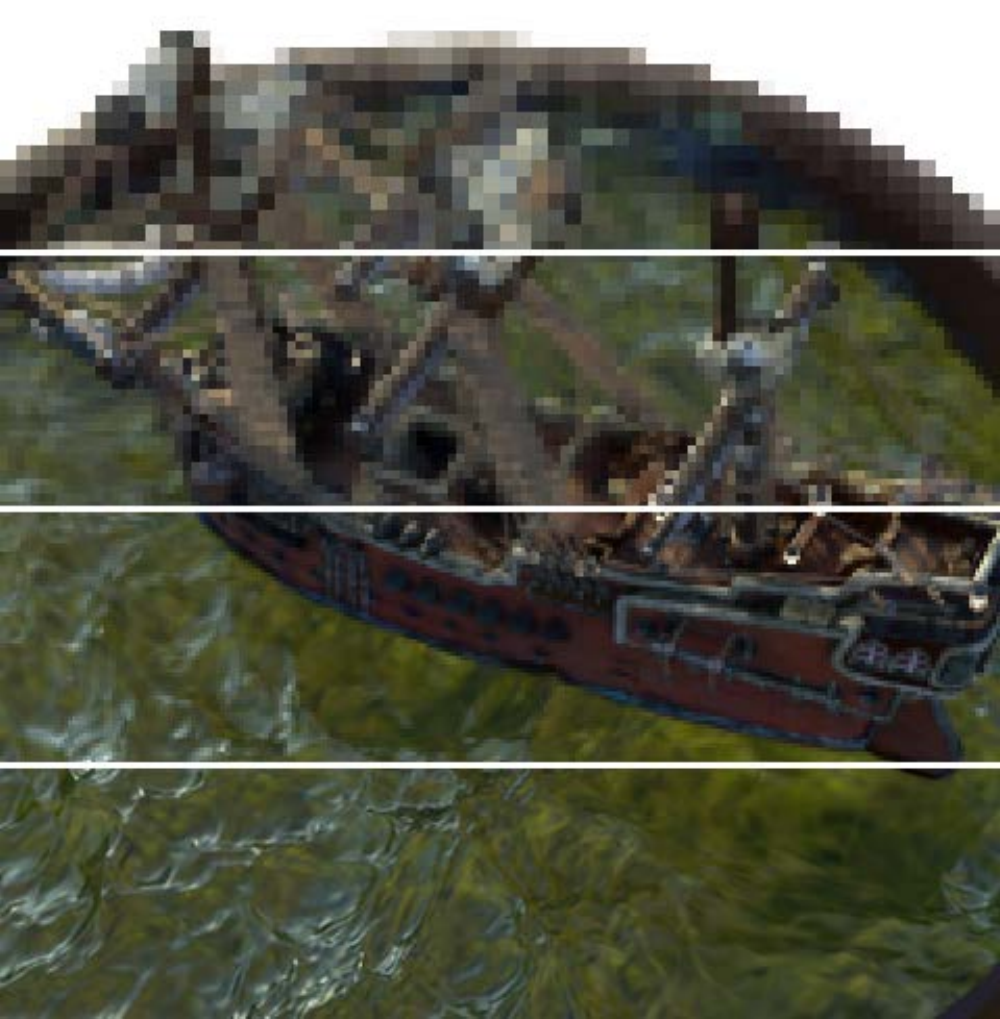}\\[2.5pt]
\AAZOROW{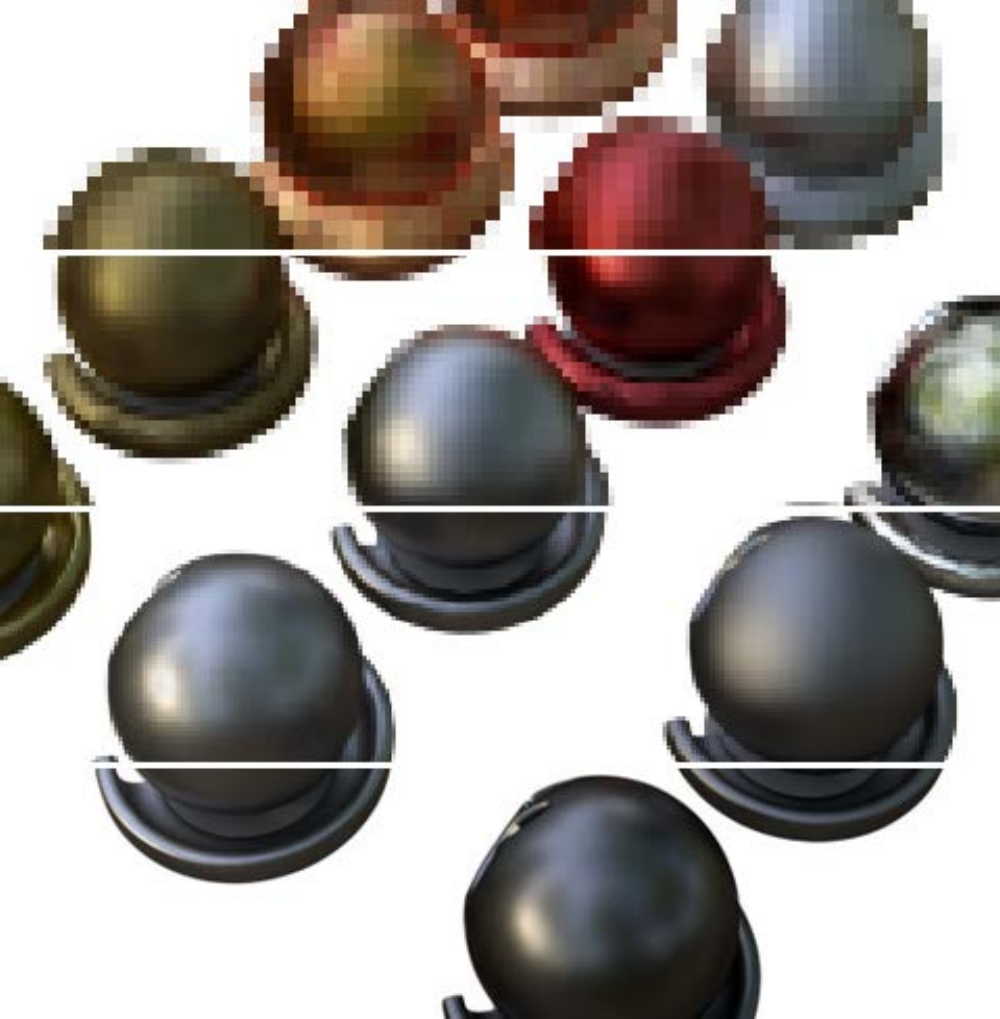}\\[2.5pt]
\AAZOROW{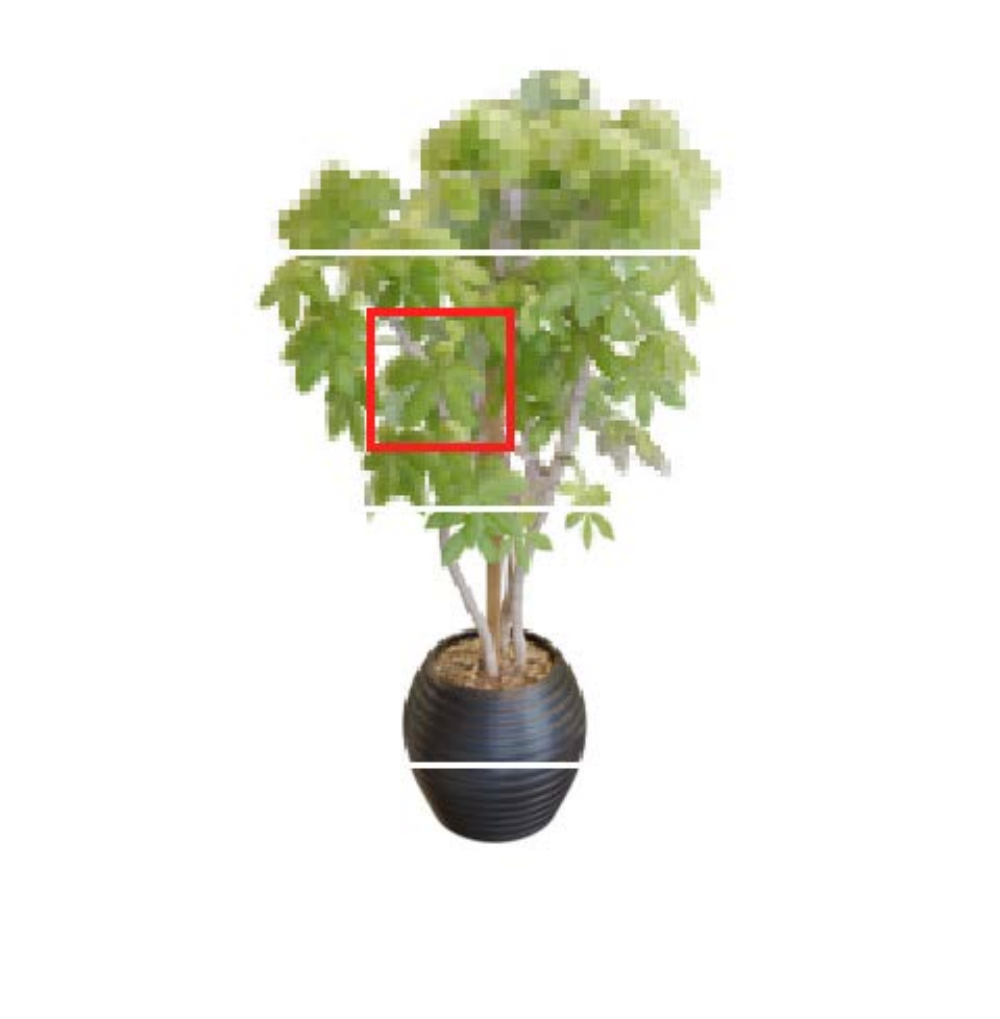}\\[2pt]
\AAMETHODS
\caption{Qualitative examples of anti-aliasing under single-scale training, multi-scale testing.}
\label{fig:aa-qual}
\vspace{-0.5cm}
\end{figure*}

\subsection{Defocus and Refocus}
\label{sec:supp-dof}

A finite aperture spreads the bundle over directions rather than over the image: the queries
of a pixel leave from across the aperture instead of from one center. A thin lens of
aperture $K_j$ focused at $F_j$ spreads them over the circle of confusion, of angular radius
\begin{equation}
\label{eq:supp-sigma}
\sigma_{ij}=\frac{K_j}{2}\Bigl\lvert\frac{1}{F_j}-\frac{1}{Z_{ij}}\Bigr\rvert,
\end{equation}
the standard thin-lens quantity also used by~\citet{wu2022dof,wang2025dof}, with $Z_{ij}$
the depth already computed for compositing (Sec.~\ref{sec:m-render}). It vanishes at
$Z_{ij}=F_j$, so a primitive on the focal plane is unaffected at any aperture, and it is
linear in $K_j$.

To reach $P$ the spread is carried into residual space. Both \eqref{eq:ray-coord} and
\eqref{eq:residual} are explicit, the first rational in the direction and the second affine
in the coordinates, so their composition has a first-order derivative in closed form: about
the central ray of the pixel,
$\mathbf{Q}_{ij}=\partial\mathbf{r}_i/\partial\mathbf{q}\in\mathbb{R}^{2\times2}$ is fixed
for primitive $i$ in view $j$, with $\mathbf{q}\in\mathbb{R}^{2}$ parameterizing the
direction on the normalized image plane. A circular aperture spreads $\mathbf{q}$
isotropically, so $P=\sigma_{ij}^{2}\mathbf{Q}_{ij}\mathbf{Q}_{ij}^{\top}$ and
\eqref{eq:supp-conv} gives
\begin{equation}
\label{eq:dof}
\Sigma_i\;\to\;\Sigma_i+\sigma_{ij}^{2}\,
\mathbf{Q}_{ij}\mathbf{Q}_{ij}^{\top}.
\end{equation}
The only quantities added are $K_j$ and $F_j$, two scalars per training image optimized with
the scene; $\mathbf{Q}_{ij}$ follows from the parameterization and is not learned. Two
approximations enter and no others: the spread is carried to first order through the
constant $\mathbf{Q}_{ij}$, and the disc is replaced by a Gaussian of matching second moment
so that \eqref{eq:supp-conv} applies. Neither the response \eqref{eq:response} nor the
compositing is approximated.

Applying \eqref{eq:dof} when rendering the training views, whose images are defocused, and
dropping it at test time forces a scene that renders sharp to account for blurred
observations, which is defocus deblurring. And since $K_j$ and $F_j$ describe the camera
rather than the scene, they stay free once training ends: sweeping $F_j$ moves the focus
through a trained scene and raising $K_j$ renders a depth of field at apertures never
captured, recovering the synthetic aperture of classical light field
rendering~\citep{isaksen2000dynamically} without its dense table.

In a representation anchored in 3D, no coordinate records where within the aperture a query
passes, so the aperture has to be stated in the operator instead: an analytic circle of
confusion attached to each projected footprint~\citep{wang2025dof}, a network predicting a
per-primitive deformation~\citep{lee2024deblurring}, or blur kernels fitted per
view~\citep{peng2024bags}. Here it lands on the covariance the renderer already reads,
beside the term for the pixel.

\subsection{The peak factor at training and at test time}
\label{sec:supp-peak}

The peak factor of \eqref{eq:supp-conv} is treated differently in the two settings, and the
reason is which quantities are still free.
While fitting a scene to defocused images, $o_i$ and the per-image $K_j,F_j$ are optimized
through the same loss and settle at whatever reproduces the observations with the factor
absent, so we do not apply it. Supplying it explicitly does not help: retraining with the
factor raised to a range of powers was worse on every metric at every power, with a monotone
return toward the released behaviour as the power approaches zero, which is what one expects
if the factor is already absorbed.

Refocusing a trained scene is the opposite case. There $o_i$ is frozen at a value fitted
under the training footprint and a new $(F,K)$ is imposed that no training image carried, so
the factor is a real change to the render and is applied exactly as in \eqref{eq:aa}, with
the aperture term in place of the scale term. Because \eqref{eq:dof} approximates a disc by
a Gaussian inside an alpha composite, the result is not exact: measured against a reference
that samples the aperture directly, the closed form is a few percent darker over a frame,
and a thin bright feature against a dark background loses about a third of its brightness,
since such a feature is carried by few primitives whose footprints grow fastest relative to
their original size.

\section{Ablation Study}
All ablations are run on the 9 scenes from~\citet{barron2022mip} under the setting of Tab.~\ref{tab:nvs}.

\begin{table}[t]
\centering
\small
\caption{Directional appearance model comparisons, i.e., SH versus ASG, on the 9 scenes of Mip-NeRF~360. The last column counts the floats each primitive spends on appearance. The same three models are attached to 3DGS for reference.}
\label{tab:ablation}
\setlength{\tabcolsep}{6pt}
\begin{tabular}{llcccc}
\toprule
& Appearance & PSNR $\uparrow$ & SSIM $\uparrow$ & LPIPS $\downarrow$ & Floats \\
\midrule
\multirow{3}{*}{\rotatebox[origin=c]{90}{3DGS}}
& RGB only    & 26.82 & 0.804 & 0.268 & 3 \\
& SH          & 27.46 & 0.815 & 0.255 & 48 \\
& ASG         & 27.69 & 0.816 & 0.255 & 30 \\
\midrule
\multirow{3}{*}{\rotatebox[origin=c]{90}{LFP}}
& RGB only    & 26.76 & 0.798 & 0.264 & 3 \\
& SH          & 27.57 & 0.808 & 0.256 & 48 \\
& ASG         & \best{28.02} & \best{0.820} & \best{0.247} & 30 \\
\bottomrule
\end{tabular}
\end{table}

\newlength{\ABc}\setlength{\ABc}{0.185\textwidth}
\newlength{\ABgap}\setlength{\ABgap}{0.0028\textwidth}
\newcommand{\ABP}[1]{\includegraphics[width=\ABc]{asg_abl_#1}}
\newcommand{\ABROW}[1]{%
  \ABP{#1_gt.pdf}\hspace*{\ABgap}\ABP{#1_asg.pdf}\hspace*{\ABgap}%
  \ABP{#1_sh.pdf}\hspace*{\ABgap}\ABP{#1_rgb.pdf}}
\newcommand{\ABL}[1]{\makebox[\ABc]{\footnotesize\sffamily #1}}
\newcommand{\ABLBLS}{%
  \ABL{GT}\hspace*{\ABgap}\ABL{+\,ASG}\hspace*{\ABgap}%
  \ABL{+\,SH}\hspace*{\ABgap}\ABL{RGB only}}
\begin{figure}[t]
\centering
{\footnotesize\sffamily\bfseries (a) LFP}\\[1.5pt]
\ABROW{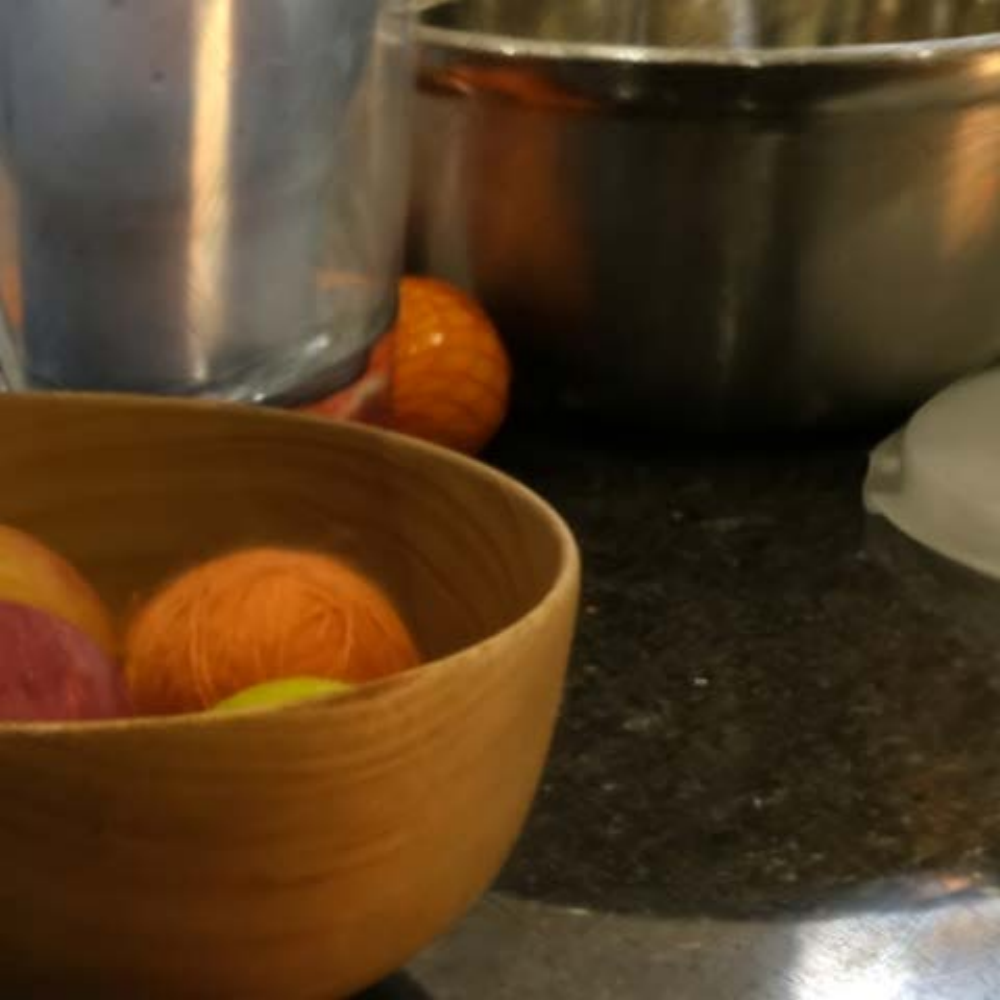}\\[2pt]
\ABROW{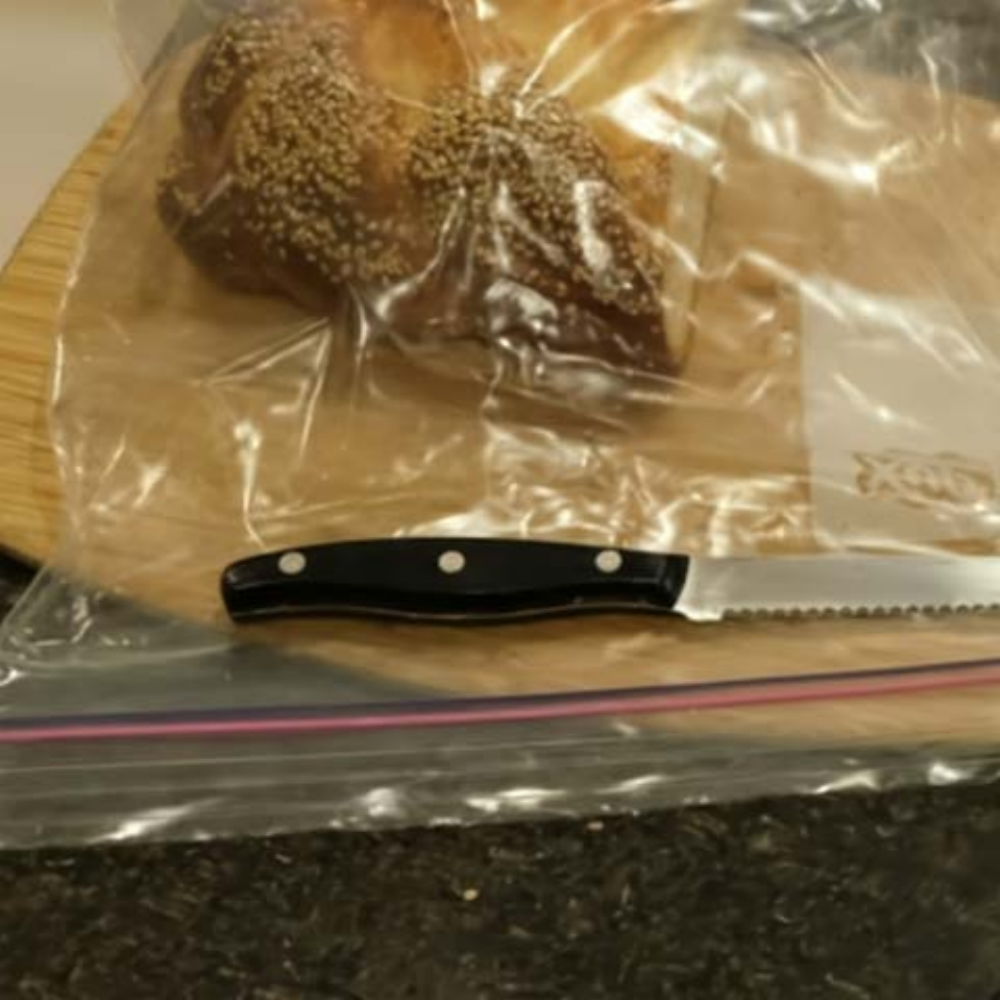}\\[1.5pt]
\ABLBLS\\[6pt]
{\footnotesize\sffamily\bfseries (b) 3DGS}\\[1.5pt]
\ABROW{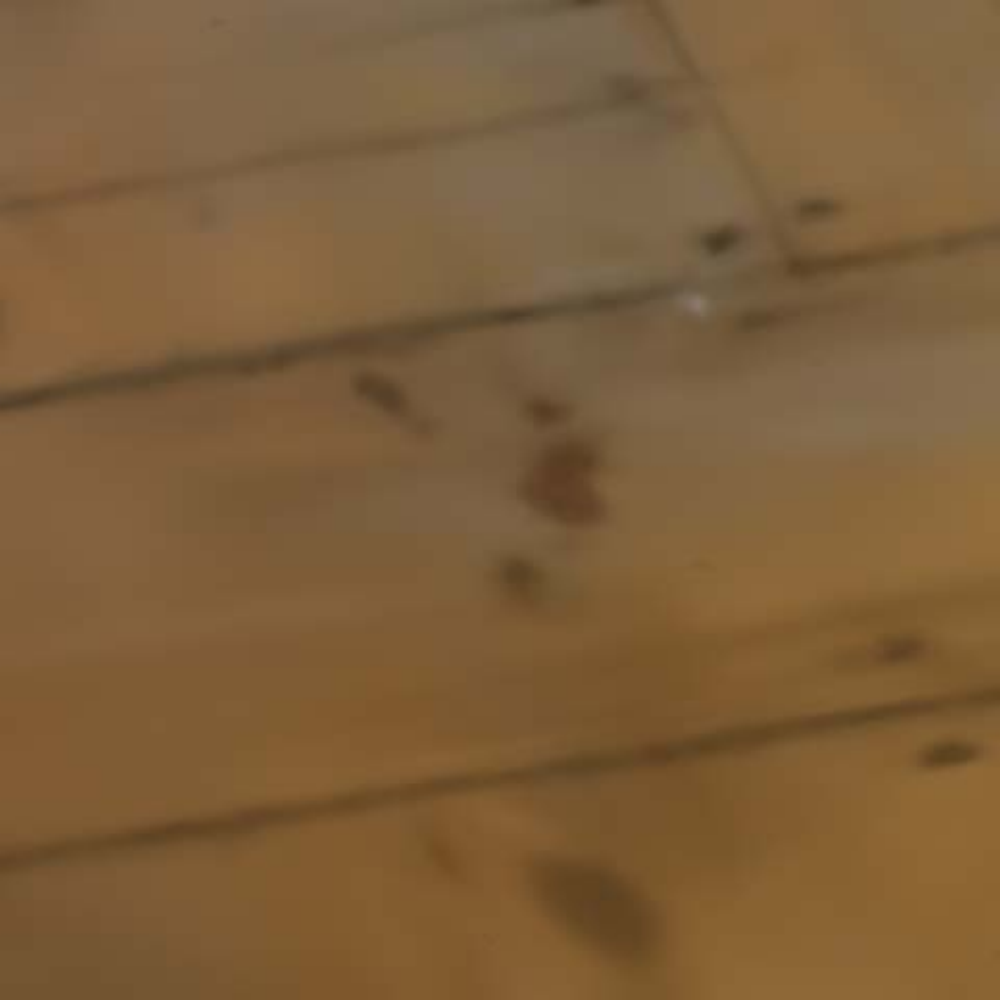}\\[1.5pt]
\ABLBLS
\caption{Appearance-model ablation. The view-dependent appearance model is
replaced by spherical harmonics (SH, degree~3) or by a view-independent
colour (RGB only), under an otherwise identical training protocol:
(a) inside LFP, (b) inside 3DGS.}
\label{fig:asg-ablate}
\vspace{-0.5cm}
\end{figure}

\paragraph{View-dependent term.}
Tab.~\ref{tab:ablation} replaces the three ASG lobes with the spherical
harmonics of 3DGS~\citep{kerbl20233d} and with a constant colour, and
repeats the three on 3DGS itself. Both sides use the SH basis and constants
of 3DGS, each evaluated in the direction that representation provides: for
3DGS the direction from the camera to a primitive, constant over its extent
on screen, and for LFP the direction of the query ray, which the
parameterization carries per pixel. Within each side the two bases share
that direction, so the comparison is controlled. The lobes differ in one
further respect: an LFP primitive carries a direction of its own, that of
its anchor ray (Sec.~\ref{sec:m-impl}), so its lobes are evaluated in the
reflection about it, while a 3D Gaussian carries none and its lobes stay in the same viewing direction its harmonics use. Lobe axes are learned on both sides, and the appearance budget is matched at 30 floats per primitive.

With a constant colour the two representations are level, so whatever
difference appears later comes from the directional term alone. On 3DGS the
choice hardly matters, lobes and harmonics being close: with only the
viewing direction to work with, the lobes act as one more basis over
directions. On LFP they are ahead by roughly twice the margin with fewer
floats, because they can reflect. A ray group collects rays with coherent radiance, so the base carries what holds across the group, and
what the lobes are left with is the highlights, which is what a reflection about a normal describes.

Fig.~\ref{fig:asg-ablate}~(a) isolates the effect inside our representation.
With a view-independent colour (RGB only), a primitive has to explain all of
its observations with a single value, so specular energy is averaged away:
the reflections on the steel pot and the specular wrinkles of the plastic bag
flatten into a matte surface, and the structure they reveal -- the rim of the
pot, the serrated blade behind the bag -- loses its contrast. Degree-3 SH
brings part of that energy back, but only at low angular frequency: the
highlights return as broad, smeared glows whose position and extent do not
follow the ground truth, while the fine creases of the bag stay washed out.
This is not a matter of capacity: at degree 3 the SH block carries 45
coefficients per primitive against the 27 of our three ASG lobes. What
differs is where those coefficients can be spent -- SH spreads them
isotropically over the whole sphere, whereas a single ASG lobe concentrates
them on the narrow, anisotropic angular support that a highlight actually
occupies. Our full model reproduces both the intensity and the shape of these
reflections.
Fig.~\ref{fig:asg-ablate}~(b) repeats the same substitution inside 3DGS,
replacing its spherical harmonics by our ASG lobes and leaving the rest of the
pipeline untouched. The trend is unchanged: SH renders the sheen of the wooden
floor pale and flat, while ASG recovers its warm tone and the grain that the
sheen reveals. The gain therefore comes from the angular basis itself rather
than from our primitive, so ASG can be dropped into existing Gaussian
splatting appearance models, and its advantage is largest wherever radiance
varies sharply with view direction.

\begin{wrapfigure}{r}{0.4\linewidth}
  \vspace{-1.2\baselineskip}
  \centering
  \includegraphics[width=\linewidth]{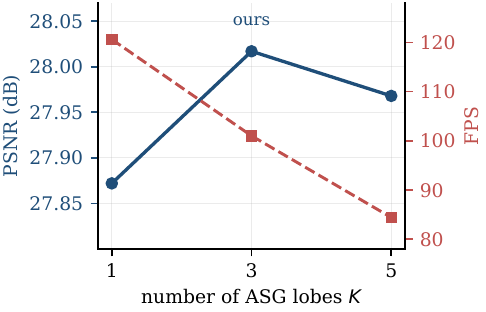}
  \vspace{-0.7cm}
  \caption{Lobe count $K$: quality peaks at $3$, cost grows monotonically.}
  \label{fig:ablation_lobes}
  \vspace{-0.8\baselineskip}
\end{wrapfigure}

\paragraph{Different lobes for ASG.}
The number of ASG lobes $K$ is a compile-time constant of the rasteriser: every per-pixel lobe
loop and per-thread cache is sized by it, so changing $K$ means rebuilding the kernel rather than
flipping a flag. We retrain all nine Mip-NeRF~360 scenes for $K\!\in\!\{1,3,5\}$, holding the atlas
count, initialisation and schedule fixed, and keeping one out-of-sigmoid lobe whenever $K\!>\!1$.
Quality peaks at $K\!=\!3$ (28.02~dB PSNR, 0.820 SSIM, 0.247 LPIPS): dropping to a single lobe
costs 0.15~dB (27.87~dB) while raising $K$ to five does not recover it (27.97~dB), because the
extra lobes add per-primitive work without adding usable angular detail -- the optimiser in fact
settles on 3\% fewer primitives at $K\!=\!5$ than at $K\!=\!3$. Rendering cost, in contrast, is
monotone in $K$: 120.6~FPS at $K\!=\!1$, 101~FPS at $K\!=\!3$ and 84.4~FPS at $K\!=\!5$. $K\!=\!3$
is therefore the knee of the quality--speed curve (Fig.~\ref{fig:ablation_lobes}) and is what we
ship; $K\!=\!1$ remains a useful operating point when throughput matters more than the last
0.15~dB.

\paragraph{Highlight regularization.}
Removing $\lambda_{\mathrm{g}}\lVert\mathbf{g}\rVert^{2}$ from
\eqref{eq:loss} costs \textbf{0.17}\,dB (28.02\,$\to$\,27.85 averaged over the
nine Mip-NeRF~360 scenes, with eight of the nine degrading), together with
0.005 SSIM and 0.008 LPIPS. Without it the lobes are free to reproduce
radiance that the base $\mathbf{c}_i$ should carry, since a highlight
centred near the training directions is cheaper to fit than a correction to
the base, and the split between the two is then set by initialization rather
than by the data. Tab.~\ref{tab:highlight} shows the effect: \textbf{the mean
penalized lobe amplitude grows $6.6\times$ (0.098\,$\to$\,0.640) while the
base colour is left essentially untouched ($1.03\times$), i.e.\ the surplus
radiance is absorbed by the lobes rather than by $\mathbf{c}_i$}. The model
also ends up \textbf{12\% smaller} (2.00\,M\,$\to$\,1.75\,M primitives),
because the lobes soak up photometric error that would otherwise drive
densification.

\paragraph{Atlas design.}
A single plane pair cannot index rays that run nearly parallel to it
(Sec.~\ref{sec:m-impl}), so the number of atlases sets how much of the ray
space is reachable. Tab.~\ref{tab:atlas} varies that number. With one atlas
the reconstruction loses $6.92$\,dB, and the loss is of the same order indoors
and outdoors, $7.12$ against $6.76$\,dB, so what is missing is coverage rather
than anything particular to a capture pattern. Four atlases recover $6.62$ of
those $6.92$\,dB, eight a further $0.18$, and sixteen another $0.12$, while
each doubling raises the per-camera precomputation and places more primitives
on the camera-frame ordering of Sec.~\ref{sec:m-render}. Thirty-two adds
$0.01$\,dB and is behind sixteen outdoors, so coverage is saturated by then.
We use sixteen, the smallest count on that plateau.

Letting the atlas normals be optimized rather than fixed before training never
helps, and at the largest step size we tried it costs $0.18$\,dB. A moving
normal changes the coordinates of every query it serves, so the residuals of
all primitives assigned to that atlas shift together, and the gradient it
receives averages over them rather than correcting any one. The measurements
show this cancellation. At $10^{-5}$ and $10^{-4}$ the normals barely rotate,
$0.10^\circ$ and $0.28^\circ$ after 30k iterations, and every metric is
unchanged to within noise. Only at $10^{-3}$ do they move appreciably,
$0.60^\circ$ on average and $6.3^\circ$ at most, and there all three metrics
degrade, the indoor scenes by $0.36$\,dB against $0.02$\,dB outdoors, which
have fewer primitives per atlas and so fewer residuals to average. Fixing the
normals at the cluster centres of the training camera directions leaves this
degree of freedom out of the optimization, and the primitives absorb the
difference through $p_i$ and $(\mu^s_i,\mu^t_i)$.

\begin{table*}[t]
\centering\small
\caption{Atlas design, each metric reported
over all scenes and split into the five outdoor and four indoor ones. Top:
number of atlases, normals fixed at the cluster centres of the training
camera directions. Bottom: the same 16 atlases with the normals left free, at three step sizes, larger learning rate (lr) means more movement for the intial normals.}
\label{tab:atlas}
\begin{tabular}{lccc@{\hskip 12pt}ccc@{\hskip 12pt}ccc}
\toprule
& \multicolumn{3}{c}{PSNR\,$\uparrow$} & \multicolumn{3}{c}{SSIM\,$\uparrow$} & \multicolumn{3}{c}{LPIPS\,$\downarrow$} \\
\cmidrule(lr){2-4}\cmidrule(lr){5-7}\cmidrule(lr){8-10}
& all & outdoor & indoor & all & outdoor & indoor & all & outdoor & indoor \\
\midrule
\multicolumn{10}{l}{\emph{Number of atlases}}\\
\quad 1  & 21.10 & 17.95 & 25.03 & 0.657 & 0.512 & 0.838 & 0.393 & 0.460 & 0.310 \\
\quad 4  & 27.72 & 24.34 & 31.95 & 0.811 & 0.718 & 0.929 & 0.257 & 0.276 & 0.234 \\
\quad 8  & 27.90 & 24.64 & 31.98 & 0.817 & 0.728 & 0.929 & 0.251 & 0.264 & 0.234 \\
\quad 16 (ours) & {28.02} & \best{24.71} & {32.15} & \best{0.820} & \best{0.731} & {0.930} & \best{0.247} & \best{0.259} & {0.232} \\
\quad 32 & \best{28.03} & {24.70} & \best{32.19} & {0.819} & {0.730} & \best{0.931} & \best{0.247} & {0.260} & \best{0.231} \\
\midrule
\multicolumn{10}{l}{\emph{16 atlases, normals learned}}\\
\quad lr $10^{-5}$ & 27.97 & 24.69 & 32.08 & 0.818 & 0.729 & 0.930 & 0.249 & 0.262 & 0.232 \\
\quad lr $10^{-4}$ & 27.96 & 24.68 & 32.08 & 0.818 & 0.729 & 0.930 & 0.249 & 0.262 & 0.232 \\
\quad lr $10^{-3}$ & 27.84 & 24.69 & 31.79 & 0.815 & 0.727 & 0.925 & 0.254 & 0.268 & 0.235 \\
\bottomrule
\end{tabular}
\end{table*}

\begin{table}[t]
\centering
\caption{Highlight regularization, averaged over the nine Mip-NeRF~360 scenes.
$\lVert\mathbf{g}\rVert$ is the mean amplitude of the penalized lobes and
$\lVert\mathbf{c}\rVert$ the mean base-colour magnitude. Dropping the term
lets the lobes grow by $6.6\times$ while the base is left untouched.}
\label{tab:highlight}
\begin{tabular}{lccccccc}
\toprule
& PSNR\,$\uparrow$ & SSIM\,$\uparrow$ & LPIPS\,$\downarrow$
& $\lVert\mathbf{g}\rVert$ & $\lVert\mathbf{c}\rVert$ & \#prim. \\
\midrule
with $\lambda_{\mathrm{g}}\lVert\mathbf{g}\rVert^{2}$ (ours)
  & \best{28.02} & \best{0.820} & \best{0.247} & 0.098 & 1.79 & 2.00\,M \\
without
  & 27.85 & 0.815 & 0.255 & 0.640 & 1.85 & 1.75\,M \\
\midrule
$\Delta$ & $-0.17$ & $-0.005$ & $+0.008$ & $\times 6.57$ & $\times 1.03$ & $-12\%$ \\
\bottomrule
\end{tabular}
\end{table}

\paragraph{Optimizing $(\mu^u,\mu^v)$.}
A primitive is anchored by two plane coordinates, $(\mu^u_i,\mu^v_i)$ on the
entry plane and $(\mu^s_i,\mu^t_i)$ on the exit plane; by default we optimize
only the latter and keep $(\mu^u_i,\mu^v_i)$ at its initial value. Enabling
gradients is not enough to change this, since the backward kernel never writes
$\partial\mathcal{L}/\partial\boldsymbol{\mu}^{uv}_i$ and the gradient reaching
those parameters is identically zero. We therefore add the missing term. By
\eqref{eq:residual} the residual depends on the entry coordinates through the
coupling coefficient, so
$\partial\mathbf{r}_i/\partial\boldsymbol{\mu}^{uv}_i=+(1-p_i)\mathbf{I}$
alongside the existing
$\partial\mathbf{r}_i/\partial\boldsymbol{\mu}^{st}_i=-\mathbf{I}$, and we
retrain all nine scenes at the learning rate used for $(\mu^s_i,\mu^t_i)$.

The result is unchanged within run-to-run noise, $28.01$\,dB / $0.820$ /
$0.247$ against $28.02$\,dB / $0.820$ / $0.247$ for the anchored model, with
per-scene differences spanning $-0.07$ to $+0.03$\,dB and no consistent sign.
The parameters do move. Over a $3000$-step probe their cumulative displacement
is $57\%$ of that of $(\mu^s_i,\mu^t_i)$, and a ten-fold learning rate moves
them five times further without improving any metric, so the outcome is not an
artefact of under-training.

This is what the parameterization predicts. A ray group is fixed by its
disparity and any one of its rays (Sec.~\ref{sec:m-prim}), so $\boldsymbol{\mu}_i$
carries one degree of freedom that names the group and one that only chooses
which member of it is stored. Optimizing $(\mu^s_i,\mu^t_i)$ already spans the
first, and what freeing $(\mu^u_i,\mu^v_i)$ adds is the second, which no
photometric loss can see: the response \eqref{eq:response} depends on the group
and not on which of its rays the record was written against. We therefore keep
$(\mu^u_i,\mu^v_i)$ fixed, which removes two learnable parameters and two atomic
accumulations per primitive at no cost in quality.

\end{document}